\documentclass[5p]{elsarticle}
\pdfoutput=1
\usepackage{subfigure}
\usepackage{lineno}
\usepackage[cmex10]{amsmath}
\usepackage{algorithm}
\usepackage{algorithmic}
\usepackage{multirow}
\usepackage{array}
\usepackage{mdwmath}
\usepackage{mdwtab}
\usepackage{eqparbox}
\usepackage{amssymb}
\usepackage{footnote}
\usepackage{threeparttable}
\usepackage{setspace}
\usepackage{stfloats}
\usepackage{url}

\begin{document}

\begin{frontmatter}

\title{Robust Face Recognition with Structural Binary Gradient Patterns}


\author[mymainaddress,mythirdaddress]{Weilin Huang}
\ead{wl.huang@siat.ac.cn}

\author[mysecondaryaddress]{Hujun Yin\corref{mycorrespondingauthor}}
\cortext[mycorrespondingauthor]{Corresponding author}
\ead{h.yin@manchester.ac.uk}

\address[mymainaddress]{Shenzhen Institutes of Advanced Technology, Chinese Academy of Sciences, Shenzhen, 518055, China.}
\address[mysecondaryaddress]{School of Electrical and Electronic Engineering, The
University of Manchester, Manchester, M60 1QD, UK.}

\address[mythirdaddress]{Multimedia Laboratory, the Chinese University of Hongkong, Shatin, N.T., Hongkong.}

\begin{abstract}
This paper presents a computationally efficient yet powerful binary
framework for robust facial representation based on image gradients. It is termed as structural binary gradient patterns (SBGP). To discover underlying local structures in the gradient domain, we compute image gradients from multiple directions and simplify them into a set of binary strings. The SBGP is derived from certain types of these binary strings that have meaningful local structures and are capable of resembling fundamental textural information. They detect micro orientational edges and possess strong orientation and locality capabilities, thus enabling great discrimination. The SBGP also benefits from the advantages of the gradient domain and exhibits profound robustness against illumination variations. The binary strategy realized by pixel correlations in a small neighborhood substantially simplifies the computational complexity and achieves extremely efficient processing with only $0.0032s$ in Matlab for a typical face image. Furthermore, the discrimination power of the SBGP can be enhanced on a set of defined orientational image gradient magnitudes, further enforcing locality and orientation. Results of extensive experiments on various benchmark databases illustrate significant improvements of the SBGP based representations over the existing state-of-the-art local descriptors in the terms of discrimination, robustness and complexity. Codes for the SBGP methods will be available at \emph{http://www.eee.manchester.ac.uk/research/groups/sisp/software/}.
\end{abstract}


\end{frontmatter}


\section{Introduction}

Face recognition has been one of the most active topics in image and pattern recognition due to its much increased attention and applications in law enforcement, surveillance, human-computer interaction, etc. Although tremendous progresses have been made over the last two decades, it is still regarded as an unsolved problem in real-world situations, where large within-class variations in facial appearance exist (e.g. illuminations, expressions and poses). A key solution lies in facial representation and a great deal of effort has been devoted to it. A desirable facial descriptor should be discriminative to inter-person differences but robust to intra-person variations, and at the same time, efficient to process.


Appearance-based methods, one of widely adopted approaches, consider face images as holistic vectors of pixel intensities in high-dimensional space. Dimensionality reduction and manifold techniques are typically applied to reduce the dimensionality and to extract intrinsic features
\cite{Huang2012}. Typical early examples are Eigenfaces \cite{Turk1991} and Fisherfaces \cite{Belhumeur1997} where linear PCA is used. They have been enhanced by nonlinear PCA and manifold methods \cite{Scholkopf1998,Roweis2000,Yin2010,Huang2009}. However, faces represented by pixel intensities are sensitive to variations such as occlusions, illumination, expression and pose.

Zhang \emph{et al}.\cite{Zhang2009b} have proposed a novel descriptor, termed as the Gradientfaces, to extract illumination insensitive features in the image gradient domain. Faces are described by using image gradient orientation (IGO) instead of intensity to achieve strong robustness to illumination change. To further take advantage of gradient features, Tzimiropoulos \emph{et al}.\cite{Tzimiropoulos2012} derived a similarity measure based on cosine of IGO differences between images (termed as $IGO_{cos}$) and showed that the measure considerably mitigated the effect of variations and enhanced PCA based recognition. However, the similarity measures of the Gradientfaces and $IGO_{cos}$ are based on pixel-wise correlations, and hence are holistic representations, which are sensitive to local deformations, rotations  and spatial scales. That is, they are prone to facial variations such as expressions and poses \cite{Heisele2003,Vu2012,Huang2010}.


Local feature descriptors have recently gained considerable attention due to their resilience to multiple visual variations by enforcing spatial locality in both pixel and patch levels. Two of the most successful local descriptors are Gabor wavelets \cite{Manjunath1996} and local binary patterns (LBP) \cite{Ojala1996, Ojala2002}. Gabor features extract both micro texture details and global shape information from spatial and spatial-frequency domains and are robust to local distortions, leading to certain successes in face recognition \cite{Liu2002,Zhang2005,Zou2007}. However, Gabor representations are time-consuming to extract and also generate a large number of features with the convolution kernels, making them prohibitive for real-time applications. Whilst, the LBP features are simple, efficient and yet resistant to illumination changes and they are also capable of detecting micro texture, e.g. spots, corners and edges\cite{Ojala2002}. However, the capability of the LBP descriptor can be severely affected by drastic changes of pixel intensity, such as extreme lighting. Most current local facial descriptors that are built on the Gabor and LBP also suffer from these inherent limitations \cite{Chen2010,Tan2011,Zhang2005,Zhang2007}.

Building on the properties gained from the IGO domain and local binary features, this paper presents a new local facial descriptor, termed as the structural binary gradient patterns (SBGP), for facial images. It measures relationships between local pixels in the image gradient domain and effectively encodes the underlying local structures into a set of binary strings, not only increasing the discriminative power but also significantly simplifying the complexity. We observe that the structural patterns of SBGP are capable of detecting stable micro edge texture from various directions. Local features built on the histogram statistics from these orientational edge textures contain the primary structural information of biological vision systems and exhibit desirable characteristics of spatial locality, orientation and scale selectivity. They show stronger orientational power than the LBP and Gabor features, leading to improved discriminative representation. Furthermore, an enhanced descriptor can be devised by building SBGP patterns on a set of orientational image gradient magnitudes (OIGM), termed as SBGPM, to further enforce its locality and orientation. Extensive experiments on several benchmark databases demonstrate the significant advantages of the SBGP-based methods over the-state-of-the-art methods with respect to discrimination, robustness and complexity.

Next, a brief review on related work is given in Section 2. The proposed SBGP descriptor is then presented in Section 3. Section 4 discusses favorable properties of the SBGP descriptor and connections and distinctions among SBGP, LBP and Gabor representations. Section 5 describes the enhanced SBGPM descriptor. Finally, experimental verifications are provided in Section 6, followed by conclusions in Section 7.

\section{Related work}
 Local histograms built on IGO statistics have been considered as visually prominent features and have favorable properties such as invariance against illumination. In\cite{Zhang2009b}, Zhang \emph{et al}. computed Gradientfaces by using IGOs representation instead of intensities to obtain an illumination insensitive measure. They showed that features extracted from gradient domain are more discriminative and robust than those in the intensity domain, and are even more tolerant to illumination variations than the methods based on the reflectance model \cite{Xie2011,Ruiz2009,Chen2006}. Similarly, Tzimiropoulos \emph{et al}.\cite{Tzimiropoulos2012} presented a simple yet robust similarity measure based on IGO representation and cosine of kernels of IGO differences between images ($IGO_{cos}$).  Then PCA subspace is adapted in the IGO space to generate a more compact, discriminant and robust representation, referred to as the $IGOPCA$\cite{Tzimiropoulos2012}.



Recently, a number of local facial descriptors have been derived from the Gabor or LBP features or their combinations. In \cite{Zhang2005}, the local gabor binary pattern histogram sequence (LGBPHS) was proposed by first running Gabor filters on face images and then building LBP histogram features on the resulted Gabor magnitude faces. Similar methods include the histogram of Gabor phase patterns (HGPP)\cite{Zhang2007} and Gabor volume based LBP (GV-LBP) \cite{Lei2011}. The advantages of these methods are built on the virtues of both Gabor and LBP descriptors. However, they commonly suffer from the difficulties of Gabor based representations, i.e. high computational complexity and high dimensionality.

As a simpler approach, Jie \emph{et al.} \cite{Chen2010} proposed a Weber local descriptor (WLD) based on the Weber's Law of human perception system, which states that the noticeable change of a stimulus is a constant ratio of the original stimulus. In \cite{Tan2011}, Tan and Triggs presented local ternary patterns (LTP) by extending LBP to 3-valued codes for increasing its robustness to noise in the near-uniform image regions. Both methods have been shown to be highly discriminative and resistant to illumination changes, extending the advantages of LBP. However, similar to LBP, both descriptors build local relationships in the intensity domain, which can be seriously affected by dramatic changes of pixel intensity.

The proposed SBGP is closely related to the center-symmetric local binary
pattern (CS-LBP) \cite{Heikkila2009} which computes local binary from symmetric neighboring pixels. However, the SBGP differs distinctly in three aspects. First, structural patterns and multiple spatial resolutions are defined in the SBGP. We show some theoretical insights that the structural patterns of SBGP work as oriented edge detectors, a key to discriminative and compact representation. The multiple spatial resolution strategy increases descriptor's flexibility with stronger discriminative power. Second, motivated by the multiple channels strategy of the invariant descriptors such as the SIFT \cite{Lowe2004} and POEM \cite{Vu2012}, we facilitate the SBGP descriptor on a set of  orientational image gradient magnitudes. This further enhances its discriminative power. Finally, the CS-LBP was originally developed for image matching, while the SBGP is proposed for face recognition. The task of face recognition often requires more detailed and robust local features than general features for matching. As it we will be shown in Section 6.2.1 that the CS-LBP is highly sensitive to significant illuminations.

\section{Structural Binary Gradient Patterns}
\begin{figure*}
\centering
\subfigure[Scream]{     \includegraphics[width=3.8cm]{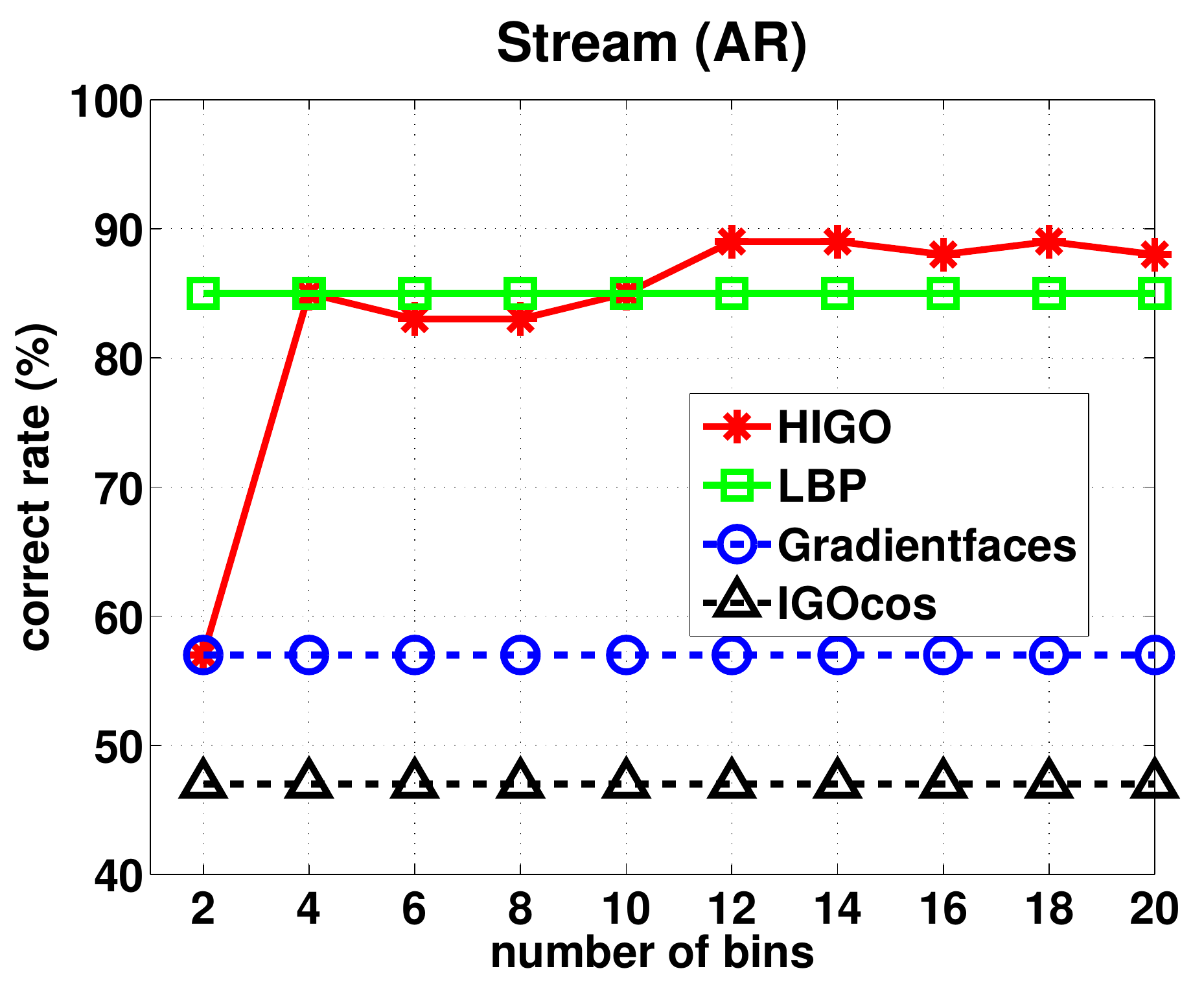}}
\subfigure[Occlusion]{     \includegraphics[width=3.8cm]{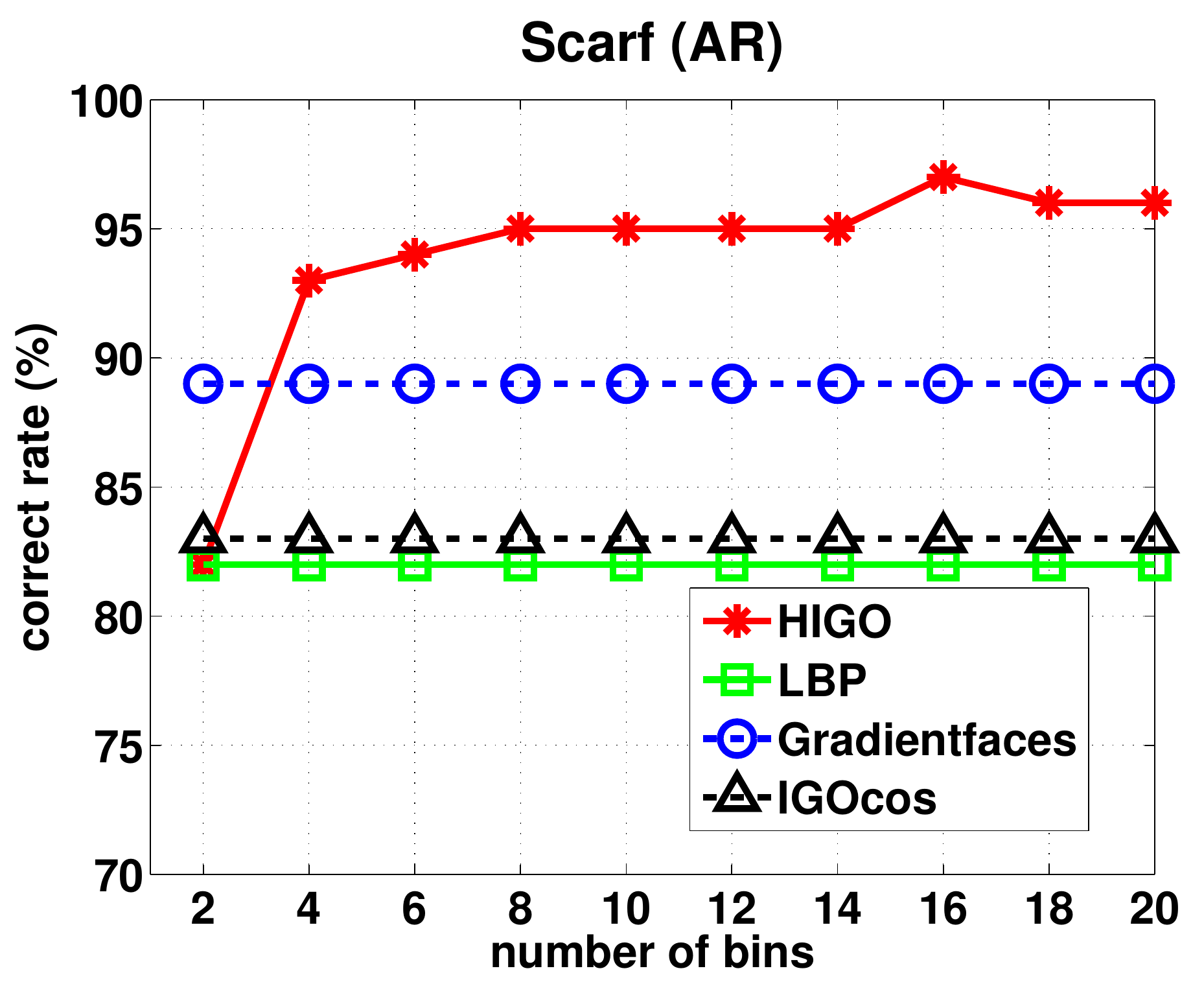}}
\subfigure[Medium illum.]{     \includegraphics[width=3.8cm]{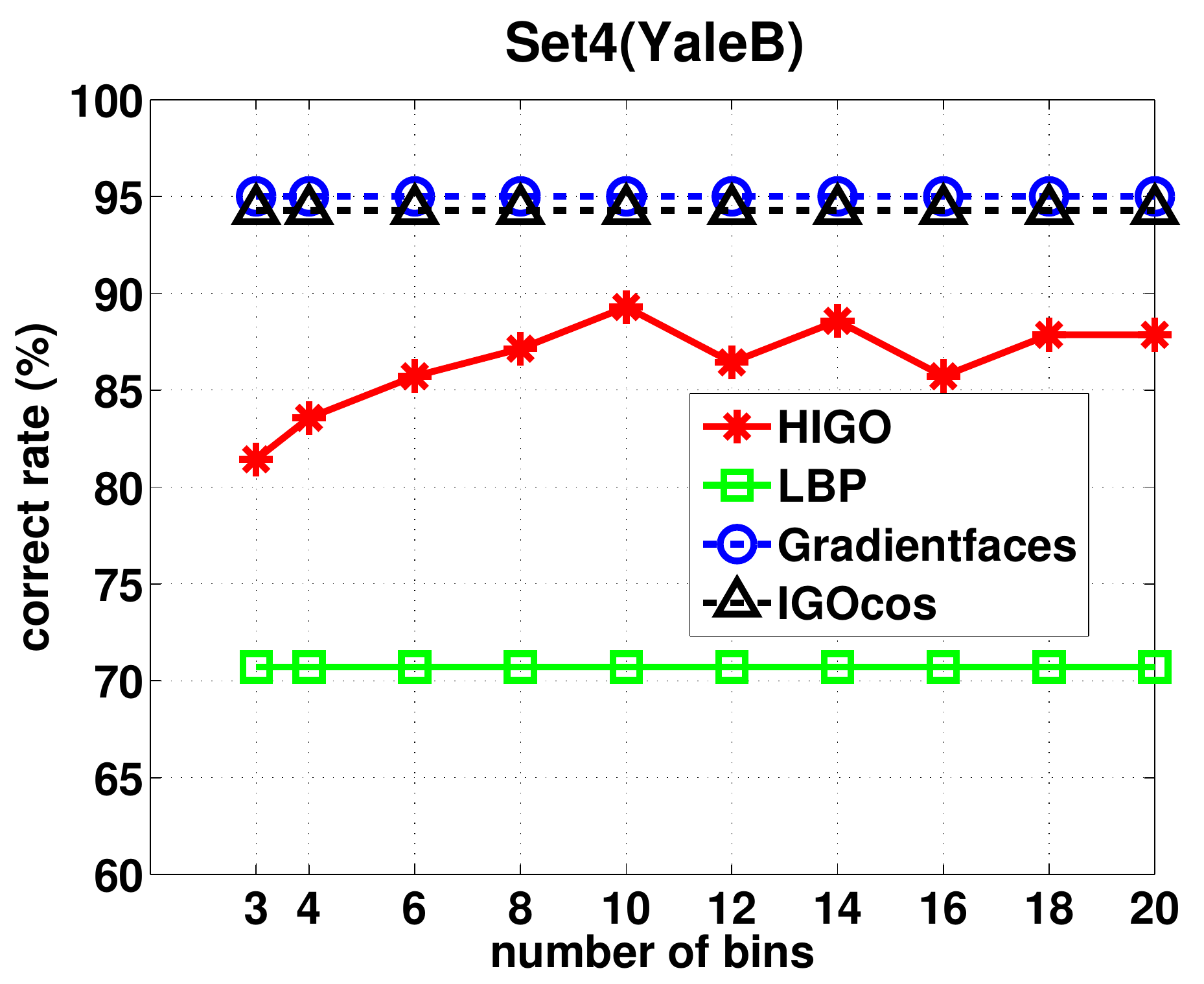}}
\subfigure[High illum.]{    \includegraphics[width=3.8cm]{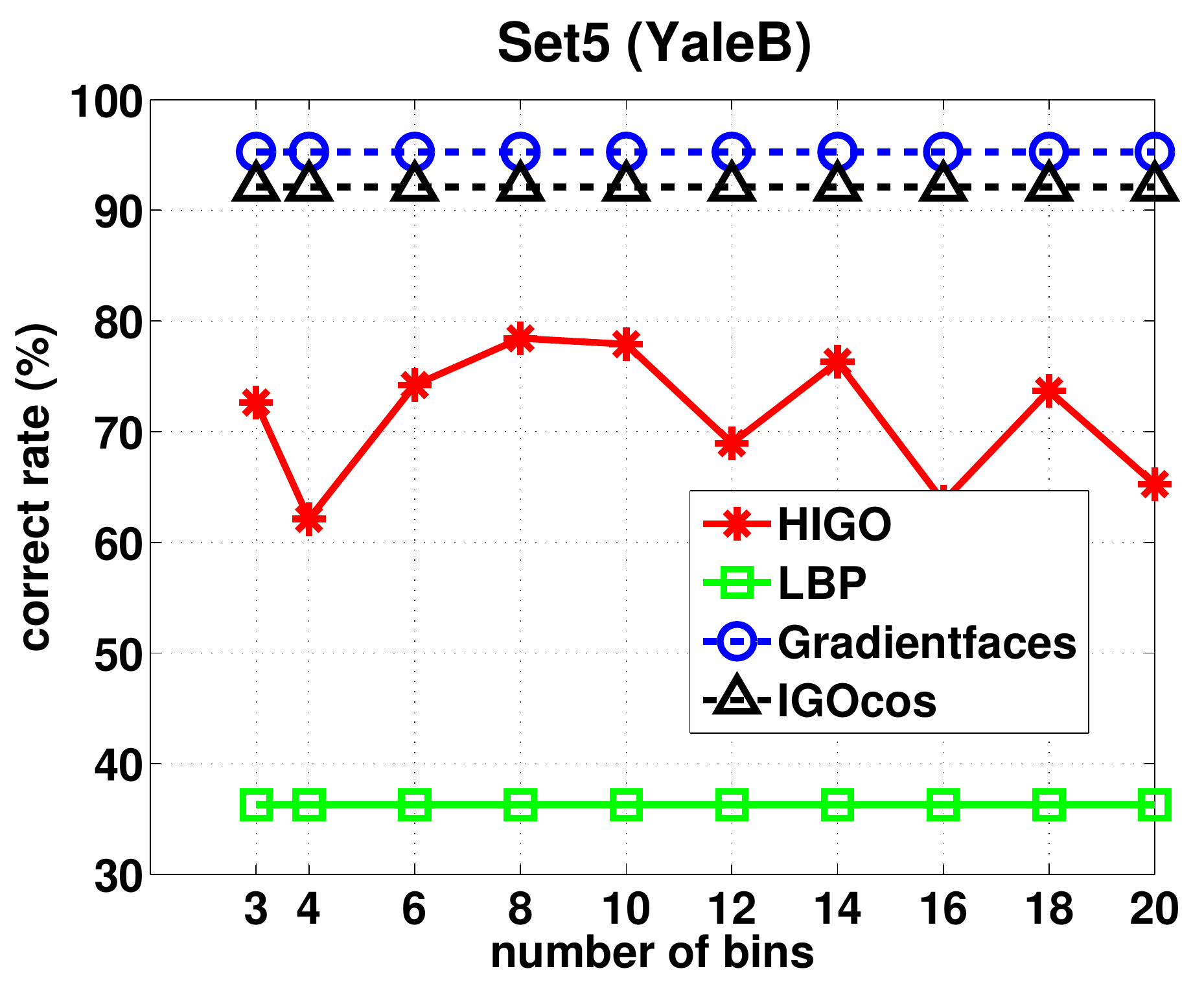}}
\caption{Performance of Gradientfaces, $IGO_{cos}$, LBP and HIGO against (a) scream, (b) occlusion, (c) and (d) illuminations. Block numbers for LBP and HIGO are the same, $20$ for AR and $24$ for YaleB, where both methods reached stable performance.}\label{fig:EXP_pre}
\end{figure*}
This section begins with a discussion of histogram statistics of IGO, followed by introduction of the proposed SBGP descriptor. SBGP computes image gradients from multiple directions in order to extract a set of binary numbers for describing local structures.

With the excellent properties of the IGO representation, we naturally consider to extract robust facial features from the IGO domain, and at the same time, to improve its robustness against local distortions by enforcing block-level locality. A straightforward approach is to directly compute histogram statistics of the IGO representation (HIGO) by dividing a face image into a number of non-overlapped blocks. Each block is represented by an IGO histogram, whose bin number is determined by the segmentations between $[0, 2\pi)$.

 To illustrate the advantage of HIGO, two simple experiments were conducted on two databases. On the AR database, 100 subjects with the group of natural faces were used as gallery images, and two groups of faces with scream expressions and scarf occlusion (both cause large-scale local distortions) were presented as probe images. Each group included 100 images from the first session. On the YaleB database, a subset of 10 subjects was used. The faces with the most natural light sources were used as galleries and two sets with medium and high illumination conditions (corresponding to sets 4 and 5 in\cite{Georghiades2001}) were presented as the probes. We evaluated the IGO based representations, such as Gradientfaces\cite{Zhang2009b} and $IGO_{cos}$ \cite{Tzimiropoulos2012}, local histogram methods, e.g. LBP\cite{Ahonen2006} and HIGO (with respect to bin numbers). The results are illustrated in Fig.~\ref{fig:EXP_pre} (systematical evaluations are reported in Section 6). The experiments here aim to show the experimental cue that motivated the derivation of the proposed descriptor. Fig.~\ref{fig:EXP_pre} evidently provides two observations. First, local features seem more robust to local deformations in expression and occlusion. Second, IGO methods are more capable than LBP in dealing with illumination changes. HIGO, taking advantages of both, achieves stronger robustness to these effects.

 Our goal was to develop a descriptor that can effectively integrate the advantages of both approaches, while still being computationally efficient. For this consideration, one has to trade off between complexity and discrimination with acceptable loss of information. Indeed, as can be seen that HIGO often yields reasonable performance by using fewer bins (e.g. four). Accordingly, it seems appropriate to build a feature model based on the four-bin IGO histograms. Some insights are discussed next.

 \subsection{Theoretical Analysis of Four-Bin HIGO}
 IGO is computed by a four-quadrant inverse tangent, which can be formulated as,
\begin{eqnarray}
\Theta_{x,y} &=&\arctan2 \left(\frac{sign(G_{x,y}^y)}{sign(G_{x,y}^x)}\frac{|G_{x,y}^y|}{|G_{x,y}^x|}\right)\label{eq:IGO2}
\end{eqnarray}
where $sign(G_{x,y}^y)$ and $sign(G_{x,y}^x)$ return the signs of the gradients in the vertical and horizontal directions. $|G_{x,y}^y|$ and $|G_{x,y}^x|$ are the gradient contrasts. In a four-bin histogram, each bin accounts the number of pixels whose IGO values located in one of four quadrants, e.g. $\theta_{x,y} \in [0, \pi/2)$. Hence IGO values are quantified into four discrete values as $\{0, \pi/2, \pi, 3\pi/2\}$, by discarding gradient contrast,
\begin{eqnarray}
\hat{\Theta}_{x,y} &=&\arctan2 \left( \frac{sign(G_{x,y}^y)}{sign(G_{x,y}^x)}\right)\label{eq:IGO4bins}
\end{eqnarray}

 In a four-bin HIGO, pattern labels can be directly computed by four different combinations of two gradient signs as [+ +], [+ -], [- +] and [- -], which are naturally applicable to  binary strategy. Similar to LBP\cite{Ojala2002}, the signs of the gradients are not affected by the changes of mean intensities, yielding a distinct ability to resist gray-scale variations. Subsequently, we generate two binary numbers to describe the patterns of four-bin HIGO as, 11, 10, 01 and 00. While LBP discards intensity contrast, two-bit HIGO discards gradient contrast to achieve illumination invariance as well as computational efficiency. To this end, we have derived the basic local binary features from the IGO domain, which serve as the basis of the proposed SBGP descriptor.

\subsection{Binary Image Gradients from Multiple Directions}
\begin{figure}
    \begin{center}
     \includegraphics[height=2.5cm,width=8.5cm]{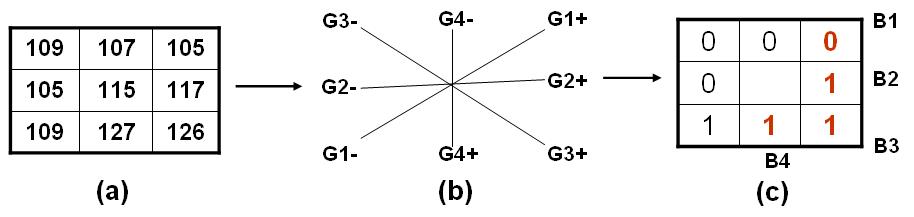}
    \end{center}
\caption{Basic SBGP operator: (a) eight neighbors of a central pixel (115), (b) four correlation directions: G1, G2, G3 and G4, (c) principal (in red or bold) and associated (in black or plain) binary numbers, resulting string, $0111_2$, or label, $L=07$.}\label{fig:SBGP_basic}
\end{figure}

The traditional IGO  is computed on gradients of horizontal and vertical directions. Its pixel-level locality is realized by using only four neighbors in two orthogonal directions.  Most current high-performing local descriptors extract meaningful local information from at least eight neighbors and their discriminative power can be improved by suitably increasing the number of neighbors \cite{Ojala1996,Ojala2002,Liu2002,Chen2010,Lei2011,Tan2011, Kumar2012}. Similarly, it can be expected that greater discrimination can be achieved in the gradient domain by involving more local neighbors from multiple directions.

Following this intuition, we further extend the four-bin HIGO to multiple directions, resulting the proposed the new facial descriptor, binary gradient patterns (BGP).  Specifically, the BGP computes binary correlations between symmetric neighbors of a central pixel from multiple ($k$) directions. The number of neighbors is twice of the number of directions. The computation is simple. A basic BGP operator of four directions is presented in Fig.~\ref{fig:SBGP_basic} and detailed bellow:

1). A set of local neighbors of a central pixel are first given (e.g. eight neighbors in Fig.~\ref{fig:SBGP_basic}(a)).

2). Then, a pair of binary numbers including a principal binary number ($B_i^+$) and an associated binary number ($B_i^-$), are computed by correlating two symmetric neighbors in each direction based on Eq.~(\ref{eq:binary}), and totally eight binary numbers are devised from four directions: G1, G2, G3 and G4, shown in Fig.~\ref{fig:SBGP_basic}(b) and (c);
\begin{eqnarray}\label{eq:binary}
B_i^+=
\begin{cases}
1 & \text{ if } G_i^+-G_i^-\geqslant 0 \\
0 & \text{ if } G_i^+-G_i^-<0
\end{cases}\\ \nonumber
B_i^-=1-B_i^+ \quad i=1,2,\ldots,k
\end{eqnarray}
where $G_i^{+}$ and $G_i^{-}$ are the intensity values of the pixels corresponding to locations in Fig.~\ref{fig:SBGP_basic}(b).

3). Finally, label of the central pixel is computed from the resulting four principal binary numbers,
\begin{eqnarray}\label{eq:label}
L=\sum_{i=1}^k 2^{i-1}B_i^+
\end{eqnarray}

Although eight binary numbers are obtained in four directions, the principal and associated binary numbers in each direction are always complementary. Hence, there are only two variances in each direction, which only require a single binary number/bit to describe. For a compact representation, only the principal binary numbers are required for computing the labels by Eq.~(\ref{eq:label}), capable of describing all possible variances of the BGP patterns. The number of BGP labels $(N_L)$ is determined by the number of the principal binary numbers/bits, equal to the number of directions ($k$), $N_L=2^k$. Thus the possible labels of a $k$-directional SBGP operator are $L^k \in \{0, 1, 2, \ldots, 2^{k-1}\}$. Note that this number ($2^k$) is substantially smaller than the number of LBP labels ($2^{2k}$). For example, in a typical model with sixteen neighbors, the numbers of labels for BGP and LBP are 256 and 65536, respectively. The proposed BGP operator efficiently integrates the merits of IGO features and local histogram representations with extremely low computational cost.


\subsection{Structural BGP}
\begin{figure*}
    \begin{center}
     \subfigure[Structural (in red boxes) and non-structural patterns (in black boxes)]{\includegraphics[height=3.6cm,width=14cm]{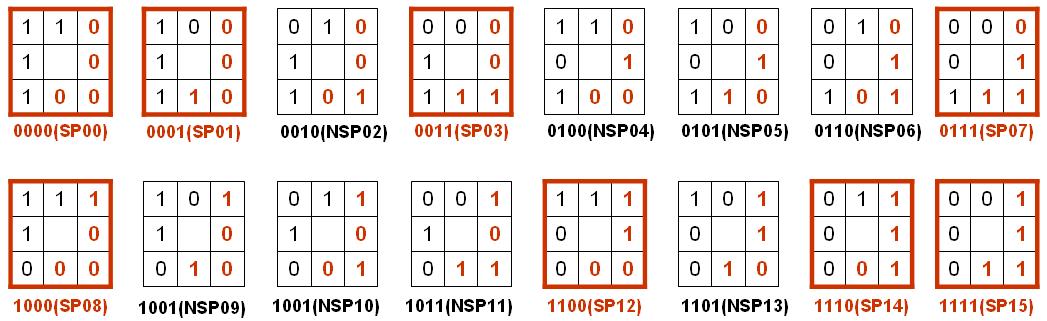}}
     \subfigure[\emph{struct.} patt.]{\includegraphics[height=3.6cm,width=3.6cm]{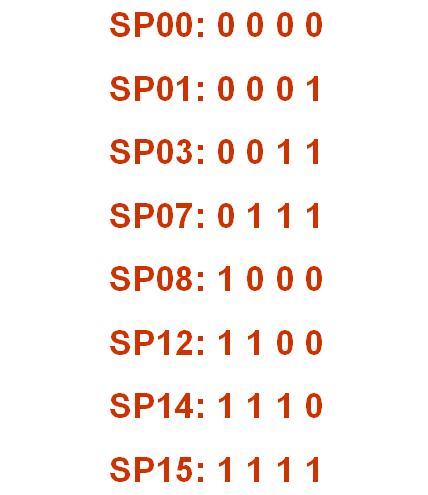}}
    \end{center}
\caption{Definition of BGP \emph{structural} and \emph{non-structural} patterns.}\label{fig:SBGP_UP}
\end{figure*}

\begin{figure}
    \begin{center}
     \subfigure[]{\includegraphics[height=2.8cm, width=4cm]{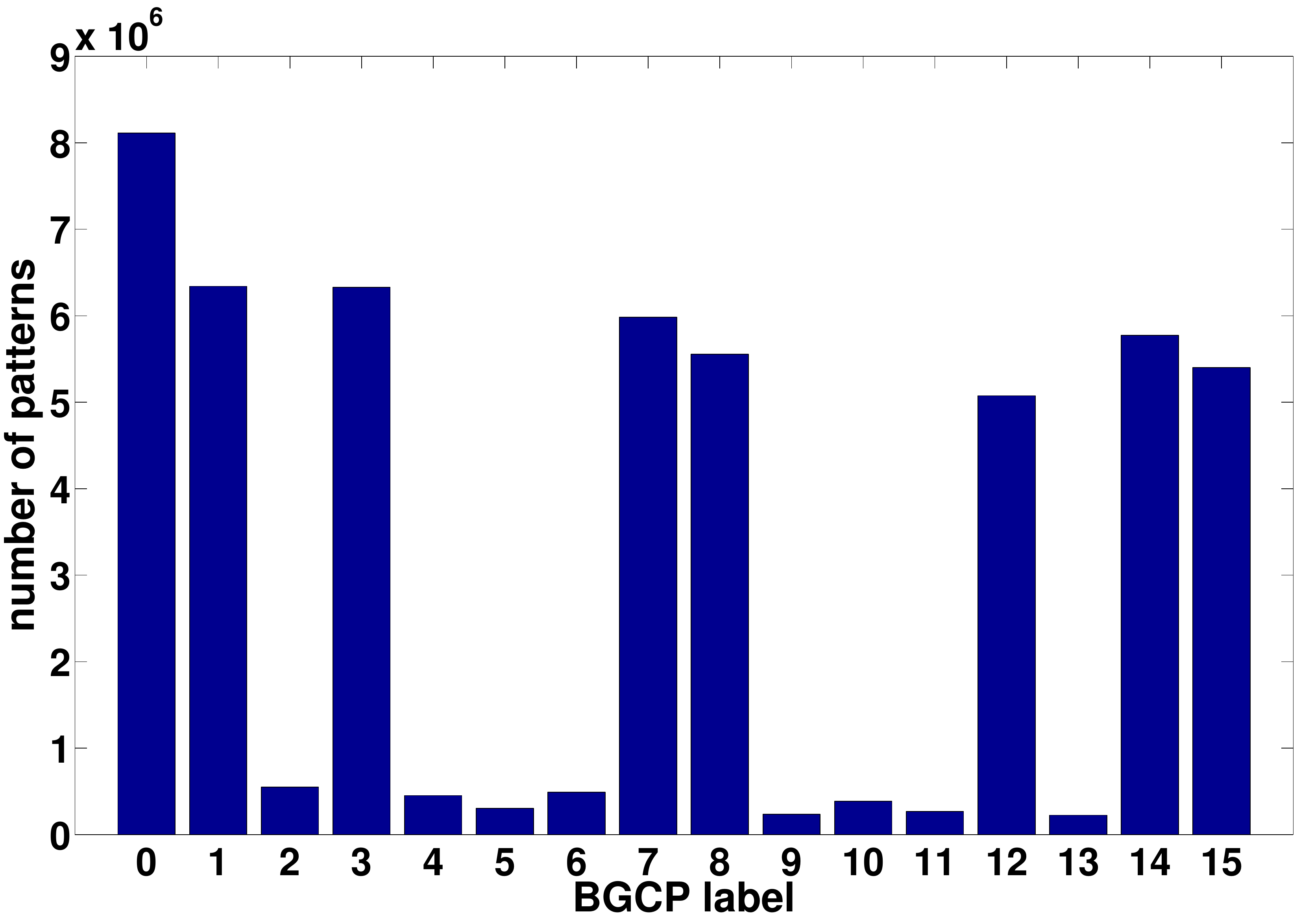}}
     \subfigure[]{\includegraphics[height=2.8cm, width=4cm]{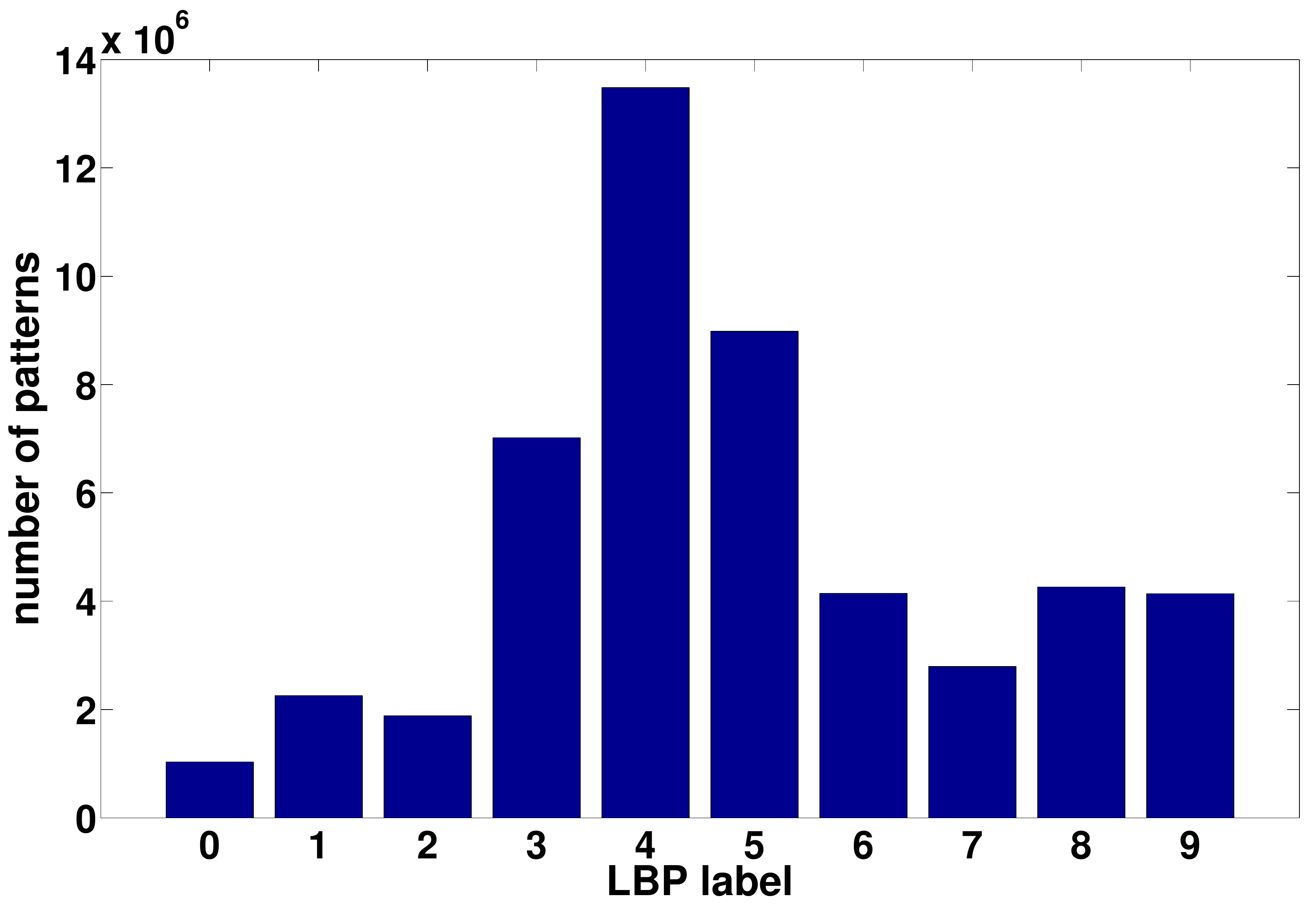}}
    \end{center}
\caption{Histogram statistics of BGP and LBP patterns on AR database, (a) BGP patterns, $x$-axis corresponds to 16 different labels presented in Fig.~\ref{fig:SBGP_UP}, (b) $LBP_{8,1}^{riu2}$ patterns, labels 0-8 and 9 are  \emph{uniform} and \emph{non-uniform} patterns, respectively.}\label{fig:histSBGP_vs_histLBP}
\end{figure}

There are sixteen different labels from a four-directional BGP descriptor. The binary structures of these labels, ranging from 0 to 15, are shown in Fig.~\ref{fig:SBGP_UP}(a). As can be seen, each label is constructed by eight binary numbers/bits, including four bits of "1" and four bits of "0". The principal bits are presented in red or bold. It is interesting to investigate the distributions of "1"s and "0"s in different labels. It can be seen that certain labels have meaningful structures where four bits of "1" are located consecutively. There are eight labels having continuous bits of "1" (marked as red or bold-lined boxes in Fig.~\ref{fig:SBGP_UP}(a)); while the "1"s in the other eight labels are discontinuous (marked as black or thin-lined boxes). These continuous "1"s indicate more stable local changes in texture and essentially describe the orientations of "edge" texture. An observation is that statistics on these patterns is highly stable and meaningful to characterize local structures. By contrast, labels with discontinuous "1"s include arbitrary changes of local texture, likely to indicate noise or outliers. Furthermore, from experimental statistics, patterns having the continuous "1"s often take up a vast majority in a typical BGP face, e.g. about 95
percent. The statistics of BGP patterns of various labels on 2600 face images from the AR databases is presented in Fig.~\ref{fig:histSBGP_vs_histLBP}(a).

Based on these observations, we define the patterns having continuous "1"s as \emph{structural} binary gradient patterns (SBGP), while refer the others as \emph{non-structural} patterns. This yields in total eight different labels for the \emph{structural} patterns (as listed in Fig.~\ref{fig:SBGP_UP}(b)) while discarding all \emph{non-structural} ones. Therefore, only eight bins are needed in the SBGP histogram. This is an appealing property, not only helping to rule out noise and outliers in face images, but also further reducing feature dimensions. For example, even with 24 neighbors, the bin number of the SBGP histogram is only 24, compared the $2^{12}=4096$ bins of the CS-LBP histogram \cite{Heikkila2009}. Further discussions and evaluations are presented in the next section.

\subsection{Spatial Resolutions}


The basic SBGP descriptor (Fig.~\ref{fig:SBGP_basic}) is computed from four directions ($k=4$) in a square neighborhood of side length of two units. Similar to LBP based descriptors, the capability of SBGP can be further improved by increasing the number of gradient directions and by enlarging the neighborhood. To this end, we define the spatial resolution of SBGP by the number of neigbors/directions and radius of the square, indicated as $(P,R)$. Typically, the maximum number of neighbors is eight times of the radius, $P_{max}=8R$, e.g. ($8,1$), ($16,2$) and ($24,3$). The SBGP descriptor with \emph{structural} patterns in spatial resolution of ($P,R$) is referred as $SBGP_{P,R}$. Assuming that the number of neighbors is maximized with respect to the radius, we present a generalized algorithm for computing the SBGP operator from a given pixel in location $(i,j)$, with spatial resolution of $(P,R)$, see Algorithm~\ref{alg_SBGP} for details.

Algorithm~\ref{alg_SBGP} returns the label values of all pixels. In this framework, features are built on histograms of the SBGP \emph{structural} patterns. One needs to know the number of the \emph{structural} labels and their values. This information is independent to face images, and is only determined by the given spatial resolution. From Fig.~\ref{fig:SBGP_UP}(a), we can find that four continuous "1"s in eight \emph{structural} labels  run through all locations of eight neighbors, indicating that the number of \emph{structural} labels ($N_{sp}$) is equal to the number of neighbors, $N_{sp}=P$, compared to $2^P$ of the LBP and $2^{\frac{P}{2}}$ of the CS-LBP \cite{Heikkila2009}. Also, based on the distributions of principal bits of the \emph{structural} labels (Fig.~\ref{fig:SBGP_UP}(b)), we device Algorithm~\ref{alg_SP} for computing \emph{structural} labels at resolution of ($P,R$).

\begin{algorithm}[H]
\small
  \caption{Computing SBGP descriptor}
  \label{alg_SBGP}
  \begin{spacing}{0.6}
  \begin{algorithmic}[1]
    \REQUIRE Location of a given pixel $(i,j)$, spatial resolution, $(P,R)$ and $I_{(i,j)}$, pixel intensity.

    \ENSURE Label of SBGP descriptor, $L_{(i,j)}$.
    \STATE \textbf{step one}: compute principal binary numbers ($B_t^+$) in $k$ directions, $k=P/2$.
    \STATE $t=1$
    \FOR{$n_1 = -R \to R$}
\STATE   \begin{eqnarray}\label{eq:SBGP_binary1}
          B_t^+=
                \begin{cases}
                1 & \text{ if } I_{(i+n_1,j+R)}-I_{(i-n_1,j-R)}\geqslant 0 \\
                0 & \text{ if } I_{(i+n_1,j+R)}-I_{(i-n_1,j-R)}<0
                \end{cases}
          \end{eqnarray}
\STATE $t=t+1$
    \ENDFOR
    \FOR{$n_2 = -(R-1) \to (R-1)$}
    \STATE
    \begin{eqnarray}\label{eq:SBGP_binary2}
    B_t^+=
          \begin{cases}
          1 & \text{ if } I_{(i+R,j-n_2)}-I_{(i-R,j+n_2)}\geqslant 0 \\
          0 & \text{ if } I_{(i+R,j-n_2)}-I_{(i-R,j+n_2)}<0
          \end{cases}
    \end{eqnarray}
    \STATE $t=t+1$
    \ENDFOR
    \STATE \textbf{step two}: compute $L_{(i,j)}$ by Eq.~(\ref{eq:label}).
    \RETURN Label of pattern, $L_{(i,j)}$.
  \end{algorithmic}
  \end{spacing}
\end{algorithm}

\begin{algorithm}[H]
\small

  \caption{Computing \emph{structural} label}
  \label{alg_SP}
   \begin{spacing}{0.7}
  \begin{algorithmic}[1]
    \REQUIRE Number of neighbors, $P$.

    \ENSURE Labels of \emph{structural} patterns, $L^{sp}$.

    \STATE Number of directions, $k=P/2$.
    \FOR{$t = 1 \to P$}

    \IF{$t\leq k$}
 \STATE
$ L_t^{sp}=2^{t-1}-1  $
    \ELSE
   \STATE
    $ L_t^{sp}=2^k-L_{2k-t+1}^{sp}-1 $

    \ENDIF
    \ENDFOR

    \RETURN $\{L_t^{sp}\}_{t=1}^P$.
  \end{algorithmic}
   \end{spacing}
\end{algorithm}

\section{Analysis, Discussions and Comparisons}

\begin{figure*}
\renewcommand{\thesubfigure}{\empty}

\subfigure[face]{\includegraphics[width=1.5cm]{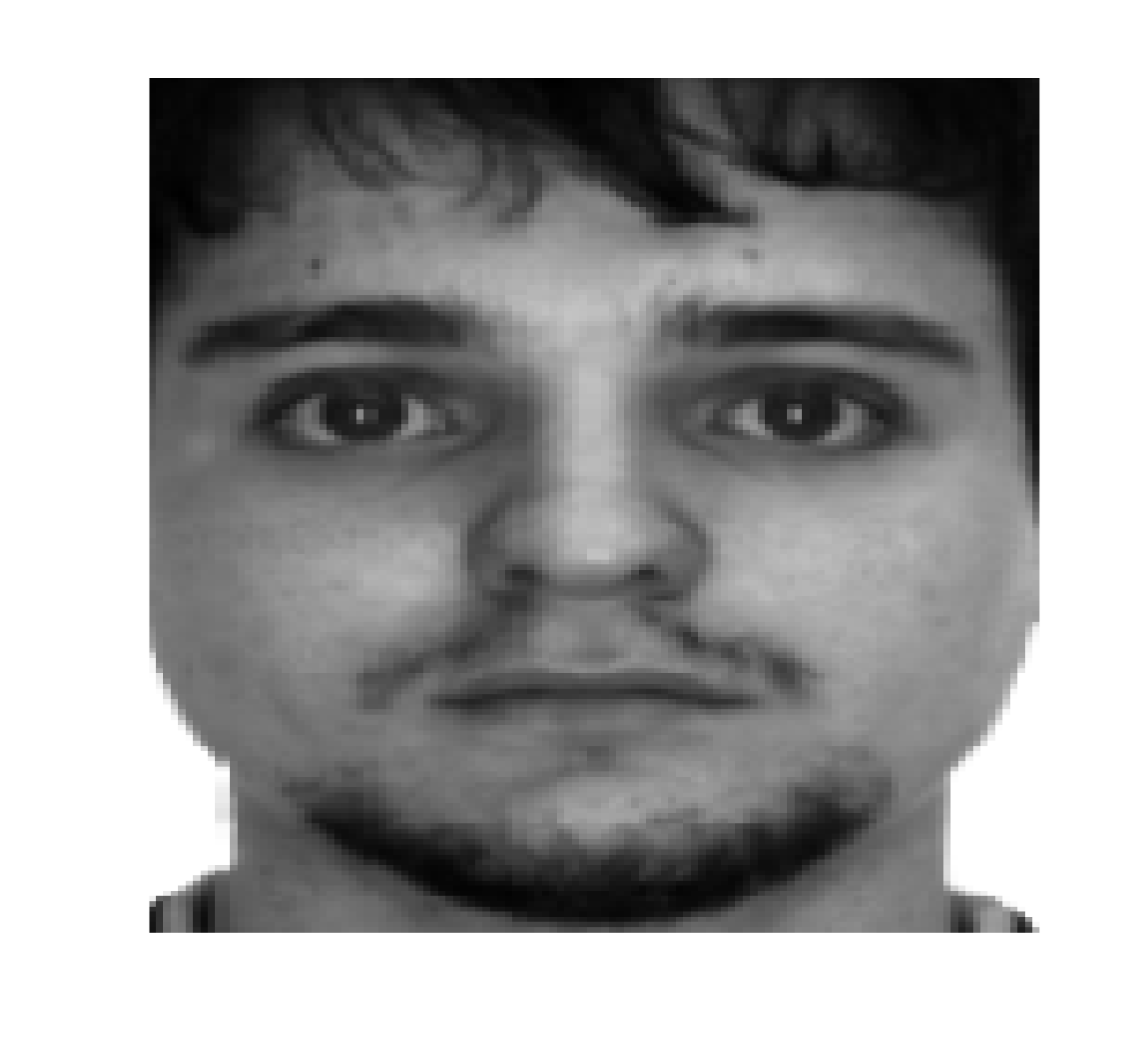}}
\subfigure[$0$]{\includegraphics[width=1.5cm]{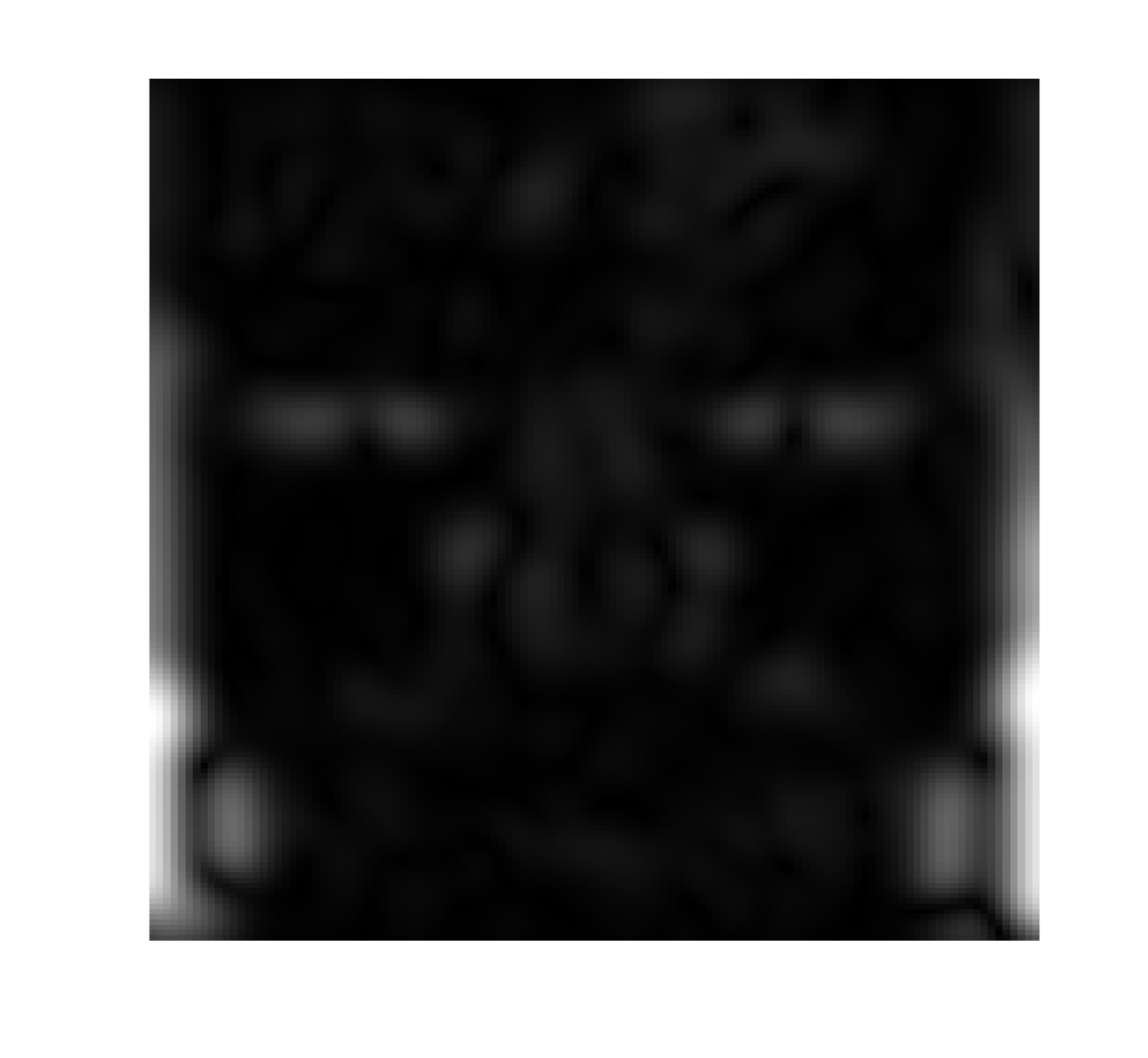}}
\subfigure[$\pi/8$]{\includegraphics[width=1.5cm]{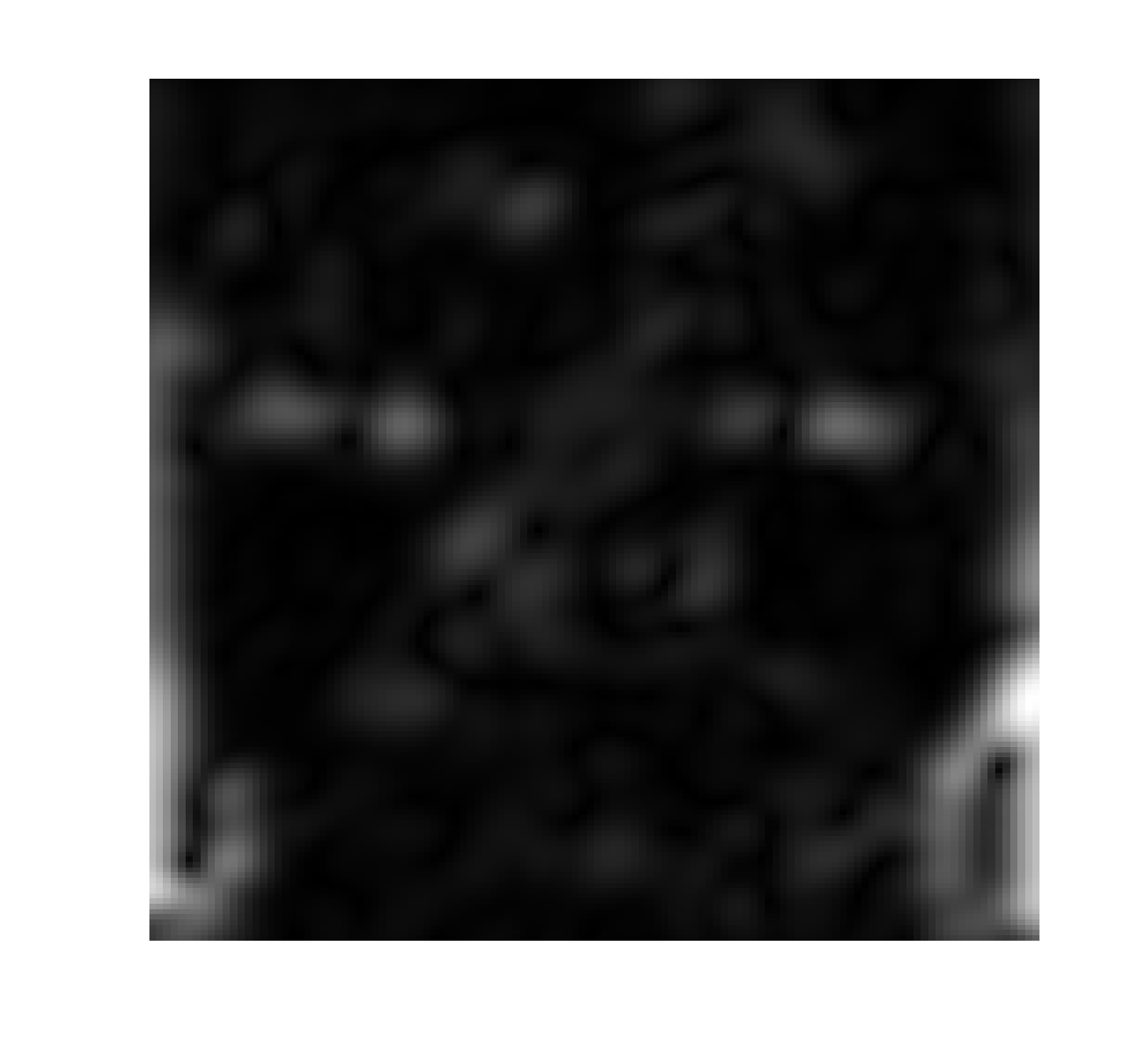}}
\subfigure[$2\pi/8$]{\includegraphics[width=1.5cm]{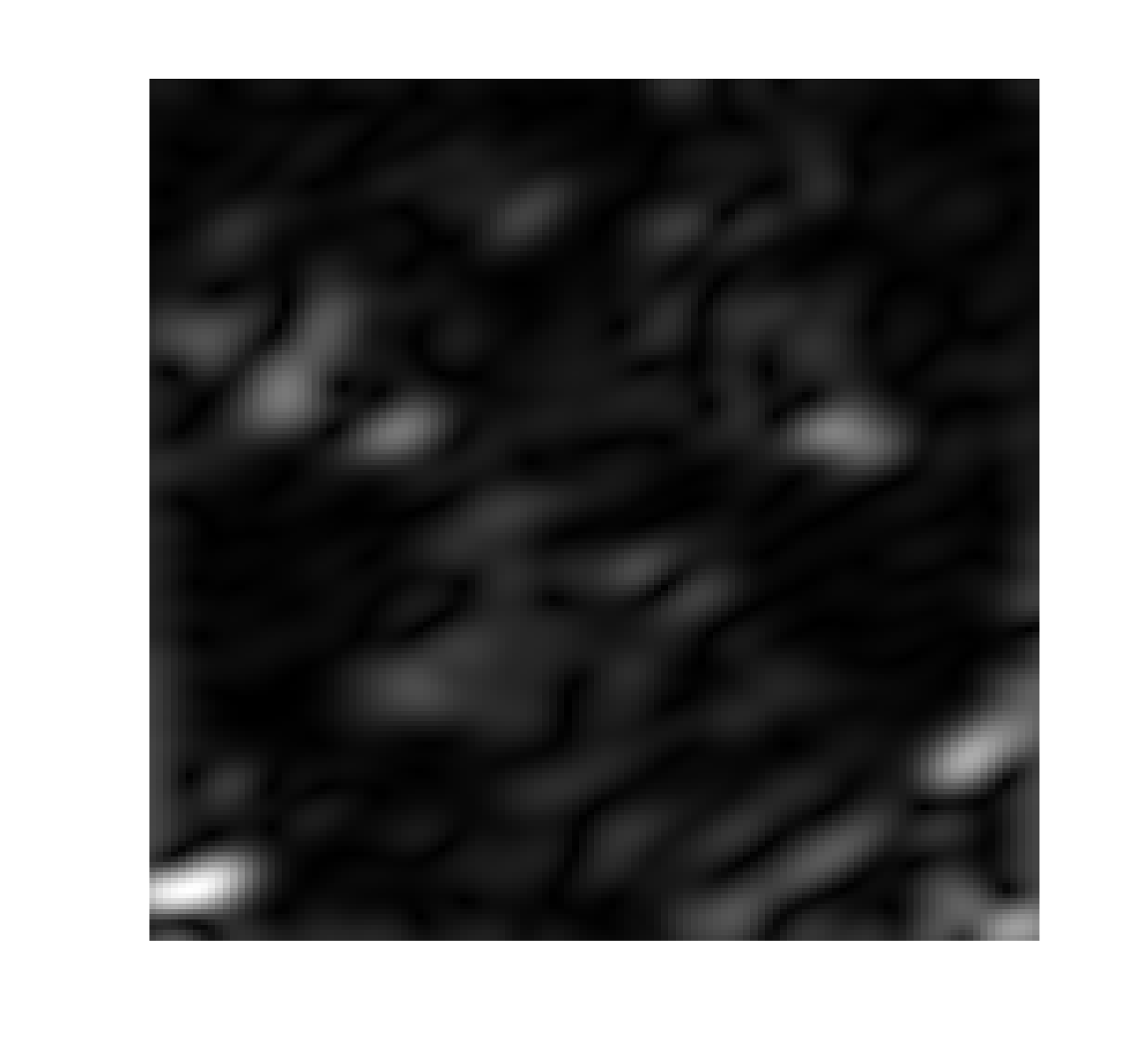}}
\subfigure[$3\pi/8$]{\includegraphics[width=1.5cm]{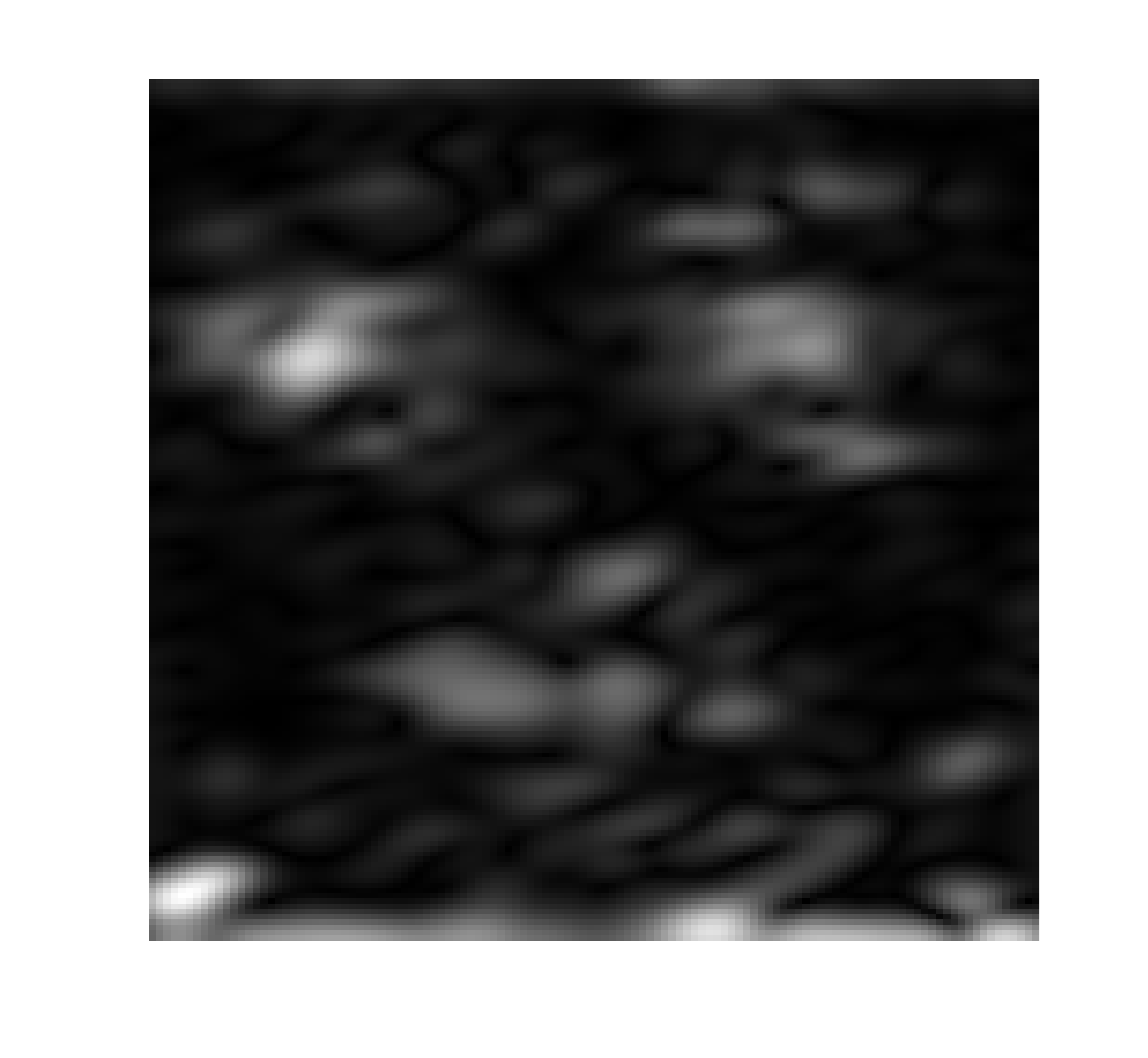}}
\subfigure[$4\pi/8$]{\includegraphics[width=1.5cm]{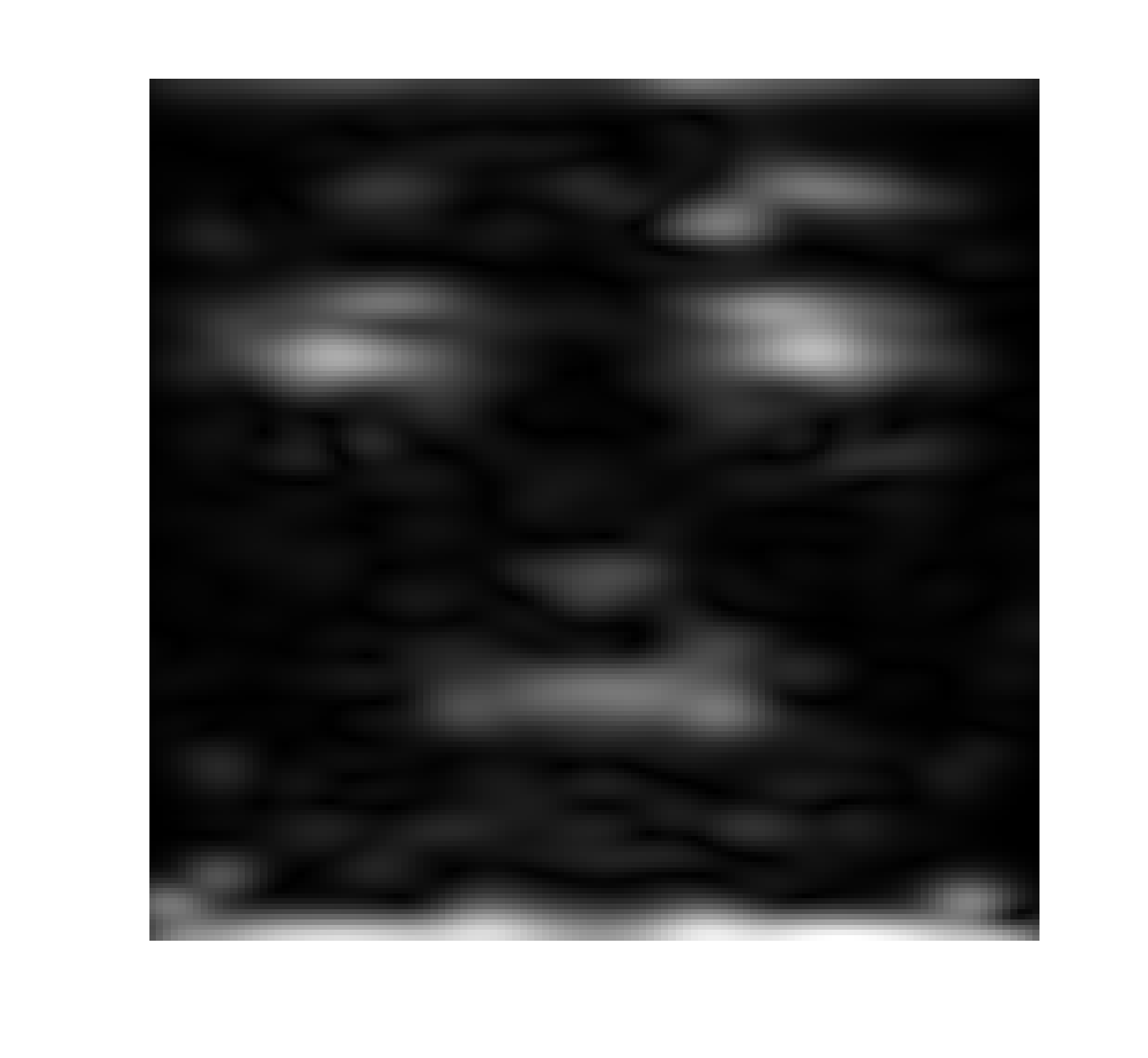}}
\subfigure[$5\pi/8$]{\includegraphics[width=1.5cm]{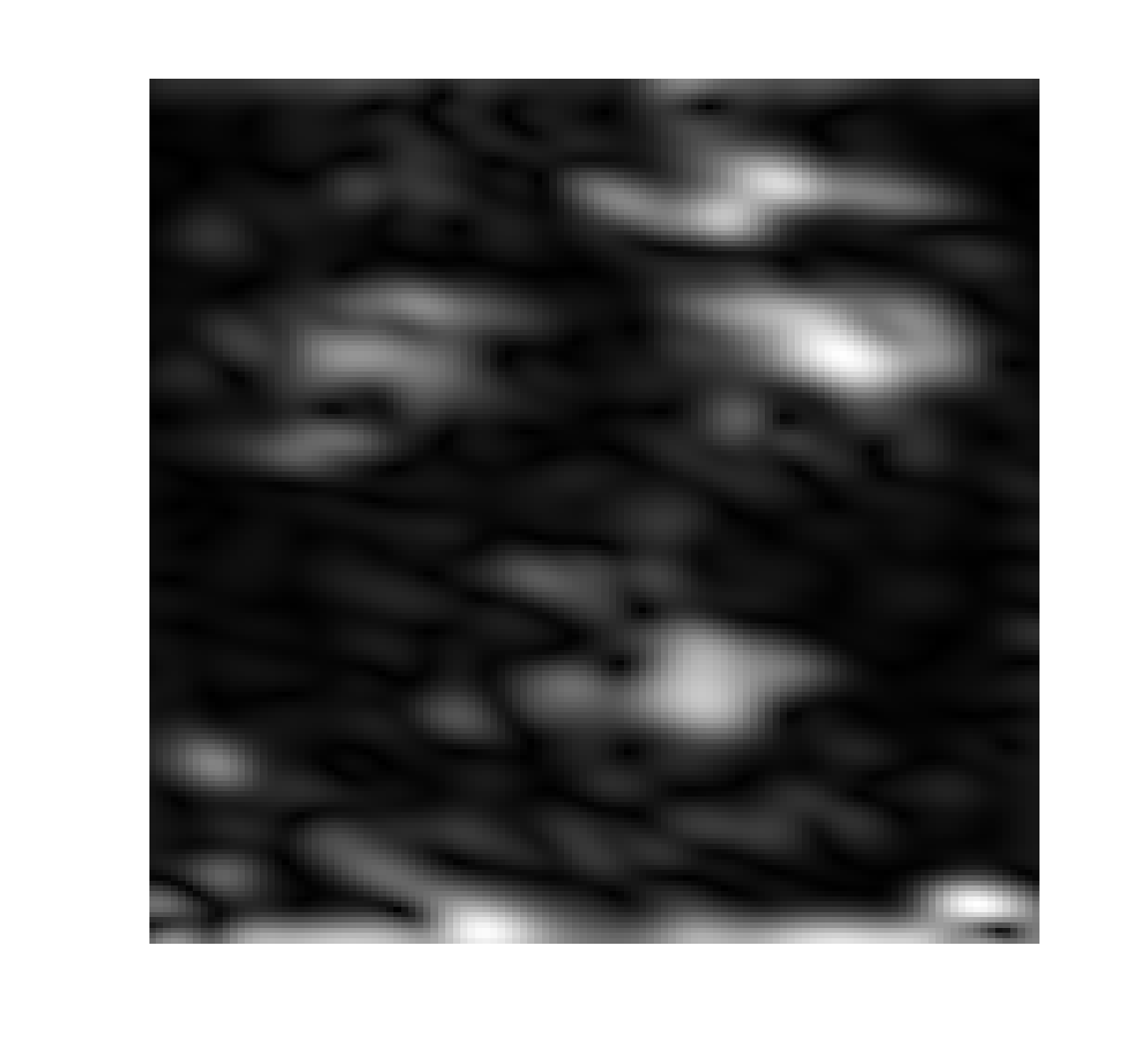}}
\subfigure[$6\pi/8$]{\includegraphics[width=1.5cm]{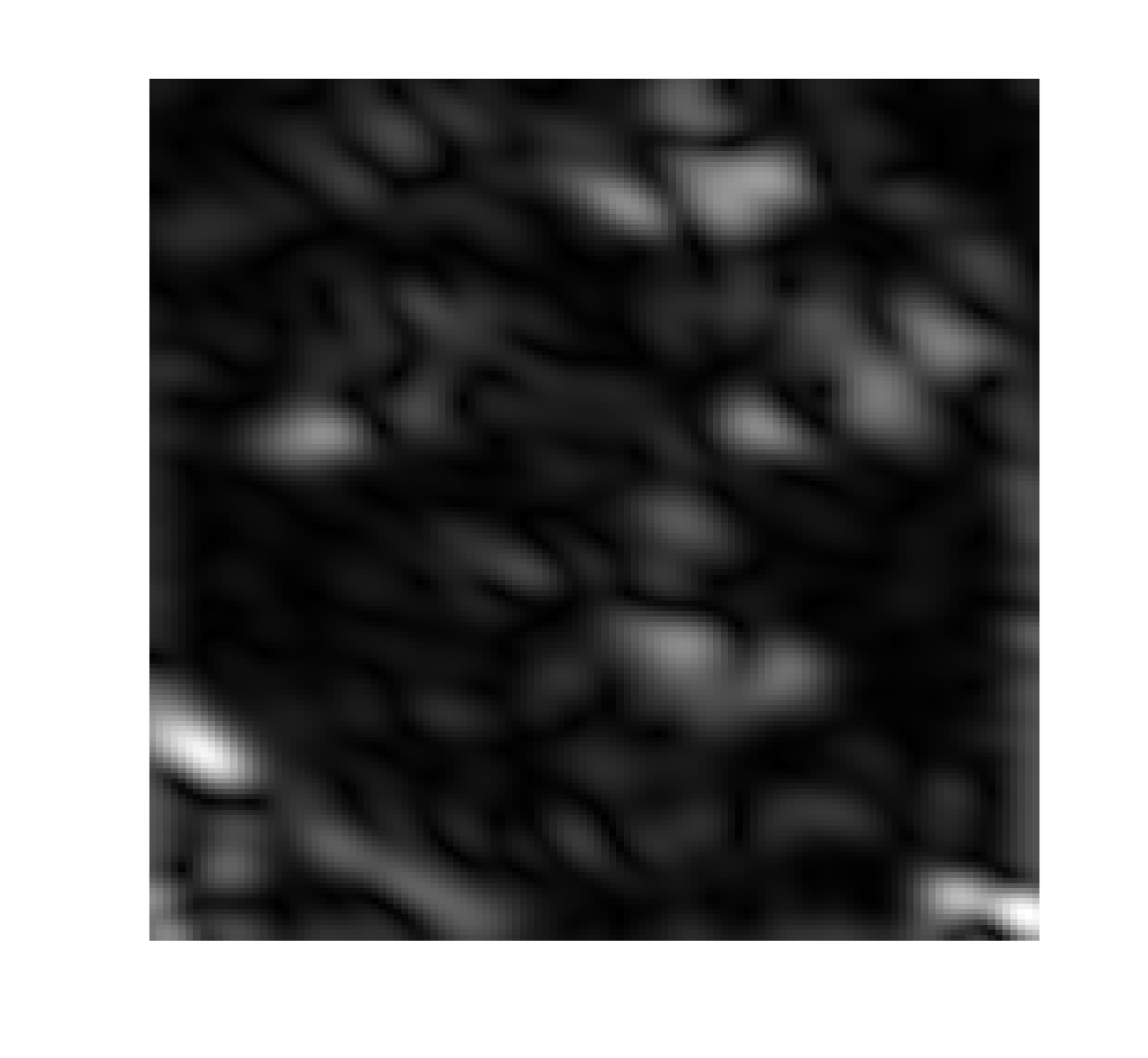}}
\subfigure[$7\pi/8$]{\includegraphics[width=1.5cm]{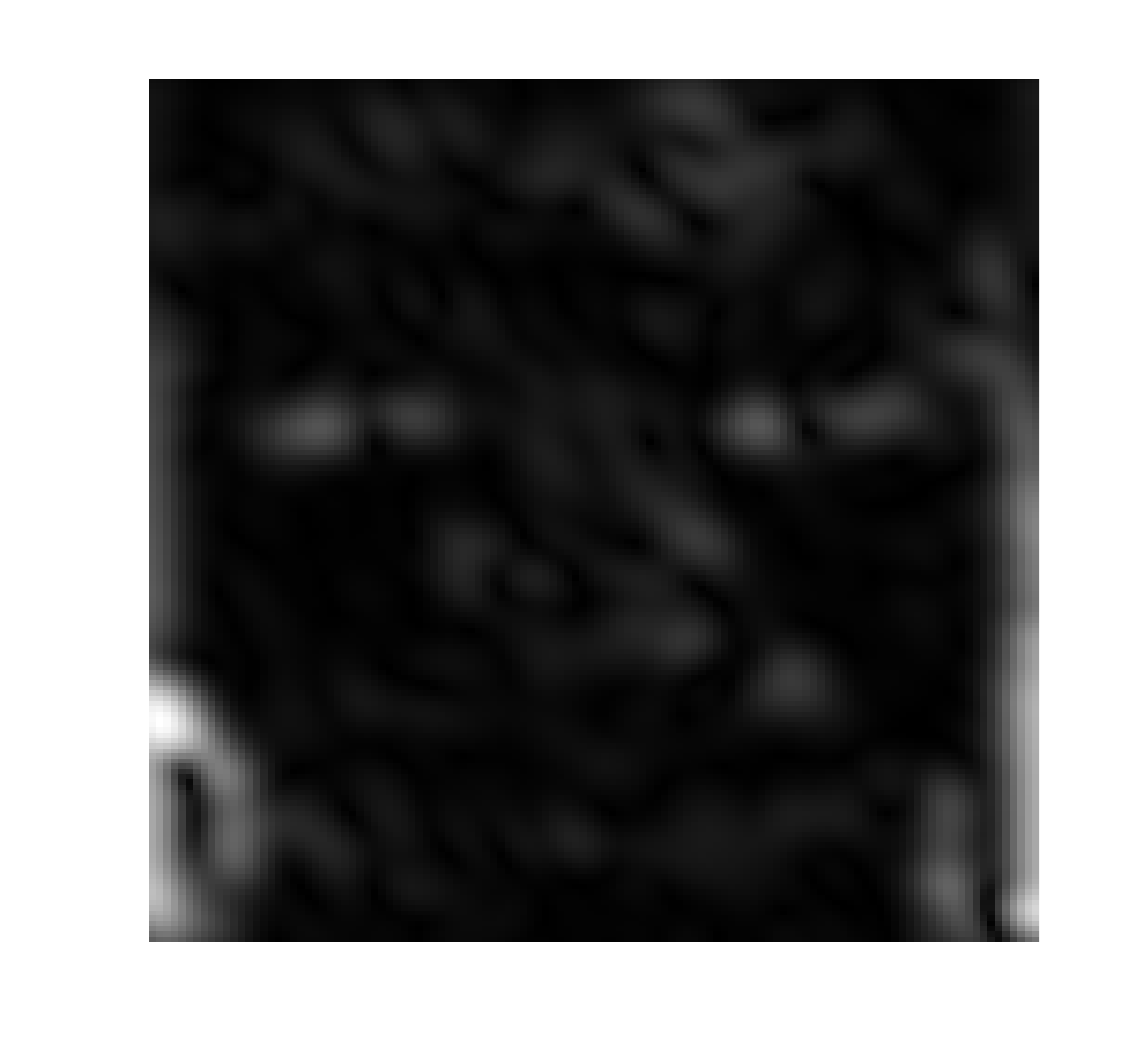}}\\

\subfigure[LBP]{\includegraphics[width=1.5cm]{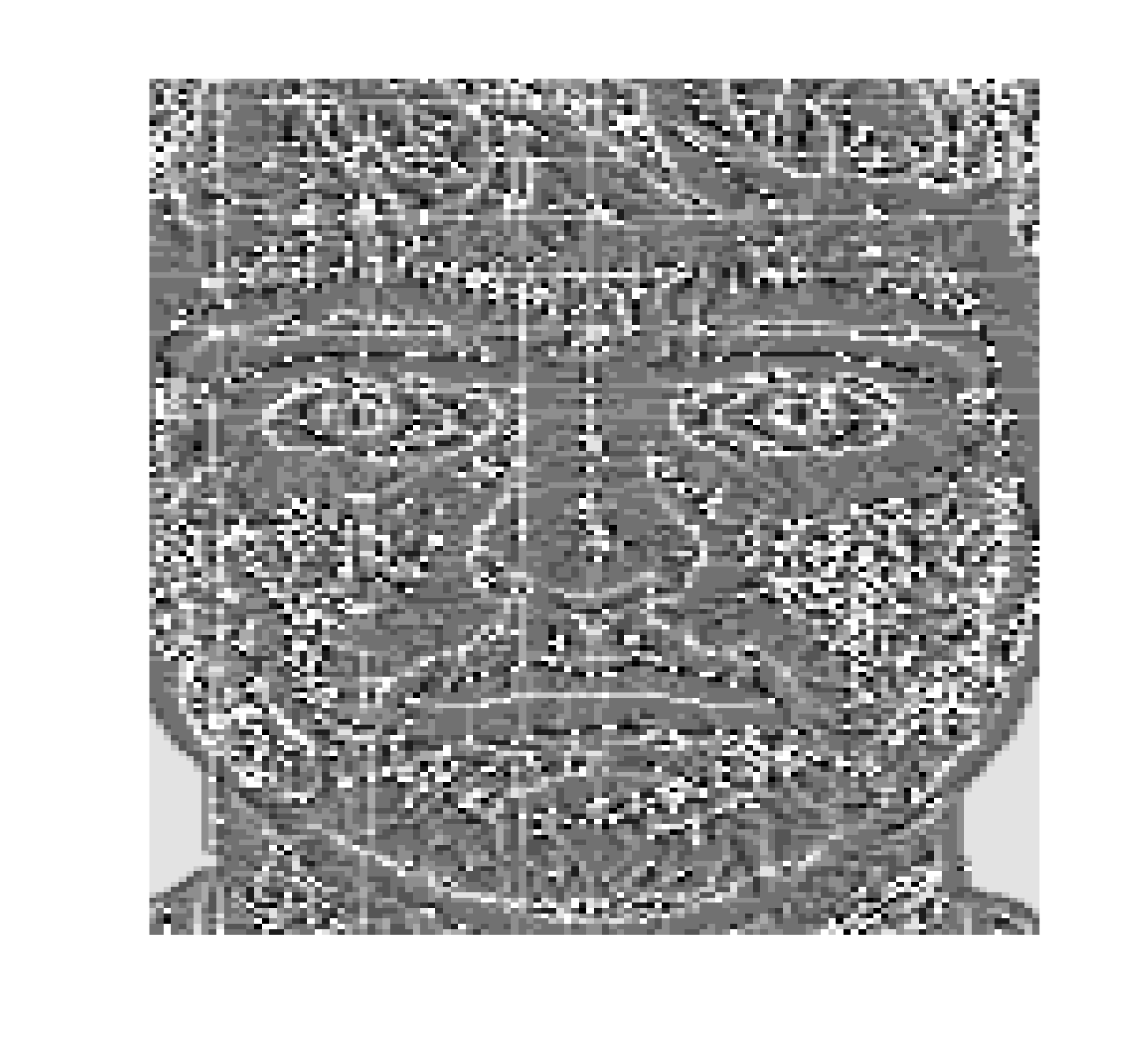}}
\subfigure[UP00]{\includegraphics[width=1.5cm]{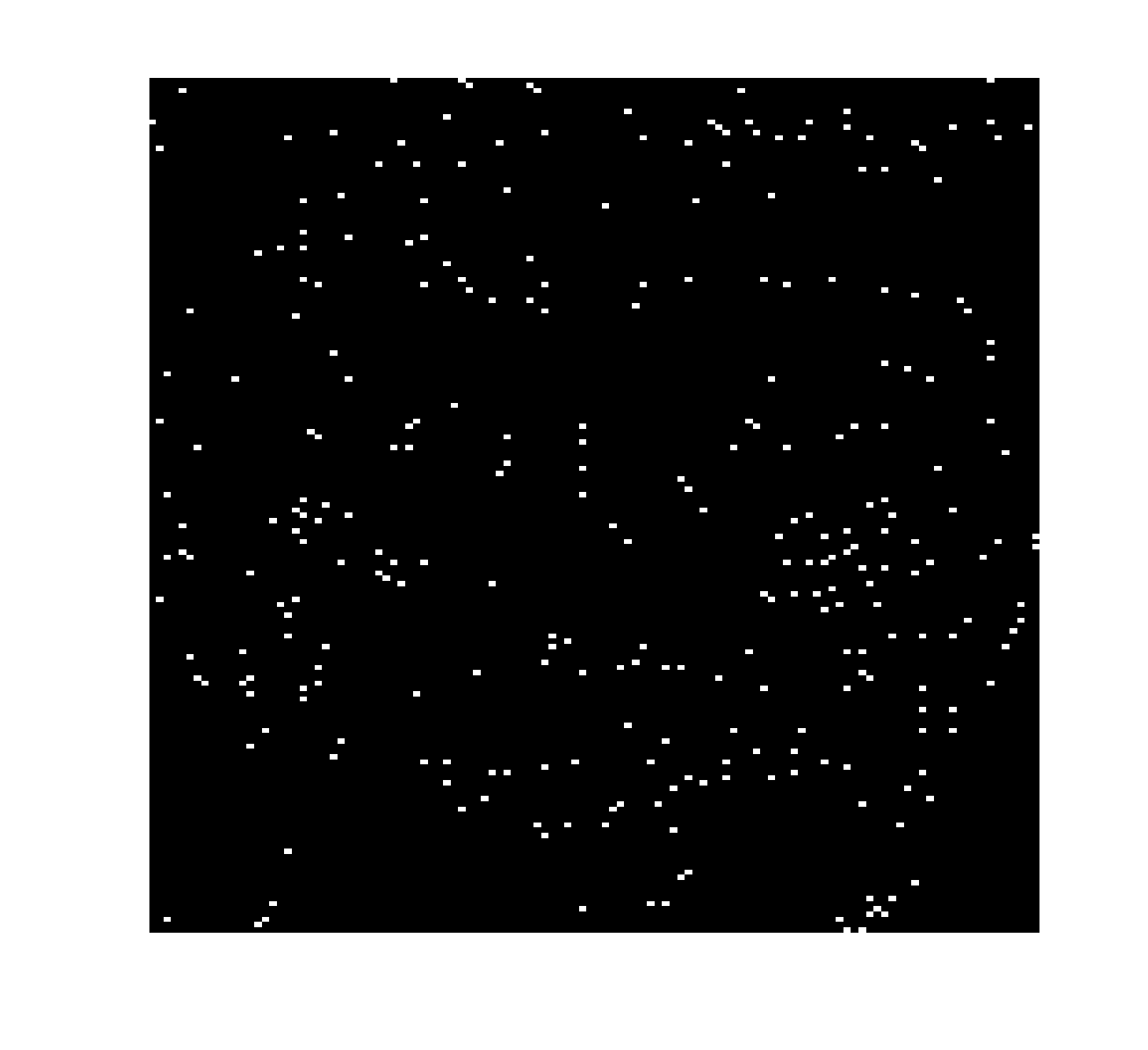}}
\subfigure[UP01]{\includegraphics[width=1.5cm]{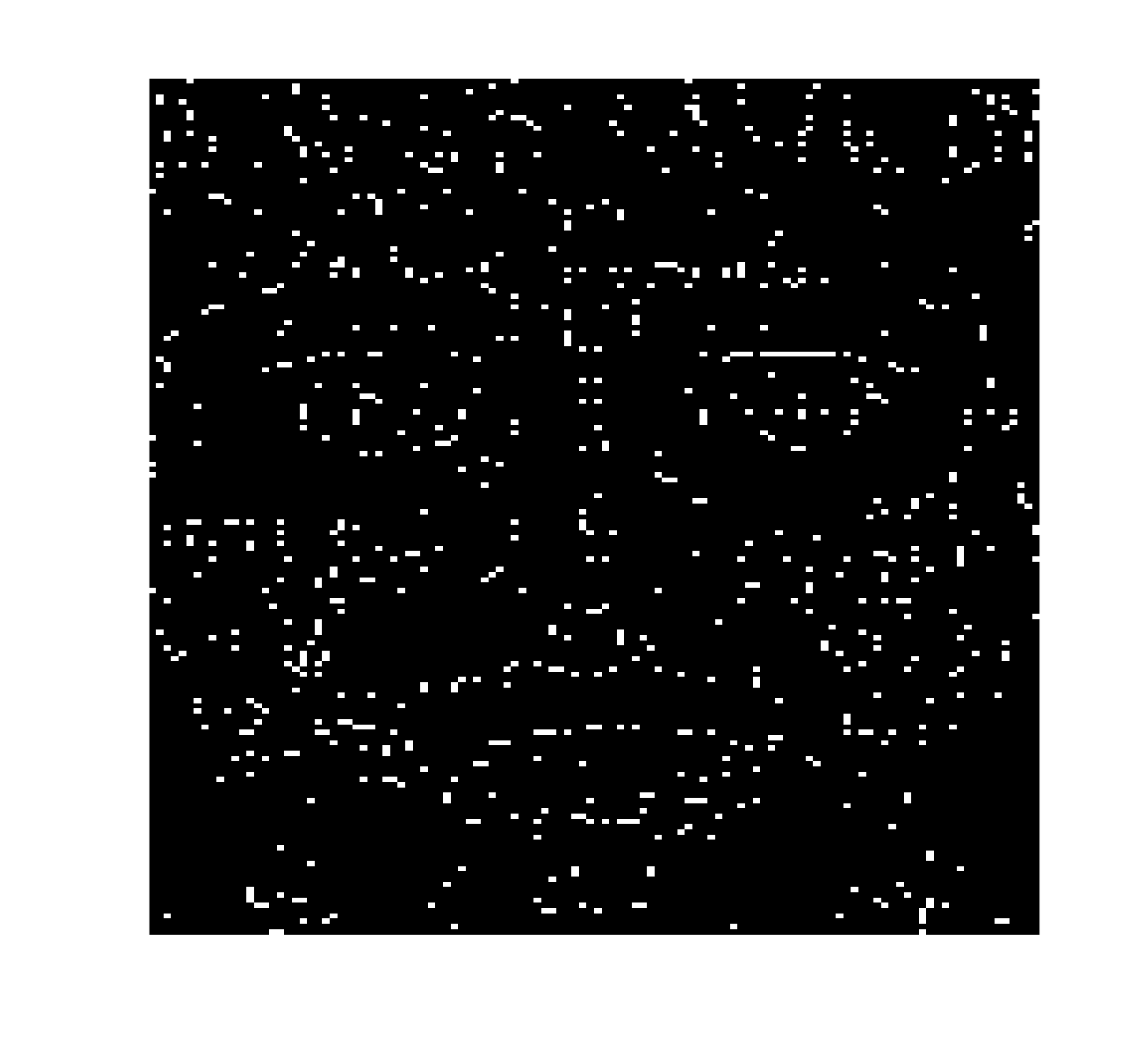}}
\subfigure[UP02]{\includegraphics[width=1.5cm]{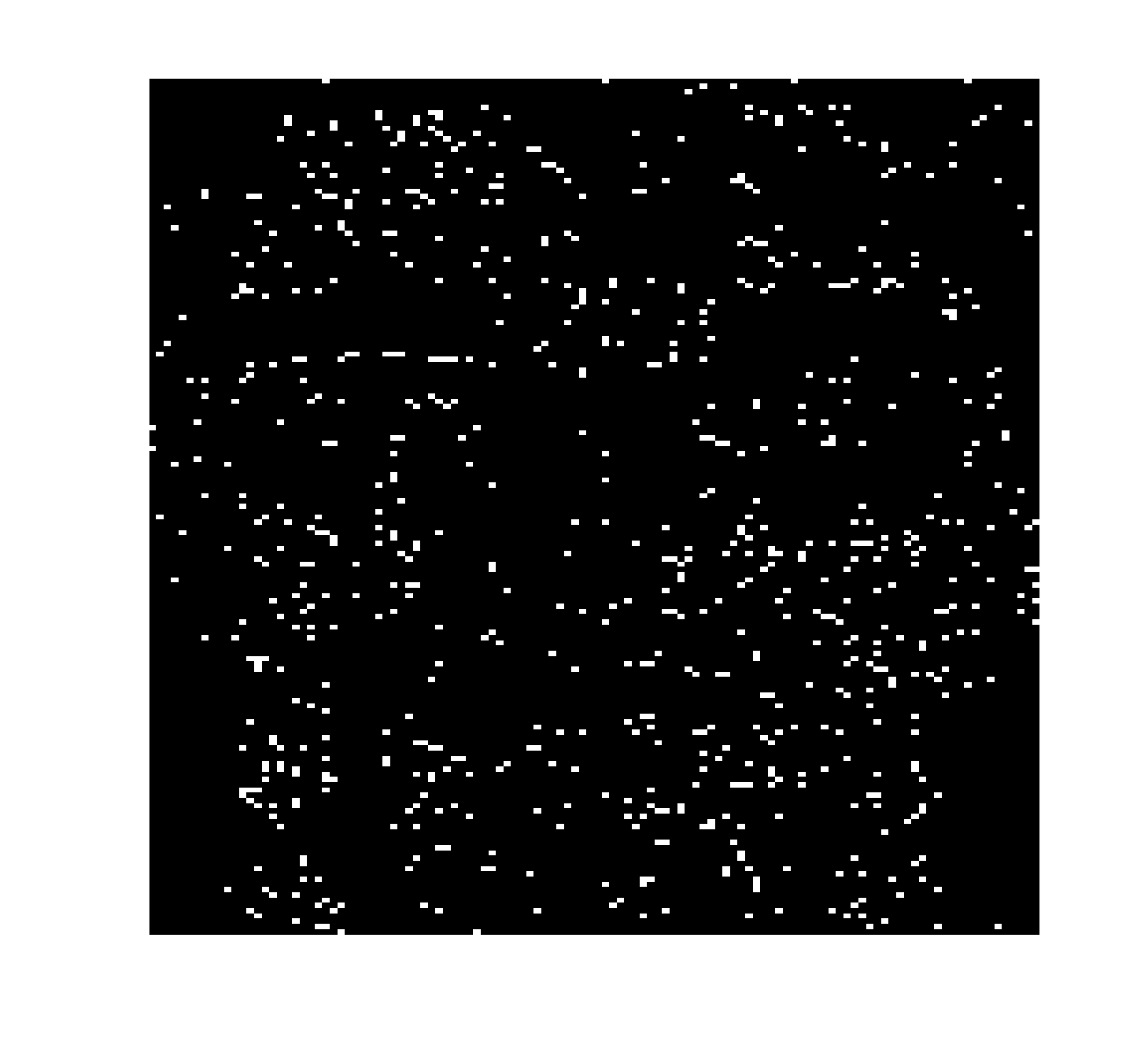}}
\subfigure[UP03]{\includegraphics[width=1.5cm]{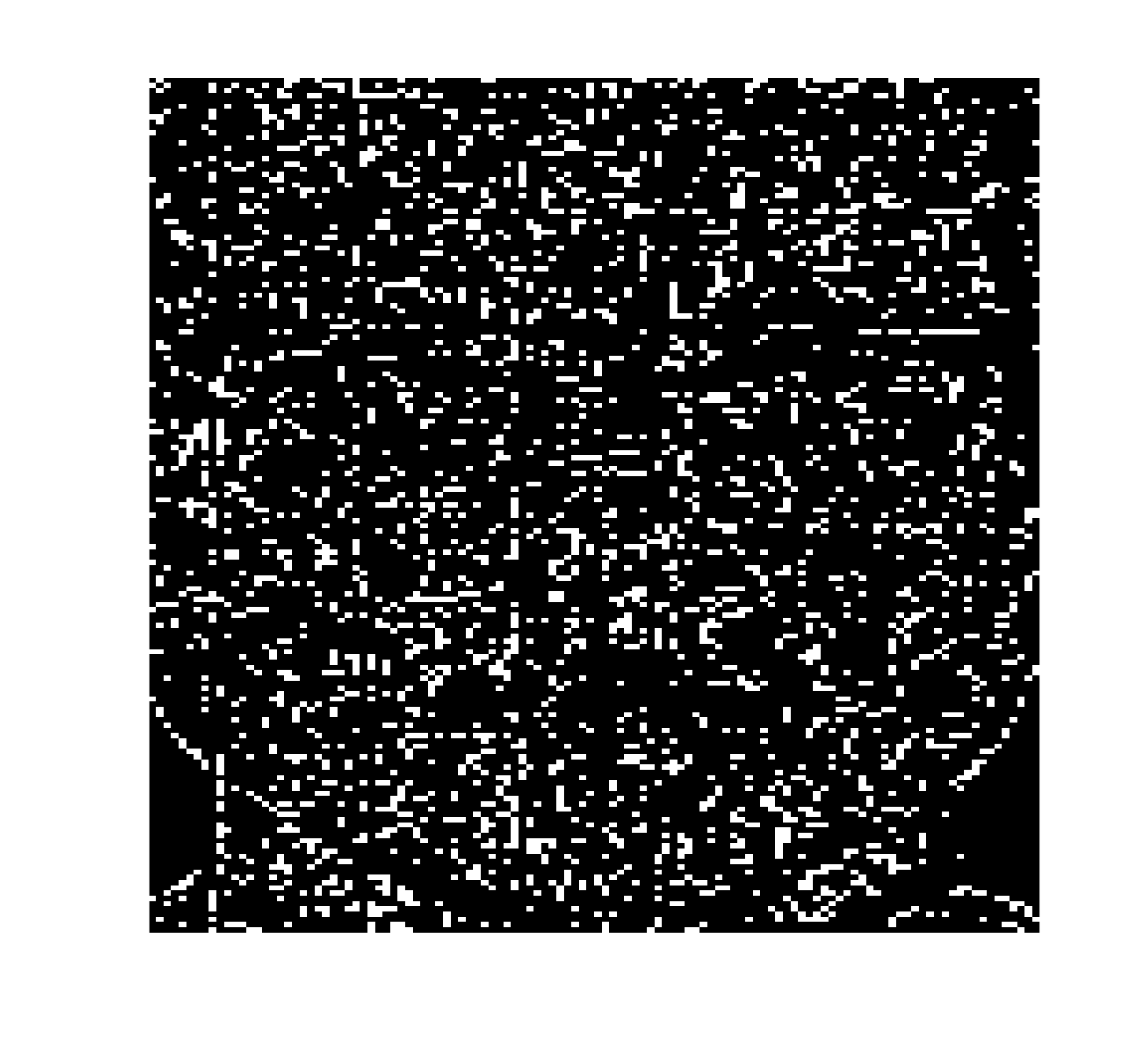}}
\subfigure[UP04]{\includegraphics[width=1.5cm]{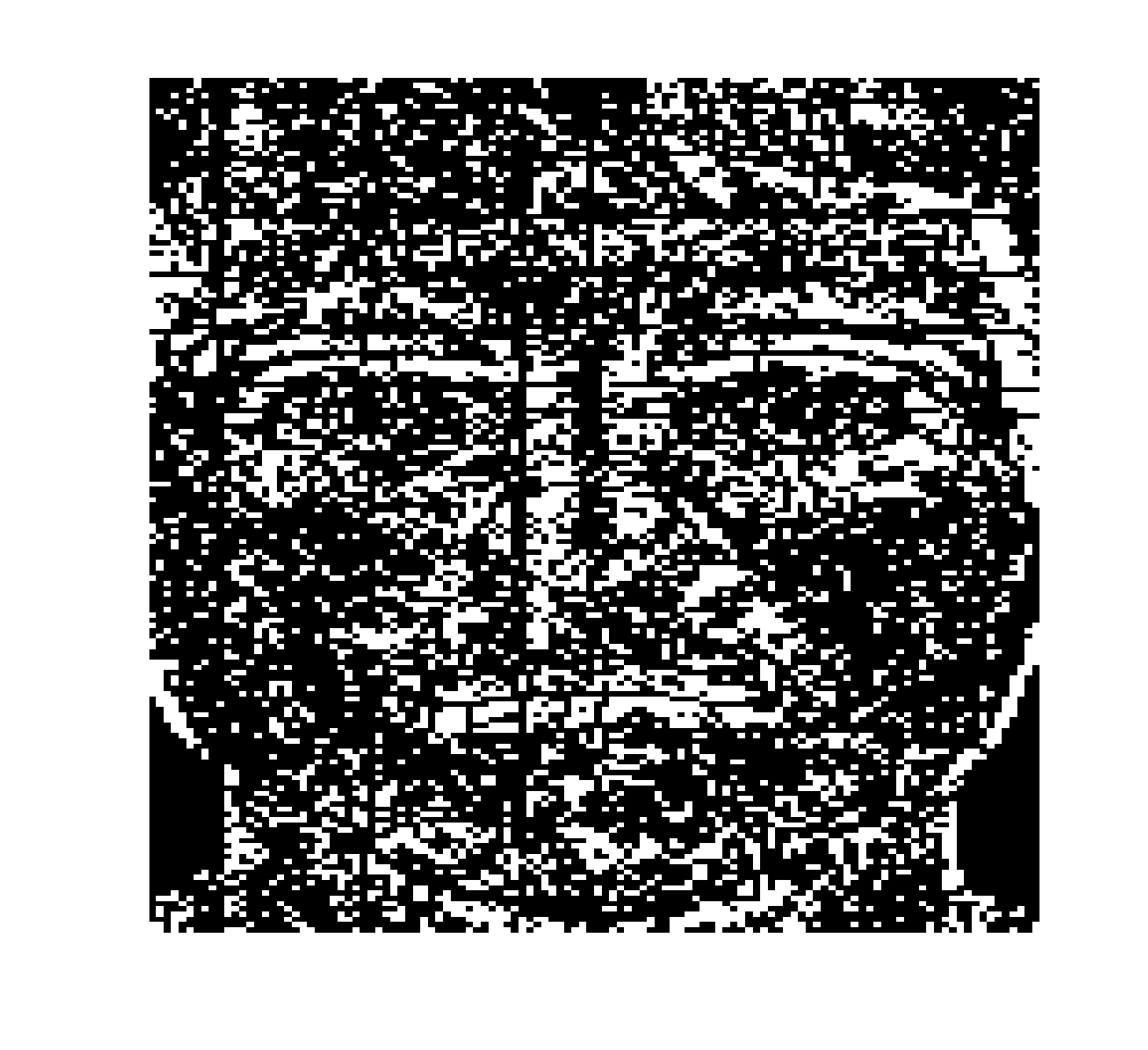}}
\subfigure[UP05]{\includegraphics[width=1.5cm]{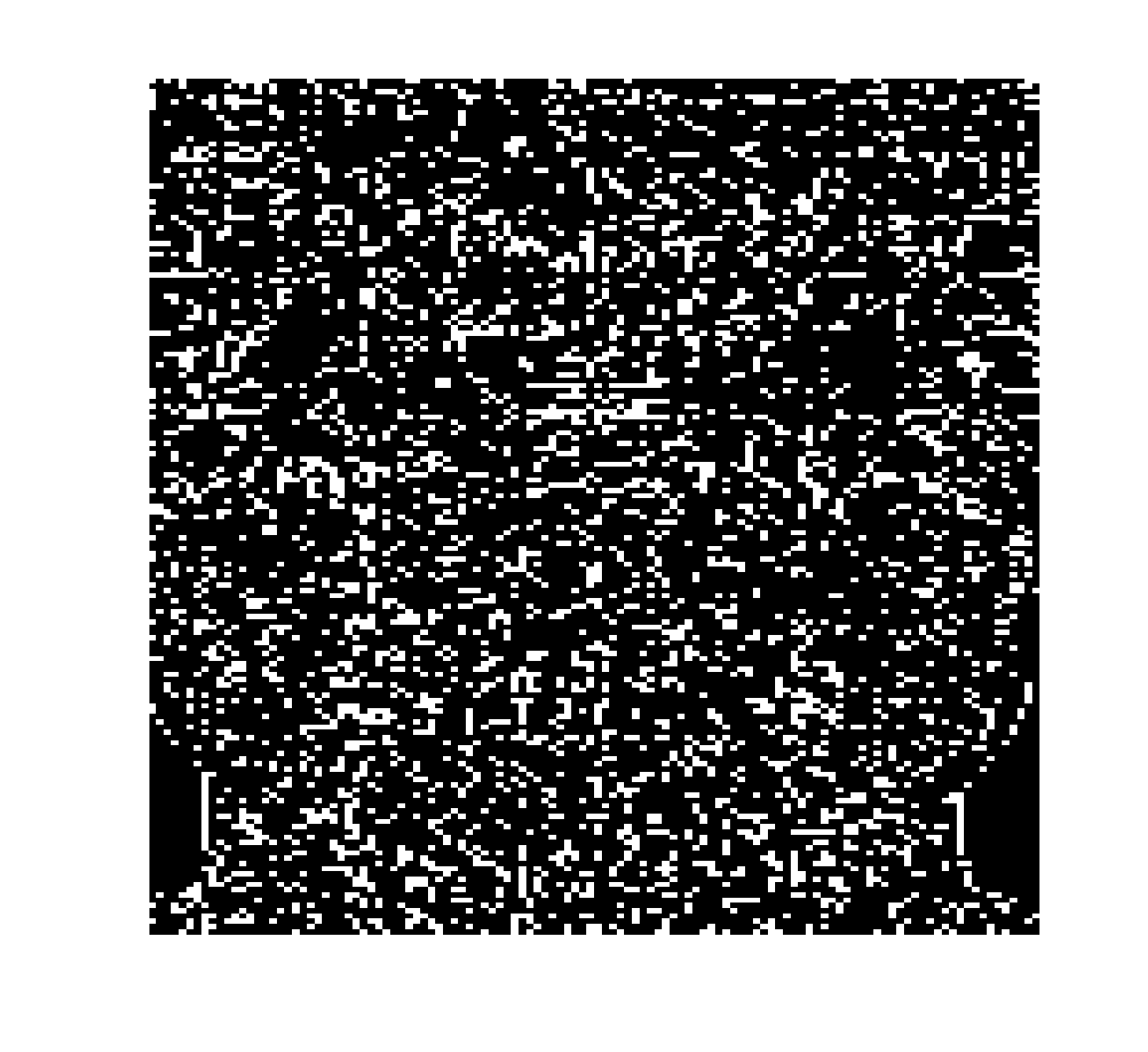}}
\subfigure[UP06]{\includegraphics[width=1.5cm]{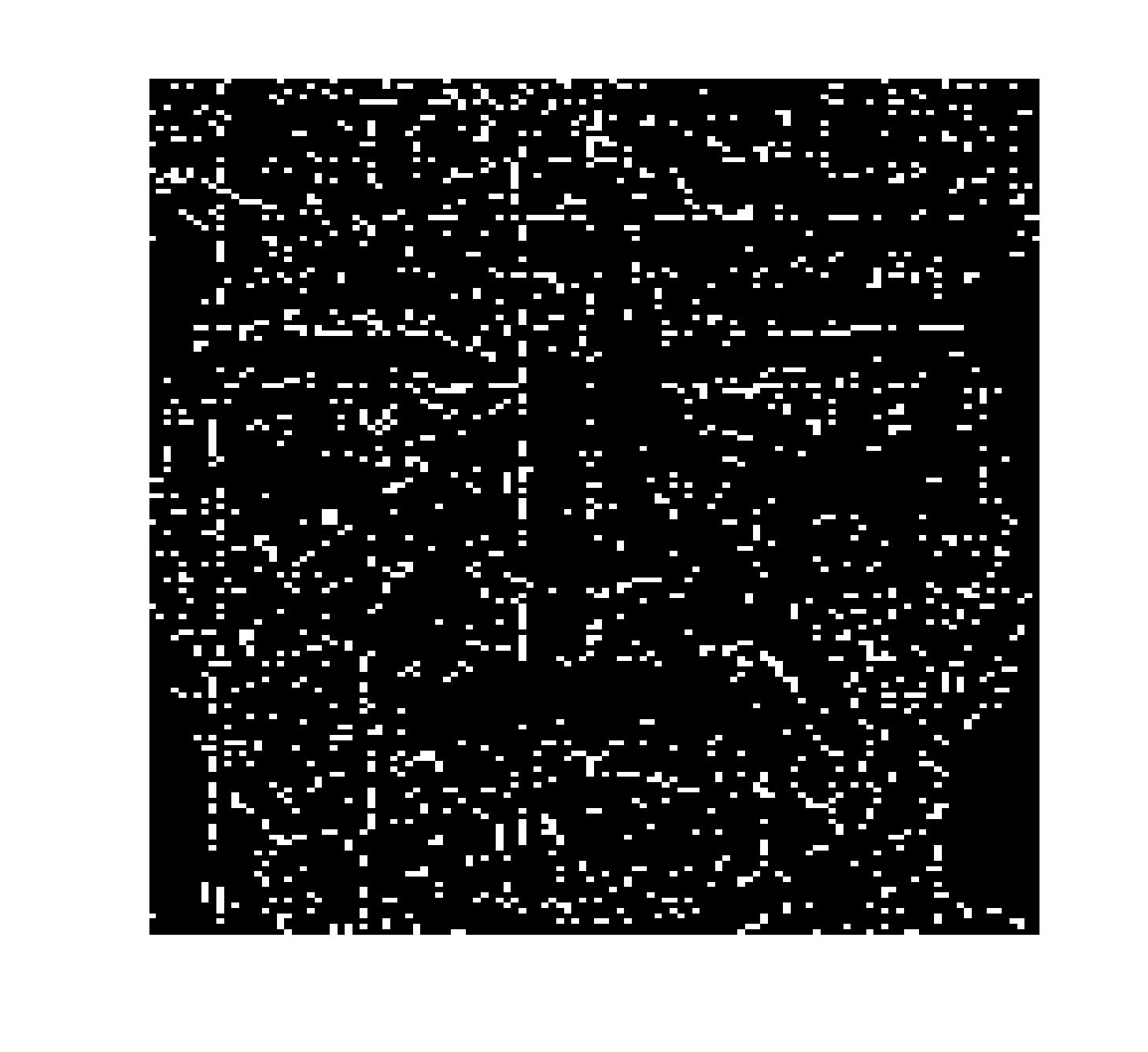}}
\subfigure[UP07]{\includegraphics[width=1.5cm]{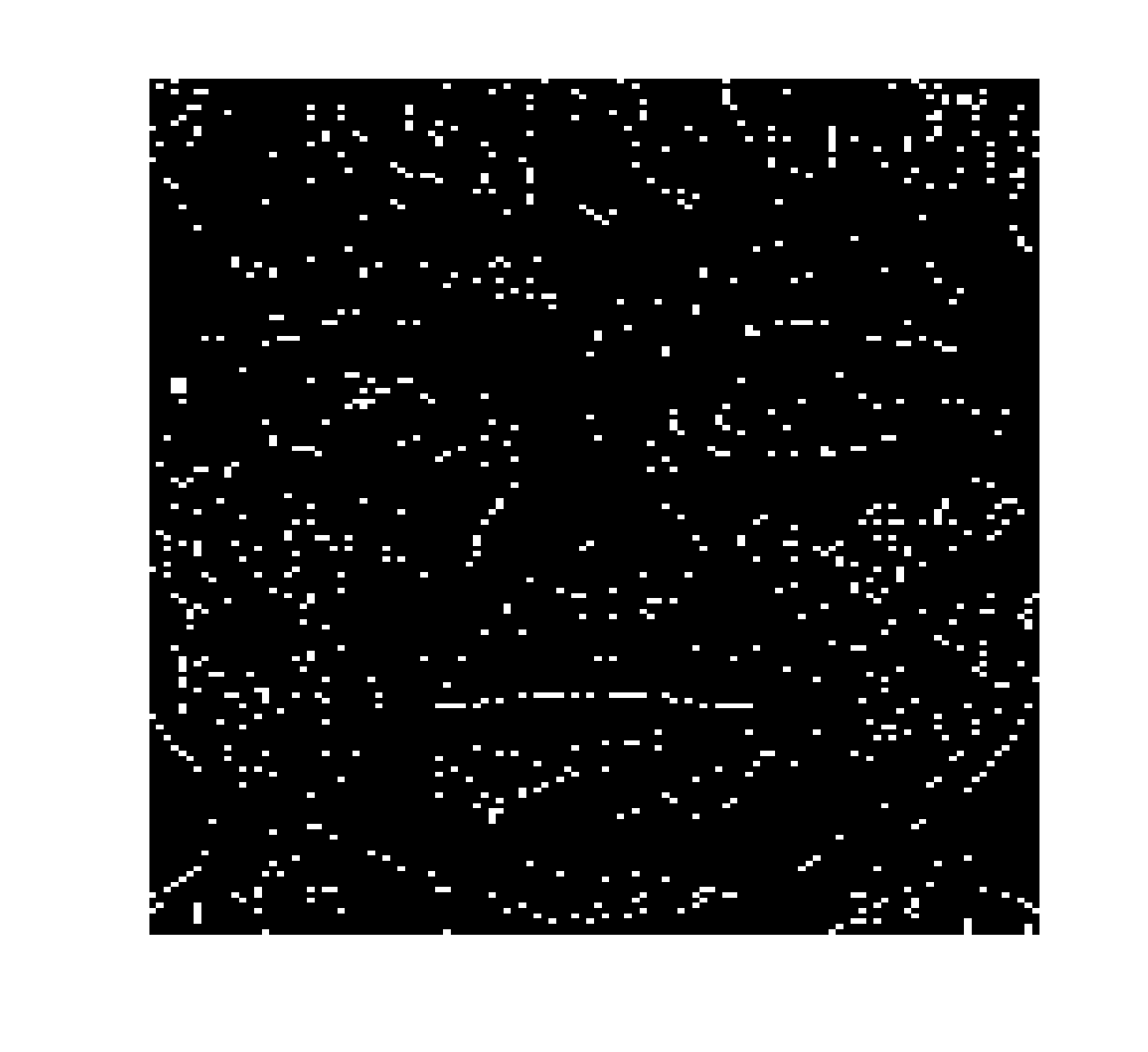}}
\subfigure[UP08]{\includegraphics[width=1.5cm]{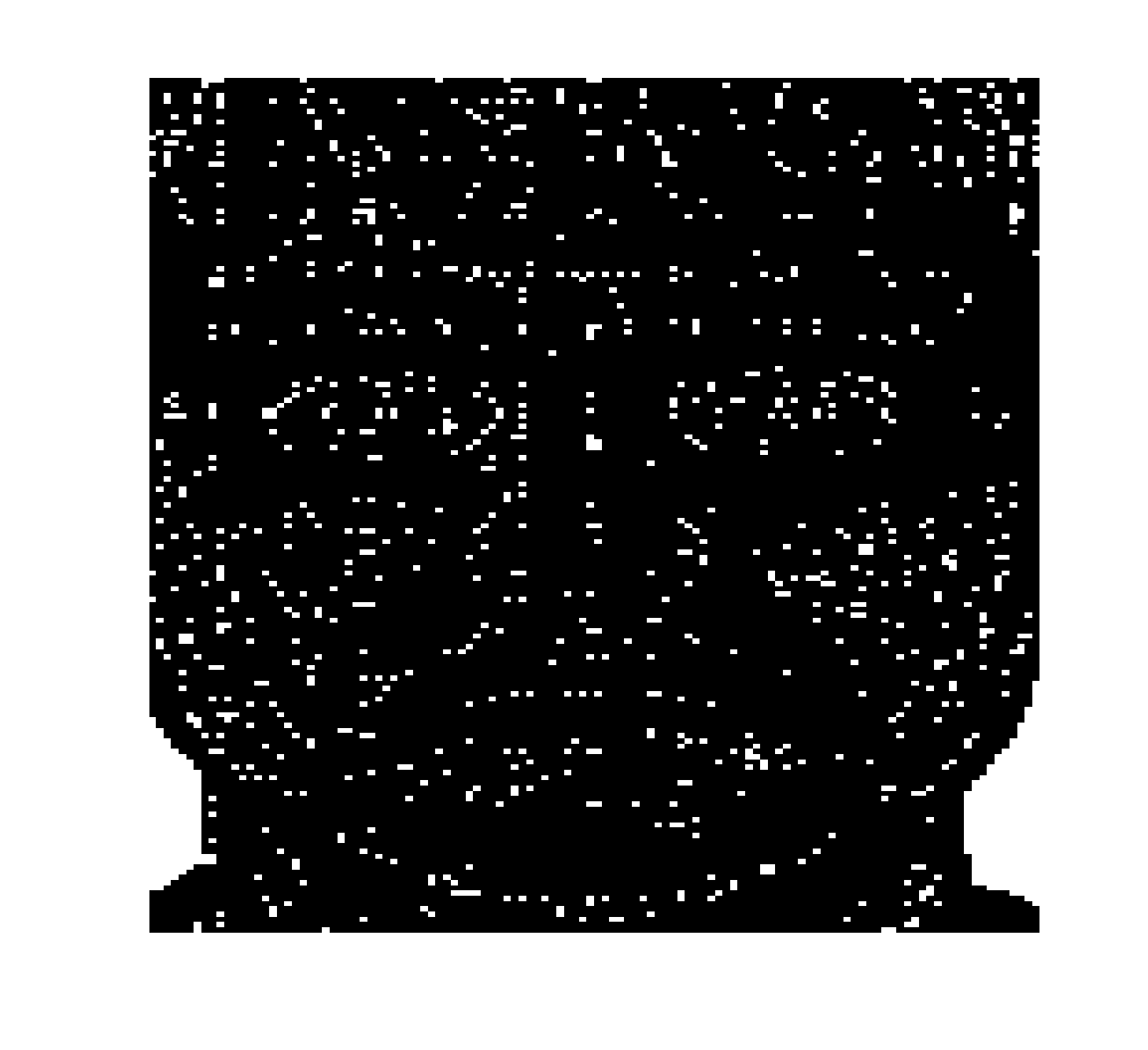}}
\subfigure[NUP]{\includegraphics[width=1.5cm]{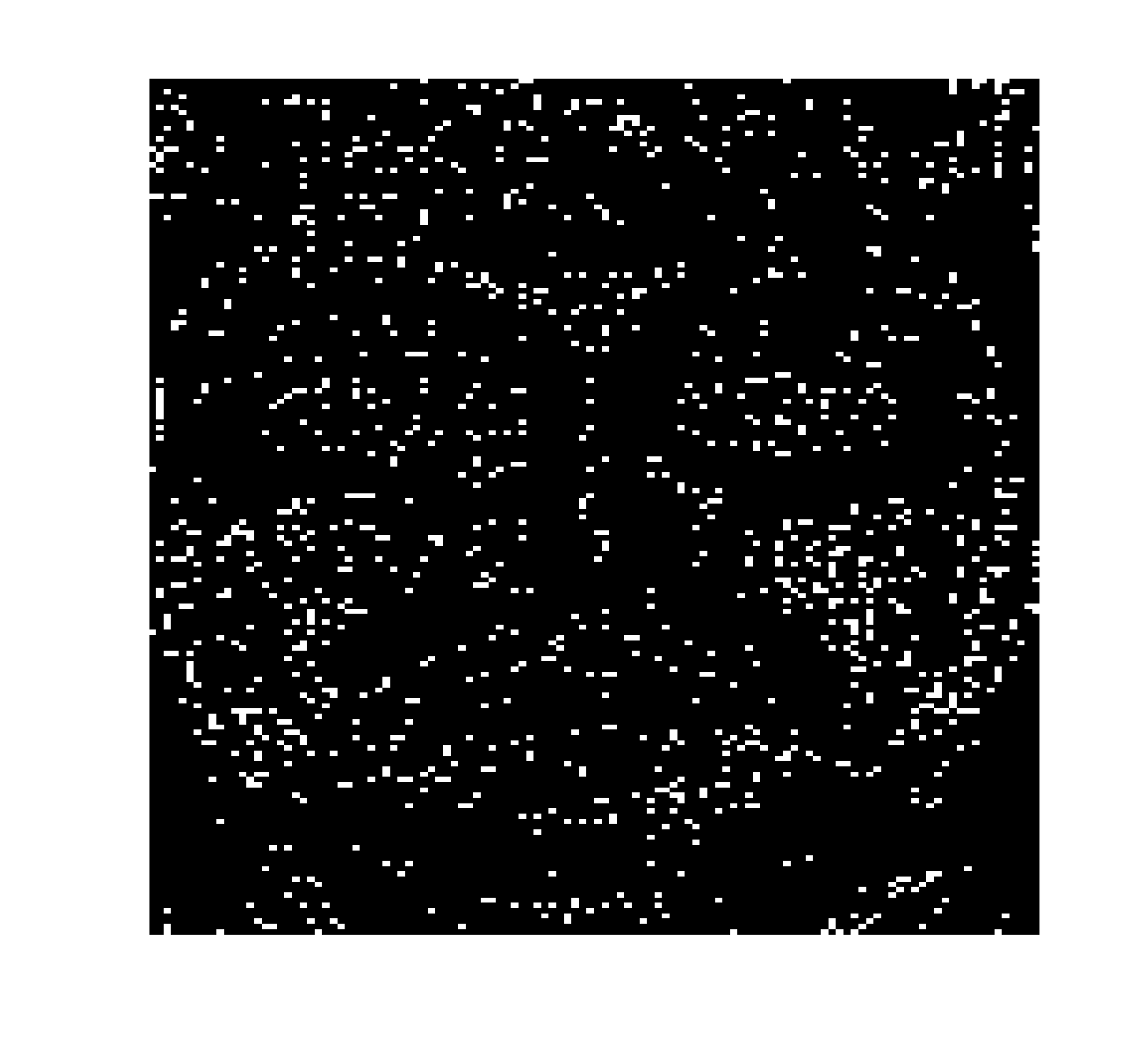}}\\

\subfigure[SBGP]{\includegraphics[width=1.5cm]{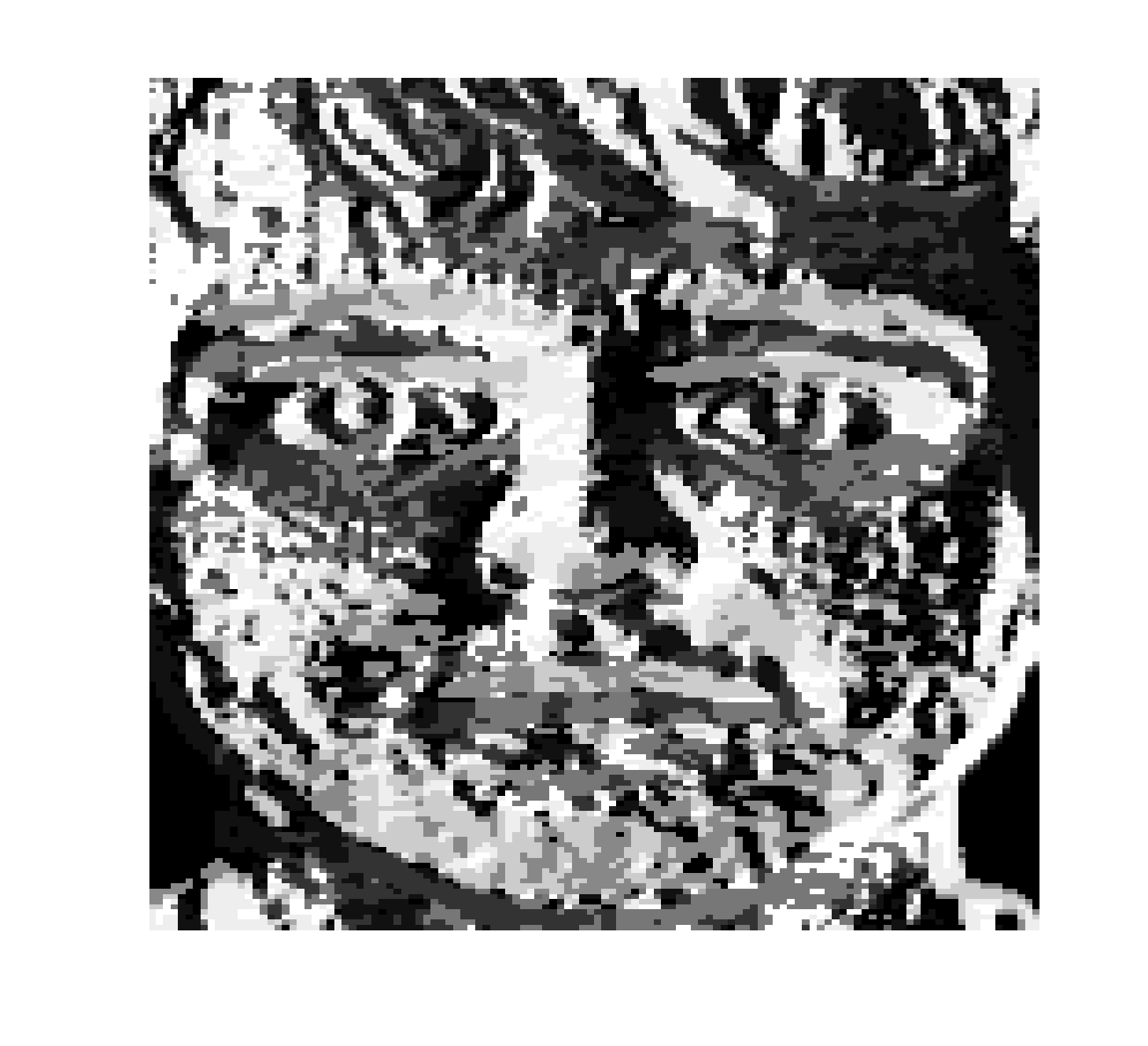}}
\subfigure[SP00]{\includegraphics[width=1.5cm]{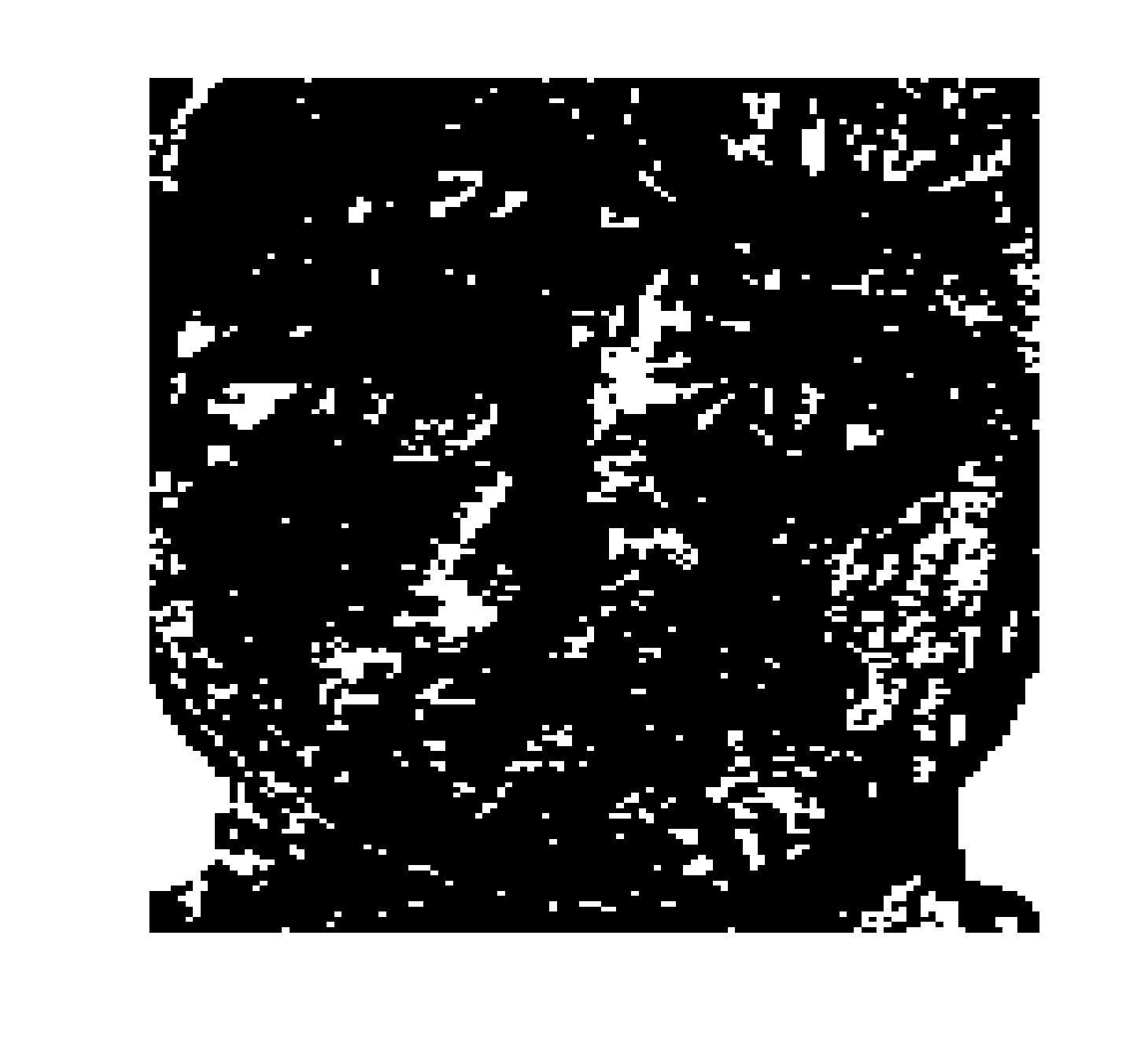}}
\subfigure[SP01]{\includegraphics[width=1.5cm]{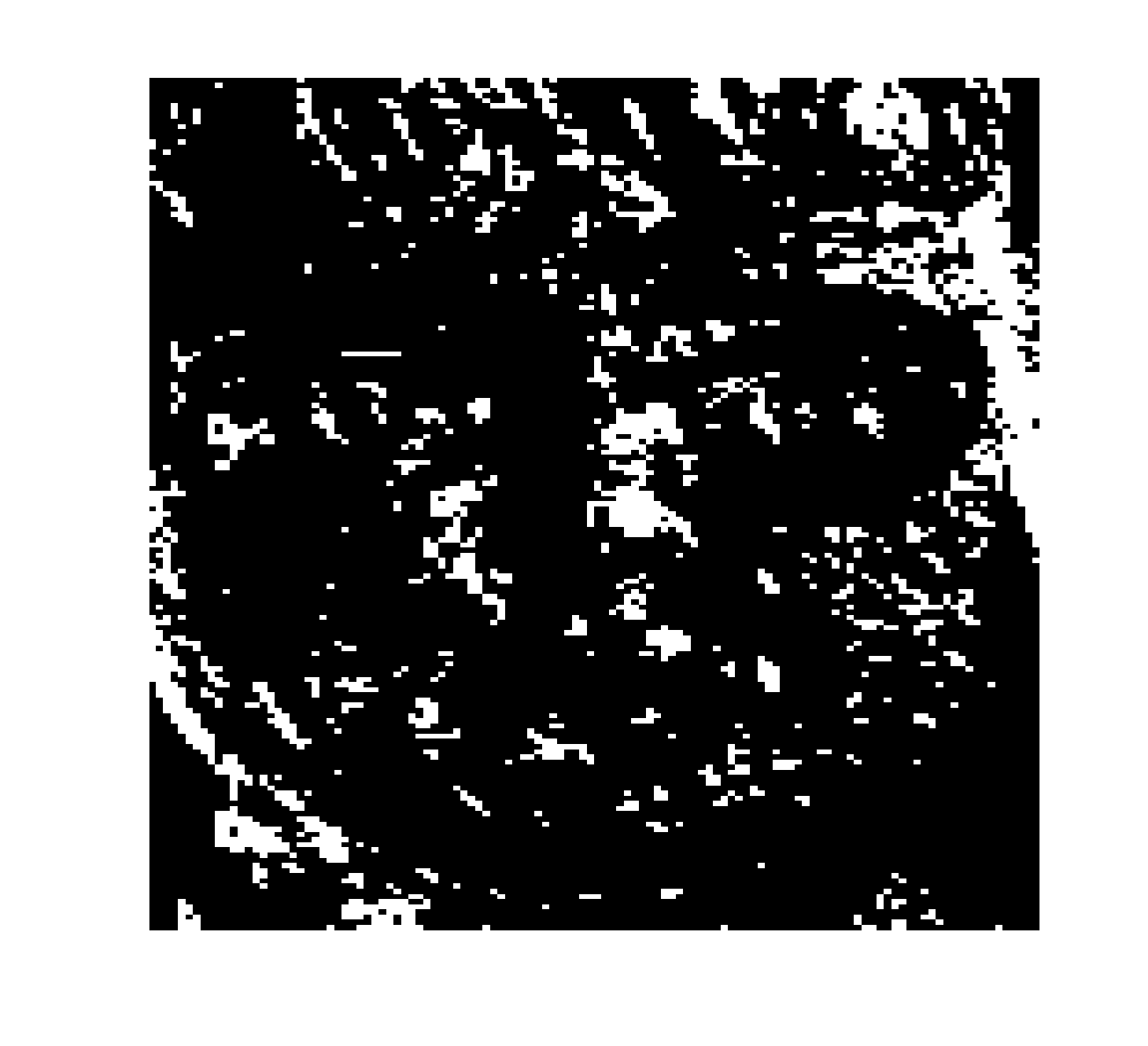}}
\subfigure[SP03]{\includegraphics[width=1.5cm]{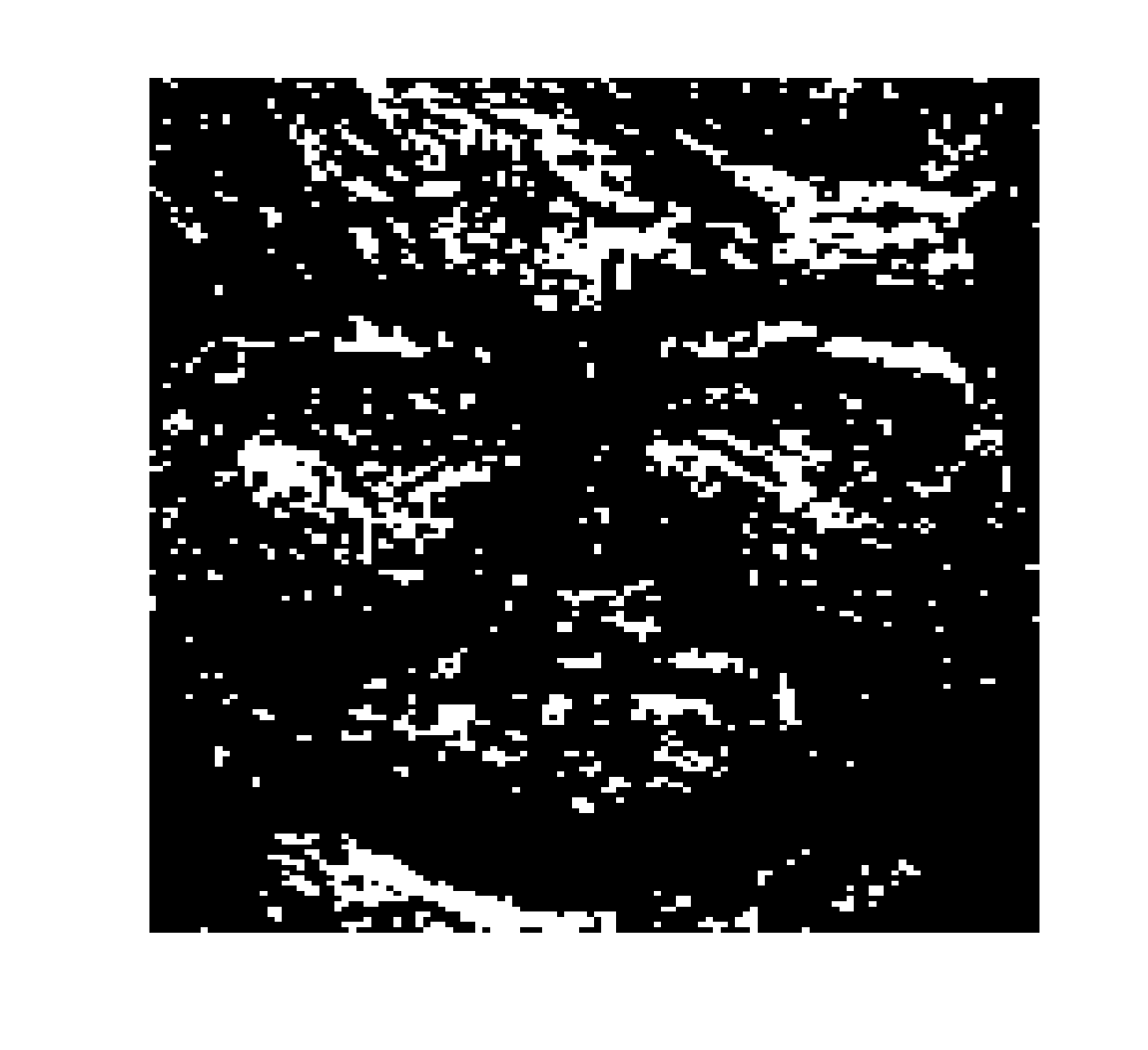}}
\subfigure[SP07]{\includegraphics[width=1.5cm]{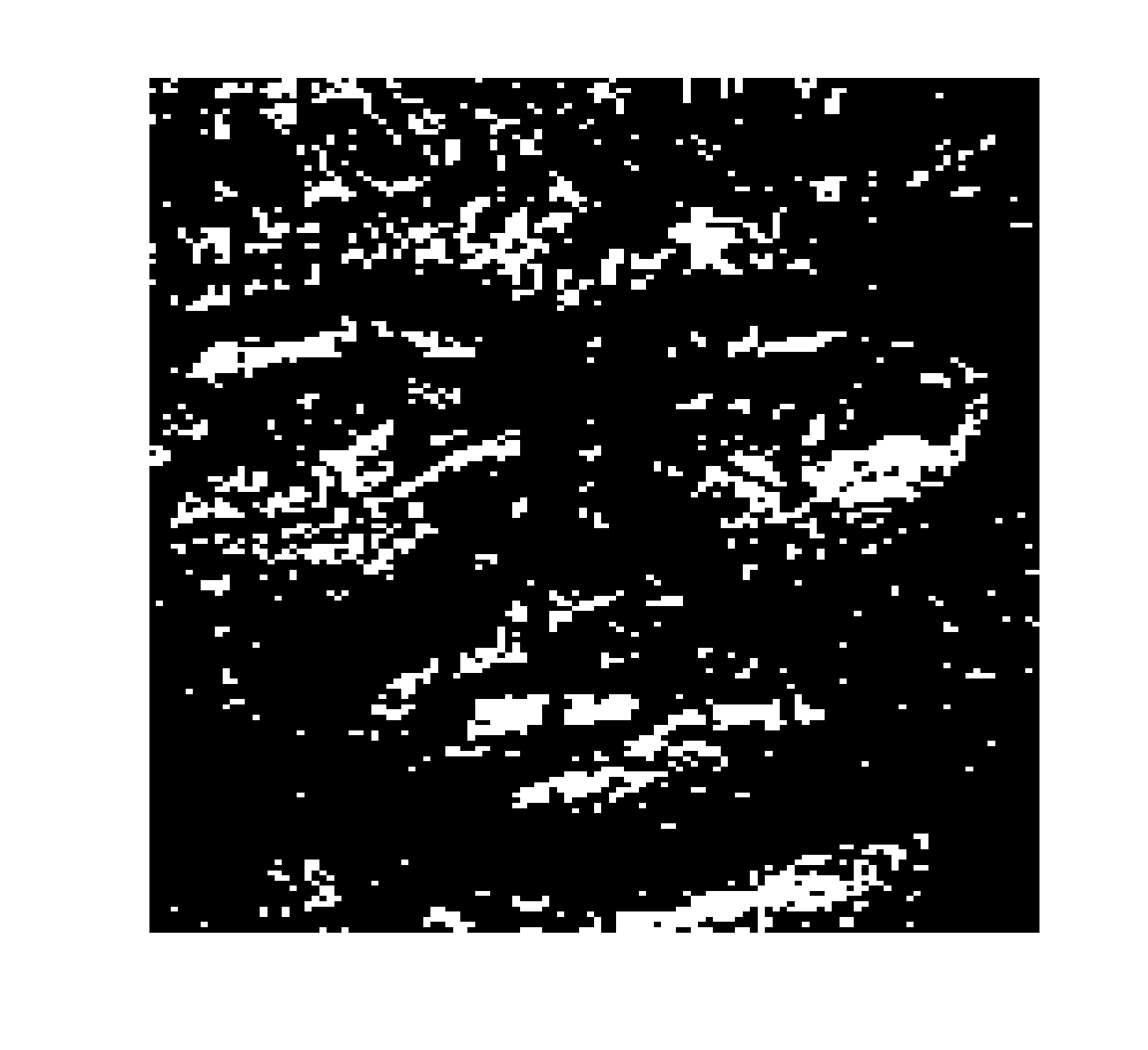}}
\subfigure[SP08]{\includegraphics[width=1.5cm]{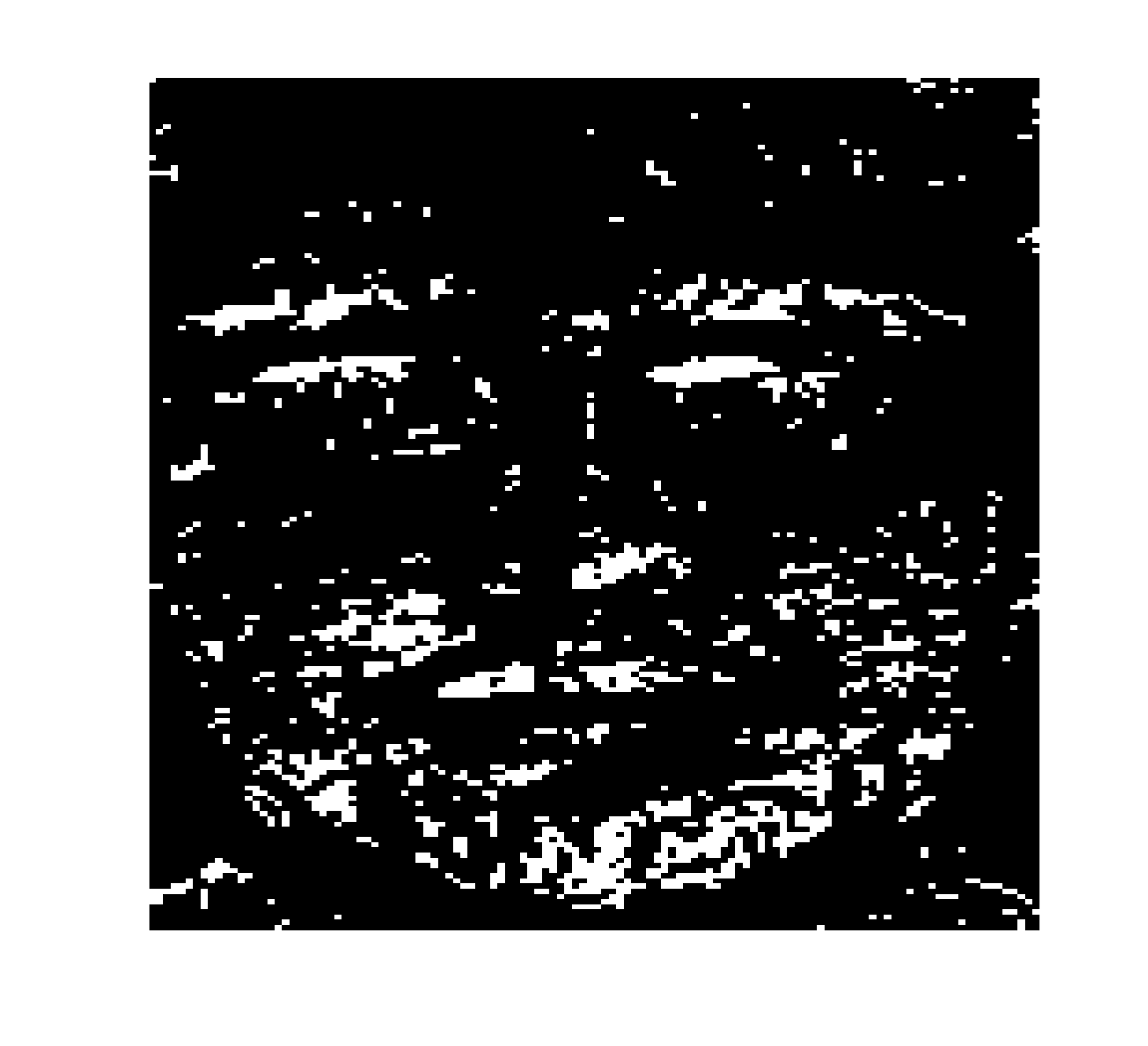}}
\subfigure[SP12]{\includegraphics[width=1.5cm]{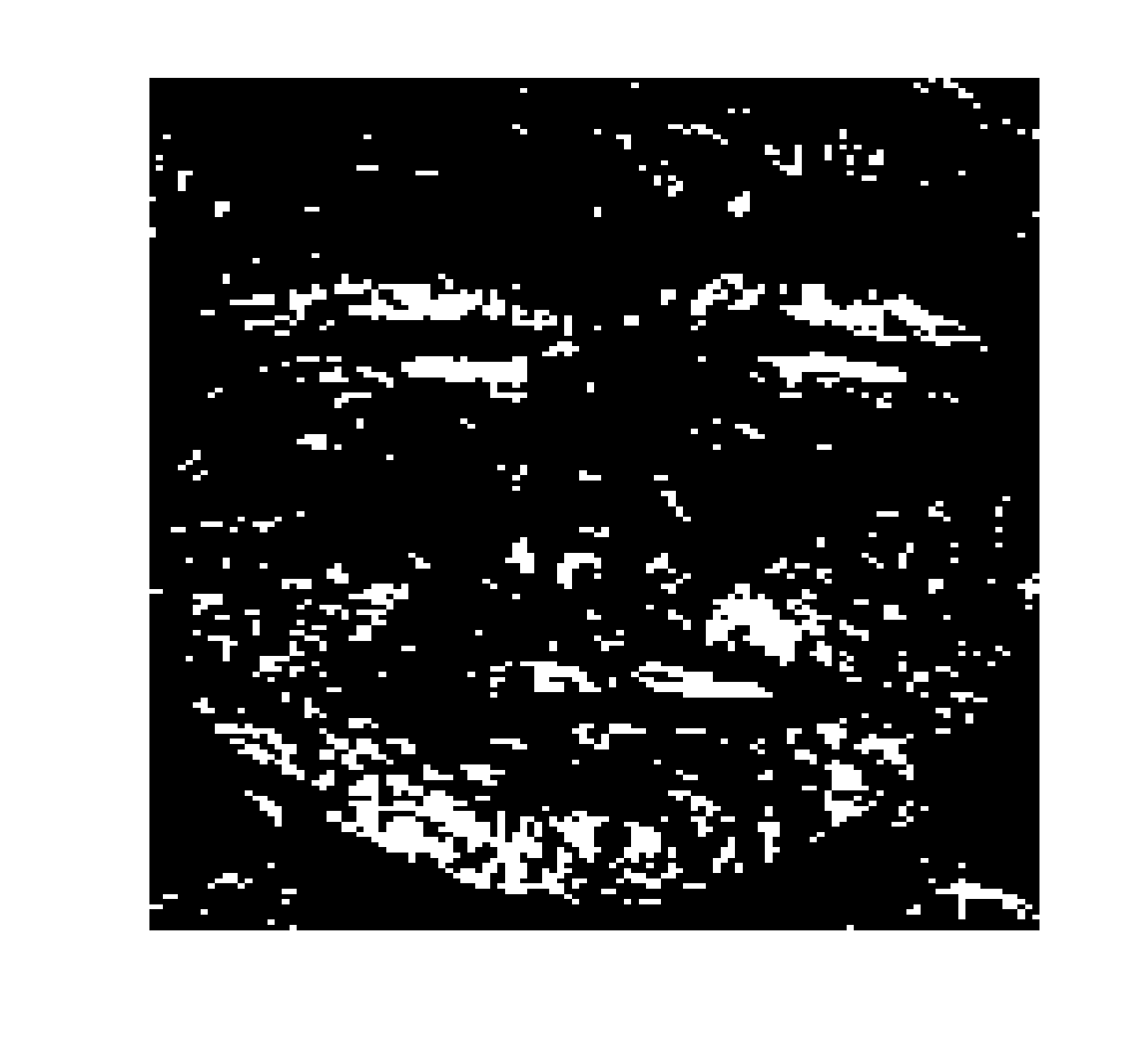}}
\subfigure[SP14]{\includegraphics[width=1.5cm]{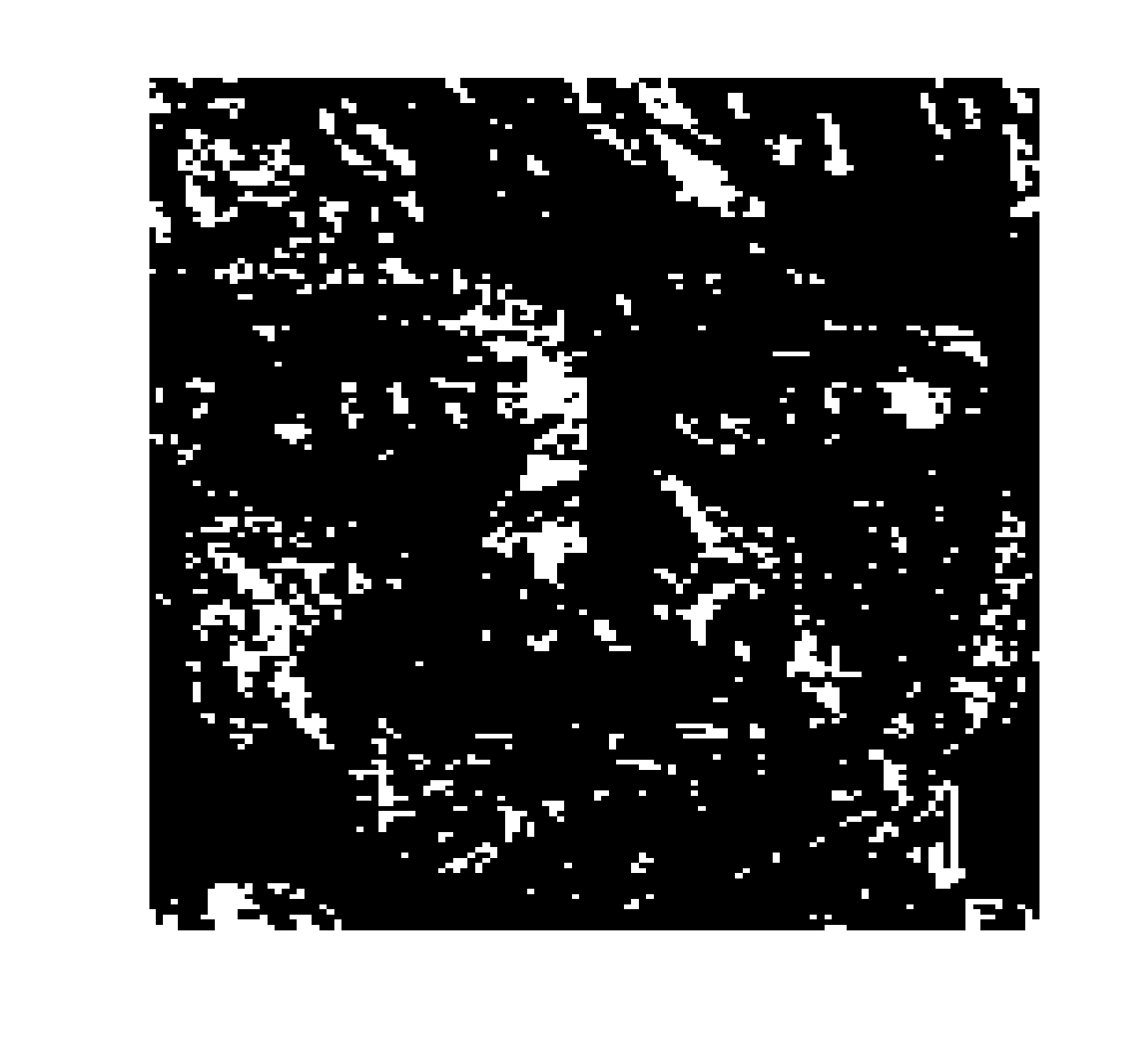}}
\subfigure[SP15]{\includegraphics[width=1.5cm]{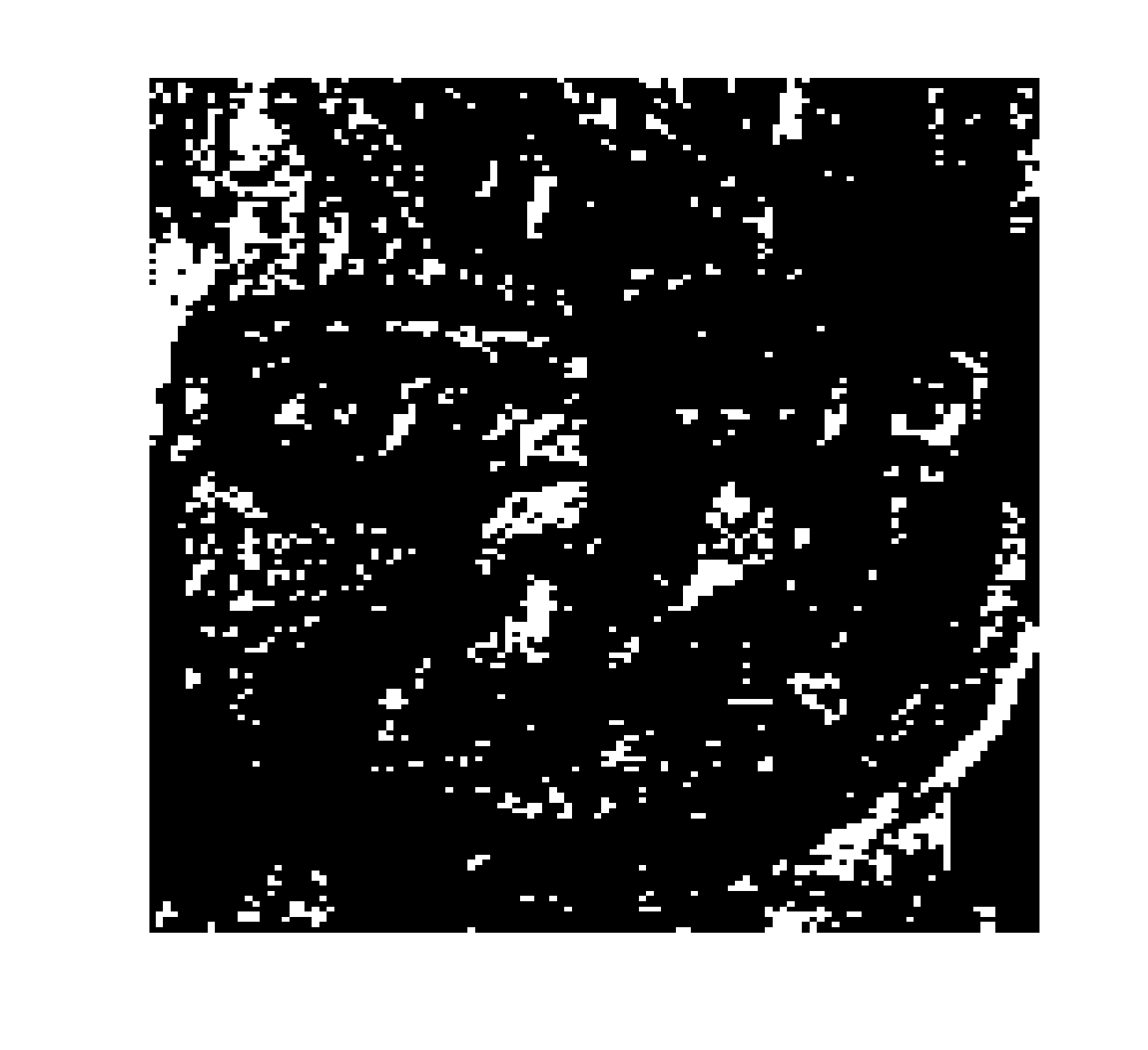}}
\subfigure[NSP]{\includegraphics[width=1.5cm]{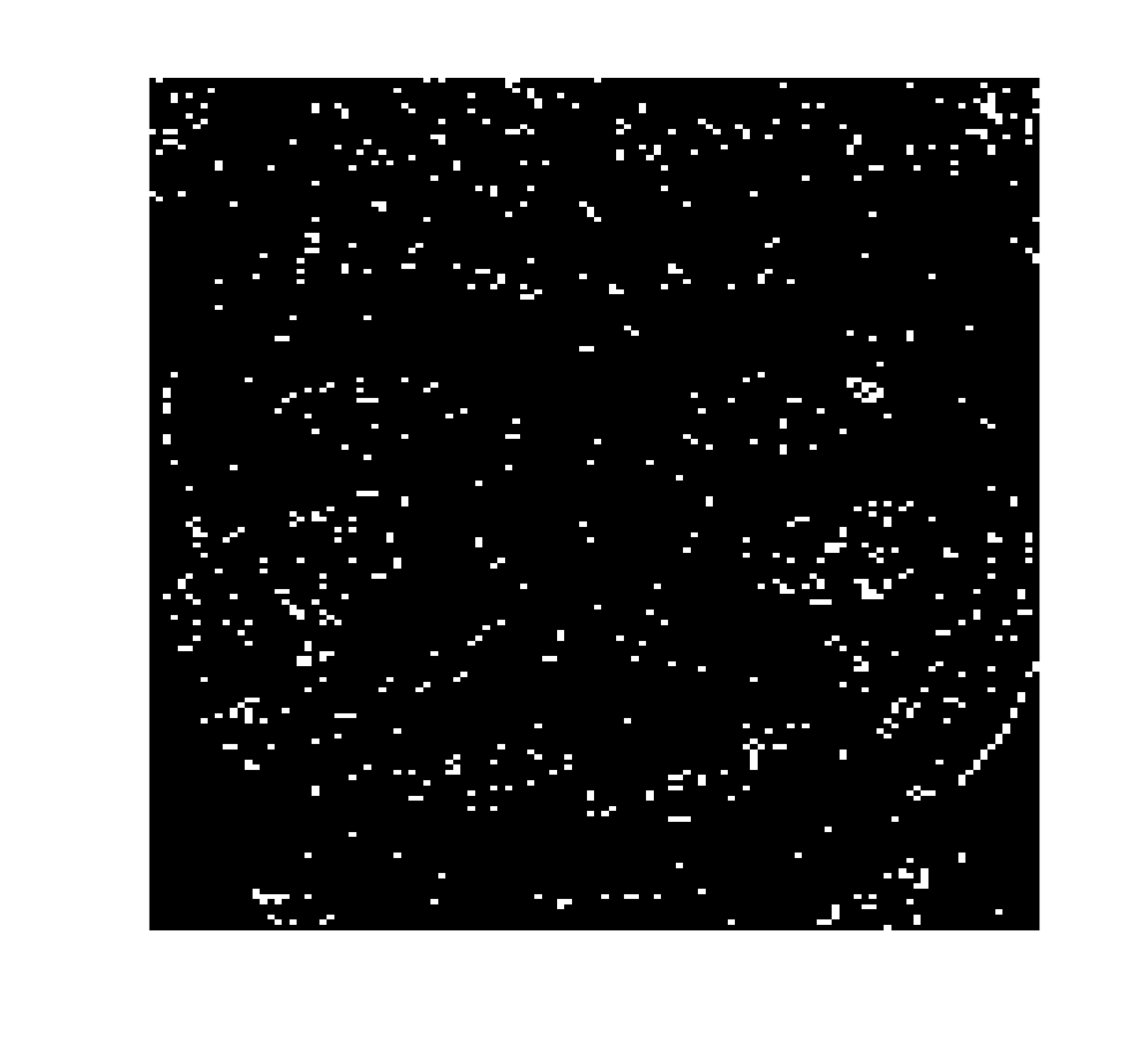}}\\

\caption{Demonstrations of Gabor, LBP and SBGP faces. Top: original AR face and Gabor magnitude faces of eight orientations and a fixed scale. Middle: $LBP_{8,1}^{riu2}$ face and  location maps of eight \emph{uniform} and one \emph{nonuniform} patterns. Bottom: $SBGP_{8,1}$ face and location maps of eight \emph{structural} and one \emph{nonstructural} patterns, corresponding to the labels in Fig.~\ref{fig:SBGP_UP}.}\label{fig:Gabor_LBP_SBGP}
\end{figure*}

This section discusses favorable characteristics of the SBGP descriptor and systematically compares it with two fundamental descriptors, LBP and Gabor wavelets, in terms of discrimination, robustness and complexity. Theoretical insights can be gained by discussing the underlying connections and distinctions among these descriptors, along with experimental studies.

Both SBGP and LBP employ advantageous binary strategy for extracting pixel correlations in local neighborhoods. However, SBGP differs from LBP in computing binary correlations, leading to distinctive properties between them.

\subsection{Discrimination}
\begin{figure}
    \begin{center}
   \includegraphics[height=3.5cm,width=5cm]{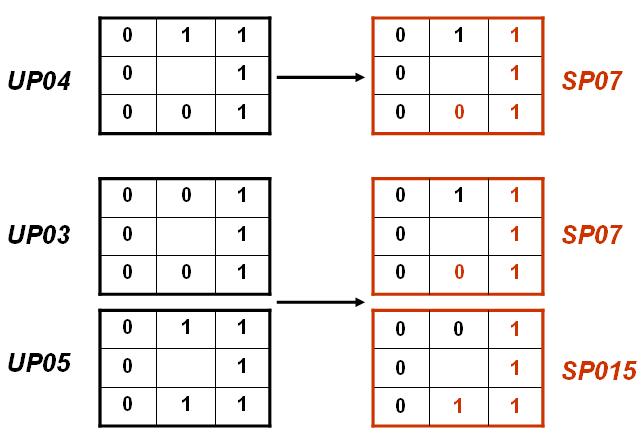}
    \end{center}
\caption{Connections between LBP \emph{uniform} patterns and SBGP \emph{structural} patterns. One of LBP $UP04$ can be transformed to SBGP $SP07$, and one of LBP $UP03$ or $UP05$ can be transformed to SBGP $SP07$ or $SP15$. But different structures of LBP $UP04$ may relate to different type of SBGP \emph{structural} patterns. Other types of LBP \emph{uniform} patterns are not guaranteed to be transformed to SBGP \emph{structural} patterns.}\label{fig:LBP2SBGP}
\end{figure}

%
%

The proposed SBGP descriptor is essentially an orientated edge detector with stronger orientational and discriminative capabilities than the LBP and Gabor representations. Fig.~\ref{fig:Gabor_LBP_SBGP} illustrates the outputs of the $SBGP_{8,1}$ and $LBP_{8,1}^{riu2}$ descriptors on a typical face image (of the AR database). It is evident that LBP detects various local textural features such as spots, corners and edges, while SBGP extracts orientated edge features. The $SBGP_{8,1}$ face assembles more facial information than $LBP_{8,1}^{riu2}$ face. Location maps of the $SBGP$ \emph{structural} patterns are more informative and discriminative than those of LBP \emph{uniform} patterns. The histogram statistics of the two descriptors in Fig.~\ref{fig:histSBGP_vs_histLBP} show that distributions of SBGP \emph{structural} patterns are fairly even, while distributions of LBP \emph{uniform} patterns mainly peaks at few patterns (UP04, UP05 and UP03).
This means that all SBGP \emph{structural} patterns contribute evenly, while LBP representation is dominated by few patterns.

As stated in \cite{Ojala2002}, these three types of LBP patterns detect edge information from textural image and lead to the finding that edge information dominates local textural features of face images. In fact, the local structures described by UP03 or UP04 or UP05 of the LBP are guaranteed to be represented by one of the SBGP \emph{structural} labels, as shown in Fig.~\ref{fig:LBP2SBGP}, determined by its orientation. In other words, LBP fuses all directions of the patterns into a single label, while \emph{structural} SBGP separately counts different oriented edges in eight labels to increase discriminative power. Although HOG also computes histogram from multiple orientations, the gradient orientations used by HOG are only computed from four local neighbors, which cannot effectively detect edge information and are insufficient to realize meaningfully the local structures.


%

Furthermore, we argue that the SBGP processes some essential properties of the human vision system, which is characterized by spatial locality, orientation and scale selectivity \cite{Olshausen1996}, and responds strongly to specifically oriented lines or edges positioned in their receptive fields \cite{Marr1980}. Gabor wavelets are a well-known model for describing these properties and have had considerable successes in image feature representation\cite{Manjunath1996,Liu2002}. The inherent characteristics of the SBGP can resemble these properties by enforcing its locality in both pixel and block levels, by taking the statistics of edge orientations to improve orientational capability, and by defining a tunable spatial resolution for scale selection. Fig.~\ref{fig:Gabor_LBP_SBGP} provides such an intuitive view that SBGP faces preserve stronger orientations than Gabor faces.



\subsection{Robustness}
%

\begin{figure}
    \begin{center}
    \subfigure[Faces]     {\includegraphics[width=1.8cm,height=2.8cm]{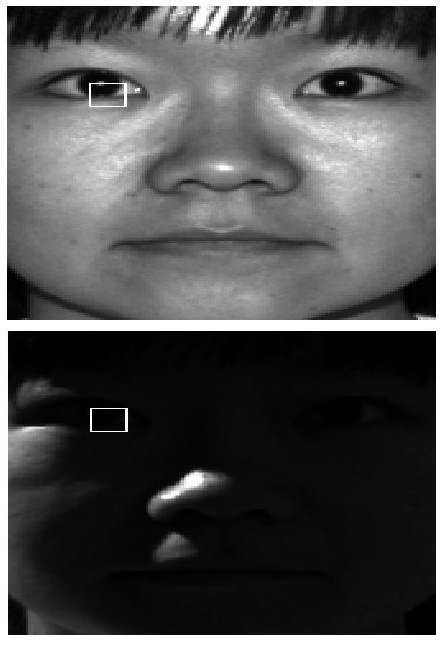}}
\subfigure[Intensity]     {\includegraphics[width=2.2cm,height=3cm]{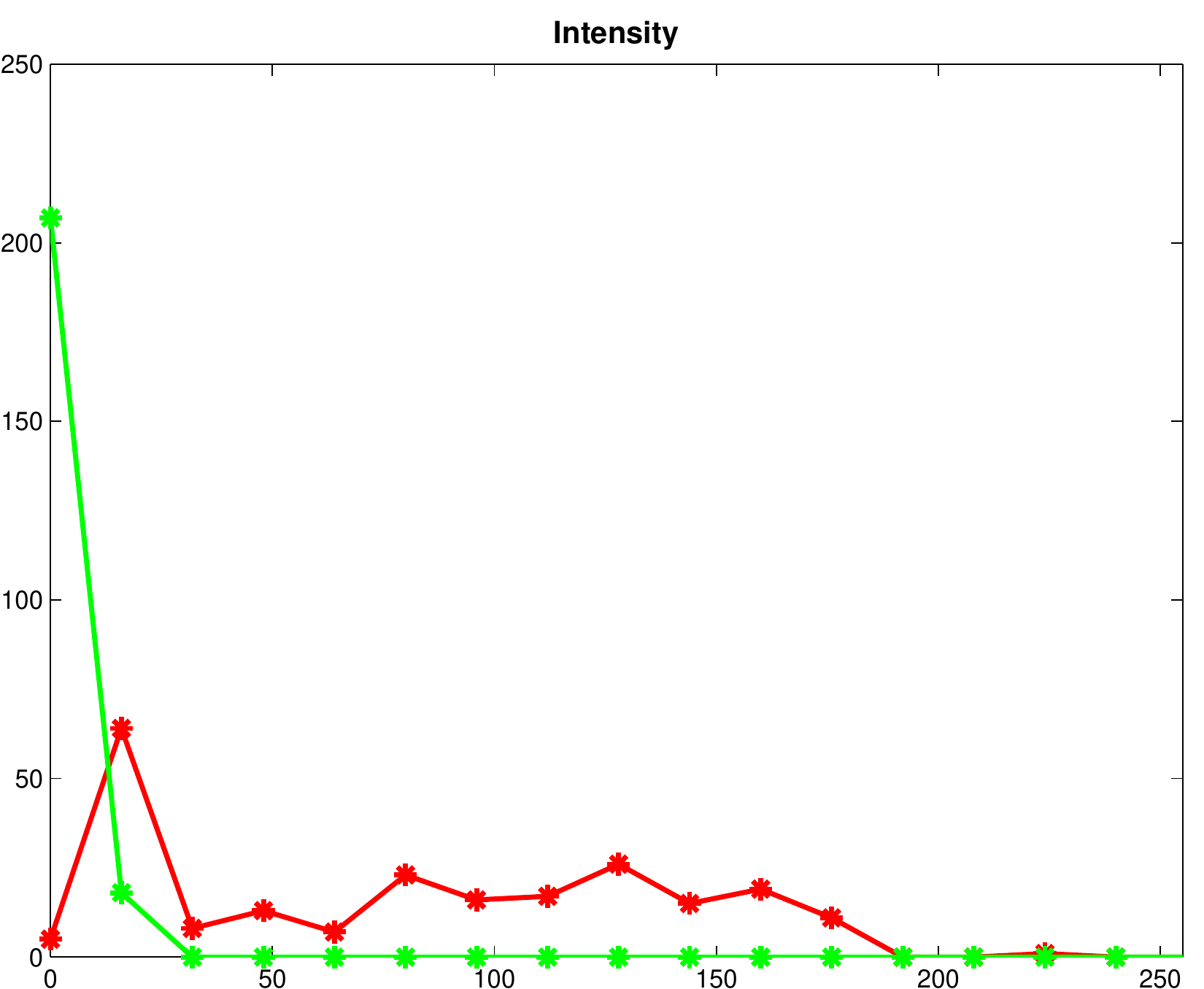}}
\subfigure[LBP]     {\includegraphics[width=2.2cm,height=3cm]{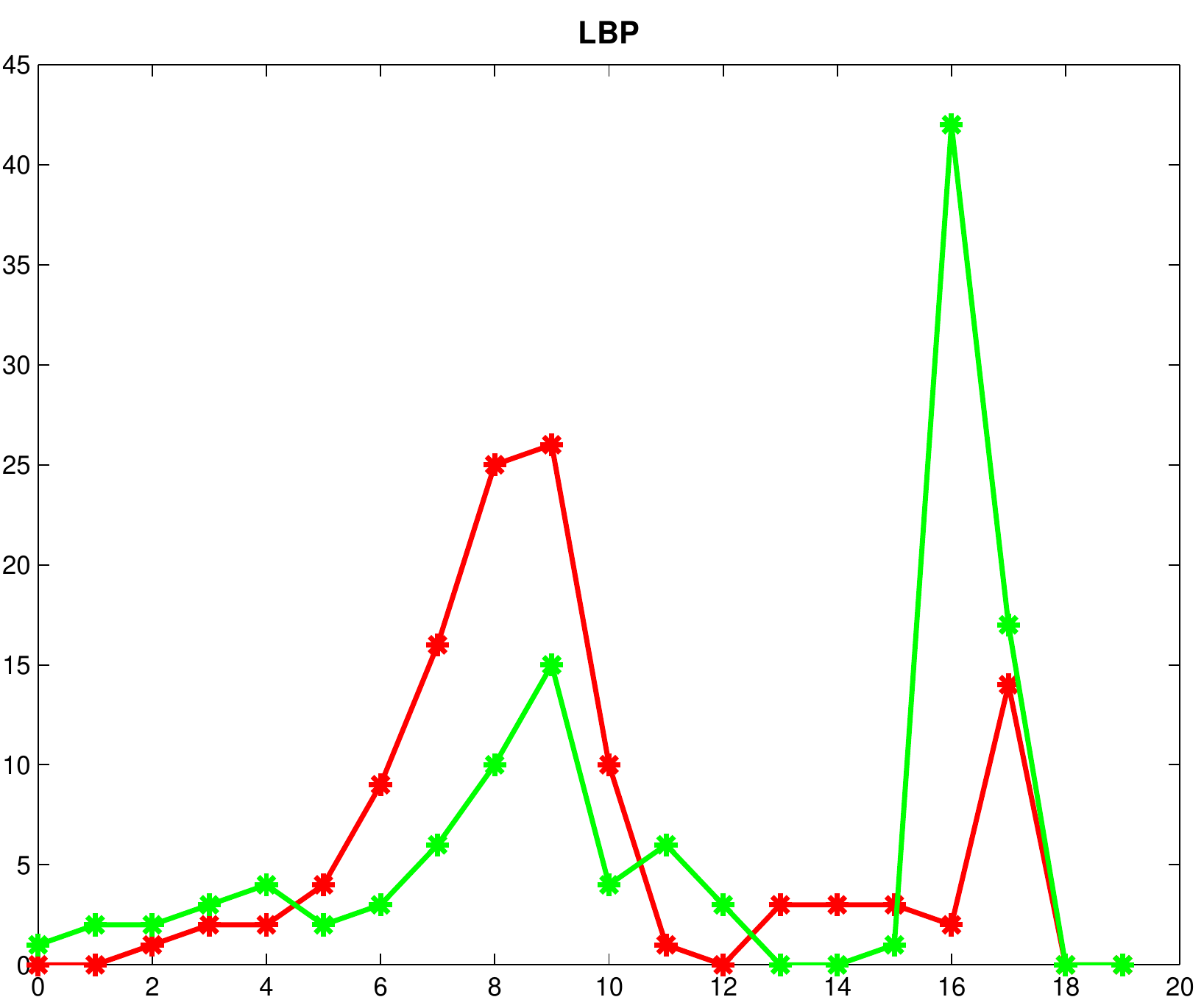}}
\subfigure[SBGP]     {\includegraphics[width=2.2cm,height=3cm]{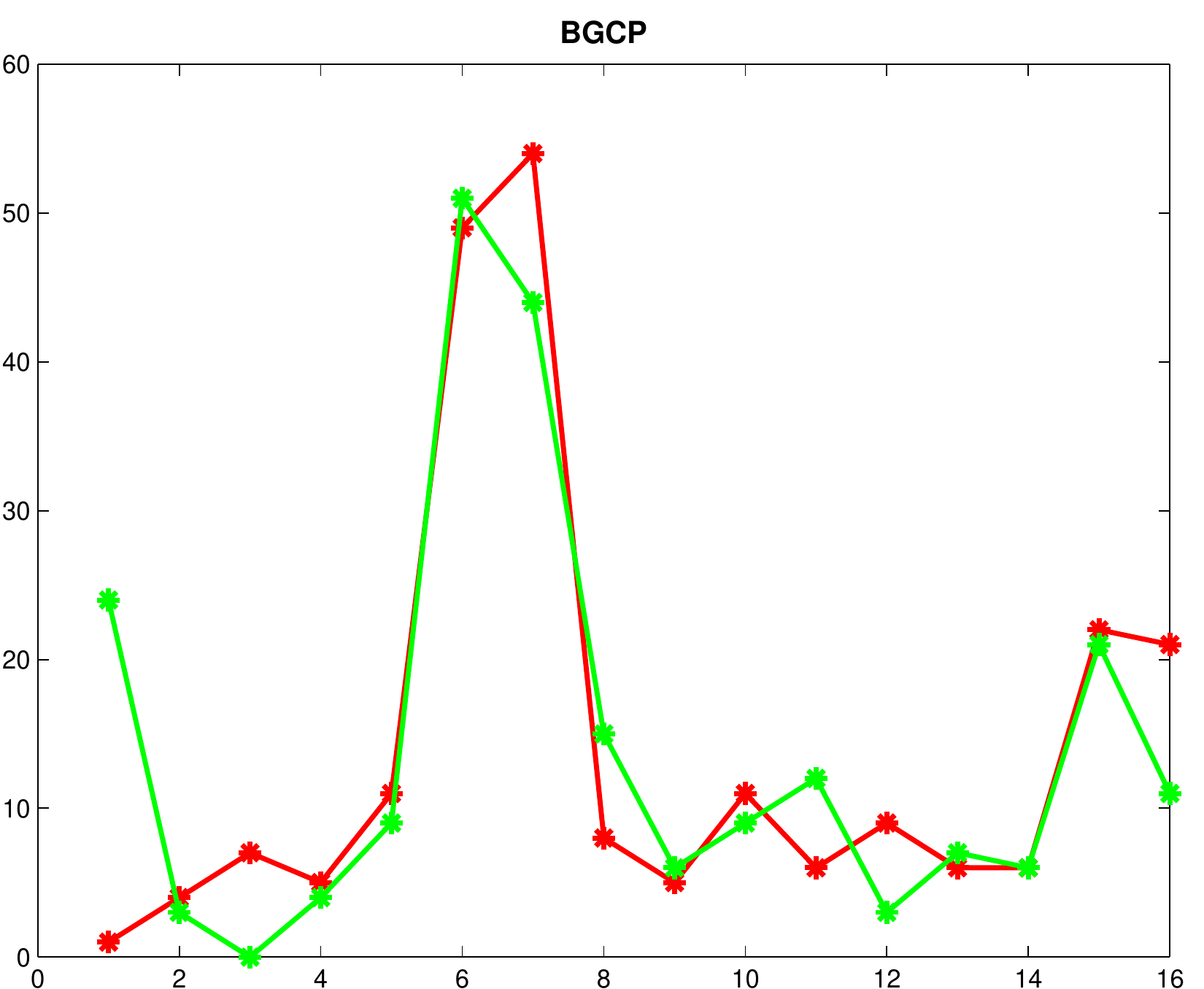}}
    \end{center}
\caption{Histogram statistics of pixel intensity, LBP and SBGP patterns on the two blocks ($15 \times 15$) of two faces in (a) (same identity) with significantly different illuminations.}\label{fig:Rob}
\end{figure}

LBP achieves gray-scale invariance by discarding intensity contrast. SBGP employs a heuristic from the IGO representation to further take advantage of the gradient domain where local representation is inherently invariant against illumination changes. The illumination invariance of IGO based representations has been verified in the reflectance model by canceling out illumination functions of different directions when computing the ratio of gradients\cite{Zhang2009b}. Benefiting from this merit, SBGP achieves stronger robustness to illumination variation than LBP by further discarding gradient contrast (as discussed in Section 2.2).  Fig.~\ref{fig:Rob} shows the histogram statistics of LBP and SBGP patterns on two exemplar face blocks, illustrating the improved robustness of SBGP against extreme illumination condition.

Furthermore, the SBGP discards \emph{non-structural} patterns, which contain non-smooth or discontinuous changes of local pixels. These patterns are often caused by noise or outliers, and contain little structural and meaningful information. The experimental statistics on 5032 face images from the AR and YaleB databases show that there is a very low proportion of these \emph{non-structural} patterns in general face images, only 5.8$\%$, lower than the proportion of LBP \emph{nonuniform} patterns, at about 8.5$\%$. In addition, some LBP \emph{uniform} labels have small numbers of patterns. For example, the UP00 and UP08, as shown in Fig.~\ref{fig:histSBGP_vs_histLBP}(b) and Fig.~\ref{fig:Gabor_LBP_SBGP} (middle row), detect bright and dark spots, respectively, in textural images \cite{Ojala2002}. These types of patterns may also include some irregular appearances, such as noisy spots and corrupted pixels.
\begin{figure}
\begin{center}
\subfigure[Faces]    {\includegraphics[width=1.65cm,height=3cm]{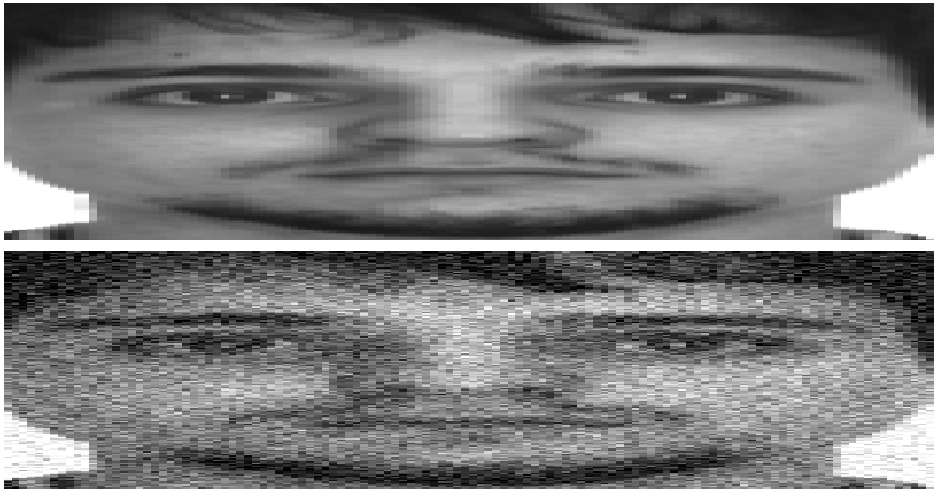}}
\subfigure[NonS]     {\includegraphics[width=1.65cm,height=3cm]{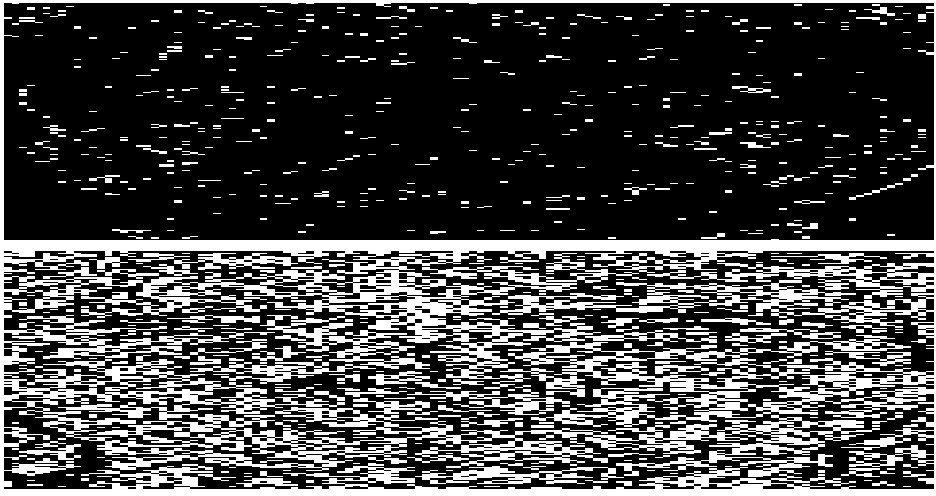}}
\subfigure[NonU]     {\includegraphics[width=1.65cm,height=3cm]{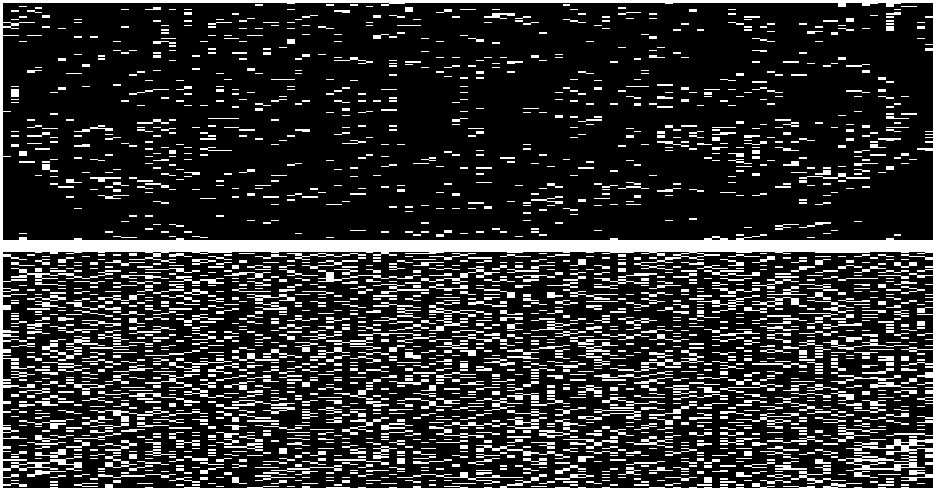}}
\subfigure[Spots]     {\includegraphics[width=1.65cm,height=3cm]{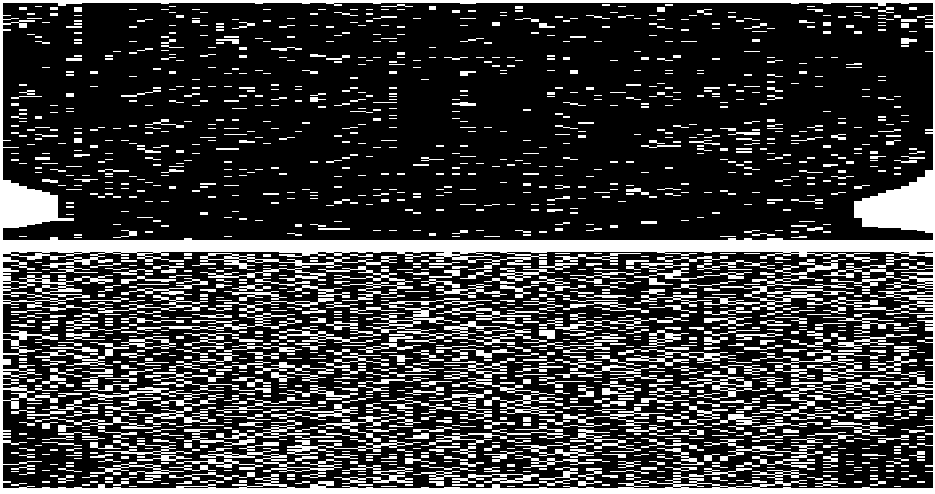}}
\subfigure[Corners]     {\includegraphics[width=1.65cm,height=3cm]{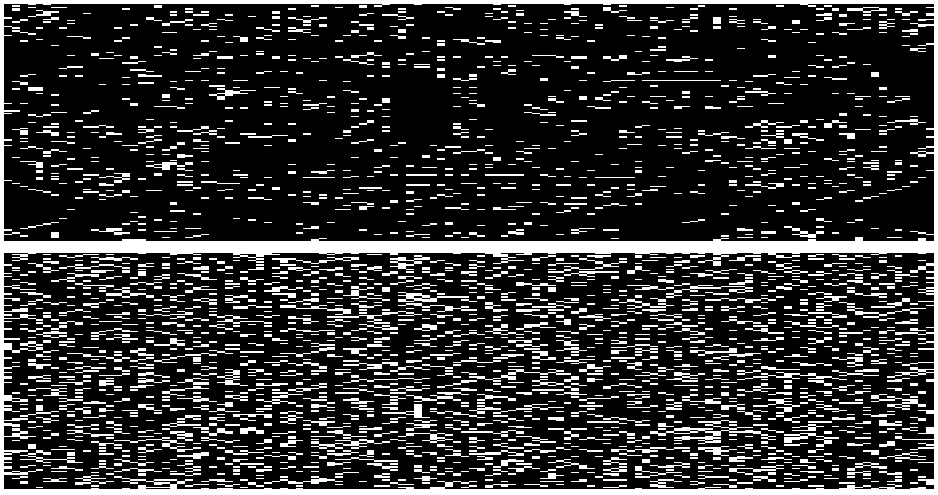}}
\end{center}
\caption{Gaussian noise on the SBGP \emph{non-structural }and LBP \emph{non-uniform} patterns. (a) Original face and with adding Gaussian noise; (b) the SBGP \emph{non-structural} and (c) LBP \emph{non-uniform} patterns; (d) the LBP UP00 and UP08 patterns(spots); and (e) the LEB UP01 and UP07 patterns (corners).}\label{fig:Rob2}
\end{figure}

As shown in Fig.~\ref{fig:Rob2} (b), the SBGP \emph{non-structural} patterns contain a large amount of noise. The SBGP discards them to mitigate the effect of noise. While the LBP retains all of its \emph{nonuniform} patterns and assign additional labels to them. Subsequently the numbers of LBP \emph{non-uniform}, spot and corner patterns increase dramatically when noise is present, as shown in Fig.~\ref{fig:Rob2} (c)-(e).

\subsection{Complexity}
 Complexity often refers to computational speed and storage demand. For SBGP and LBP, the computational speed depends on the number of binary correlations and the number of resulting (principal) binary numbers (labels), both of which are determined by the number of neighbors/directions. The computation of Gabor features depends on the number of convolutions, applying multiple Gabor kernels with various scales and orientations for a set of local pixels. So, the speed is determined by the numbers and sizes of Gabor kernels. The storage demands of these three descriptors are measured by the feature dimensions. As mentioned, the dimensions of local histogram based features are computed by multiplying the numbers of labels (bins) and blocks.

In this comparison, the numbers of neighbors were set to their maximum numbers with respect to radii for both binary descriptors, $P=8R$, e.g. ($8,1$), ($16,2$) and ($24,3$). The Gabor faces were run by using their typical parameter setting with kernel size of $31 \times 31$,  eight different orientations and five various scales \cite{Liu2002}. We computed the average running time per face and the feature dimensions, together with the numbers of computational units for a given pixel and the numbers of labels generated by binary descriptors. The experimental results on the AR and YaleB face databases (5053 faces in total) are given in Table~\ref{tab:complexity}. The image size was $100 \times 100$ and the numbers of blocks used by LBP and SBGP were the same as, 36. The experiments were run on a typical PC with AMD Dual Core processor of 2.2GHz and RAM of 2.0GB. SBGP was run by our unoptimized MATLAB code. LBP code was from the authors of \cite{Ojala2002,Ahonen2006} (also in MATLAB). Gabor representation was run based on the MATLAB codes of \cite{Yang2010, Liu2002}, which integrates the C/C++ codes for computing the convolutions.
\begin{table}[tb]
\footnotesize
\centering \caption{Complexity of SBGP, LBP and Gabor features.}
\begin{tabular}{l|c|c|c|c}

\hline

Descriptors       & $\sharp$ comp. units & time(s) &$\sharp$ labels & $\sharp$ dimensions\\
\hline
&\multicolumn {4}{c}{}\\

\hline
Gabor  & 38440   & 0.9699 & - & 400000\\
\hline

&\multicolumn {4}{c}{$(P,R)=(24,3)$}\\\hline
$LBP^{u2}$     &48&0.0189    &555       &19980\\ \hline
$LBP^{riu2}$   &48&0.0188   &26           &936 \\ \hline
$SBGP$         &\textbf{24} &\textbf{0.0097}       &$\textbf{24}$            &\textbf{864} \\ \hline

&\multicolumn {4}{c}{$(P,R)=(16,2)$}\\ \hline
$LBP^{u2}$        &32&0.0128    &243       &8748 \\ \hline
$LBP^{riu2}$      &32&0.0124   &18           &648  \\ \hline
$SBGP$            &\textbf{16}&\textbf{0.0065}       &$\textbf{16}$            &\textbf{576}  \\ \hline

&\multicolumn {4}{c}{$(P,R)=(8,1)$}\\ \hline
$LBP^{u2}$     &16&0.0056    &59        &2124  \\ \hline
$LBP^{riu2}$   &16&0.0055   &10           &360    \\ \hline
$SBGP$         &\textbf{8} &\textbf{0.0032}       &$\textbf{8}$            &\textbf{288}\\ \hline
\end{tabular}\label{tab:complexity}
\end{table}


As can be seen, the complexities of the two binary descriptors are significantly lower than that of Gabor. The ratios of computational cost, execution time and final feature dimension between Gabor feature and basic SBGP descriptor are about 4800:1, 300:1 and 1400:1, respectively. Compared to LBP, the costs and execution times of SBGP are about half of that of LBP in all resolutions. Running a basic SBGP operator on a regular face image takes only $0.0032s$, making it applicable to real-time applications. Furthermore, SBGP uses even fewer pattern labels than the LBP descriptor, i.e. much lower dimensionality of its features. For example, the dimensions of SBGP features are only about 13.6\%, 6.6\% and 4.3\% of $LBP^{u2}$ in three spatial resolutions. Comparing to the CS-LBP \cite{Heikkila2009}, in the ($24,3$) case, the number of the CS-LBP dimension is increased to $2^{12} \times 36 \approx 1.5 \times 10^5$, which is more than 170 times of our SBGP. Hence, the SBGP descriptor is extremely efficient and compact.

\section{SBGP on Orientational Image Gradient Magnitude}

%

\begin{figure}
    \begin{center}
    \includegraphics[width=9.5cm,height=5cm]{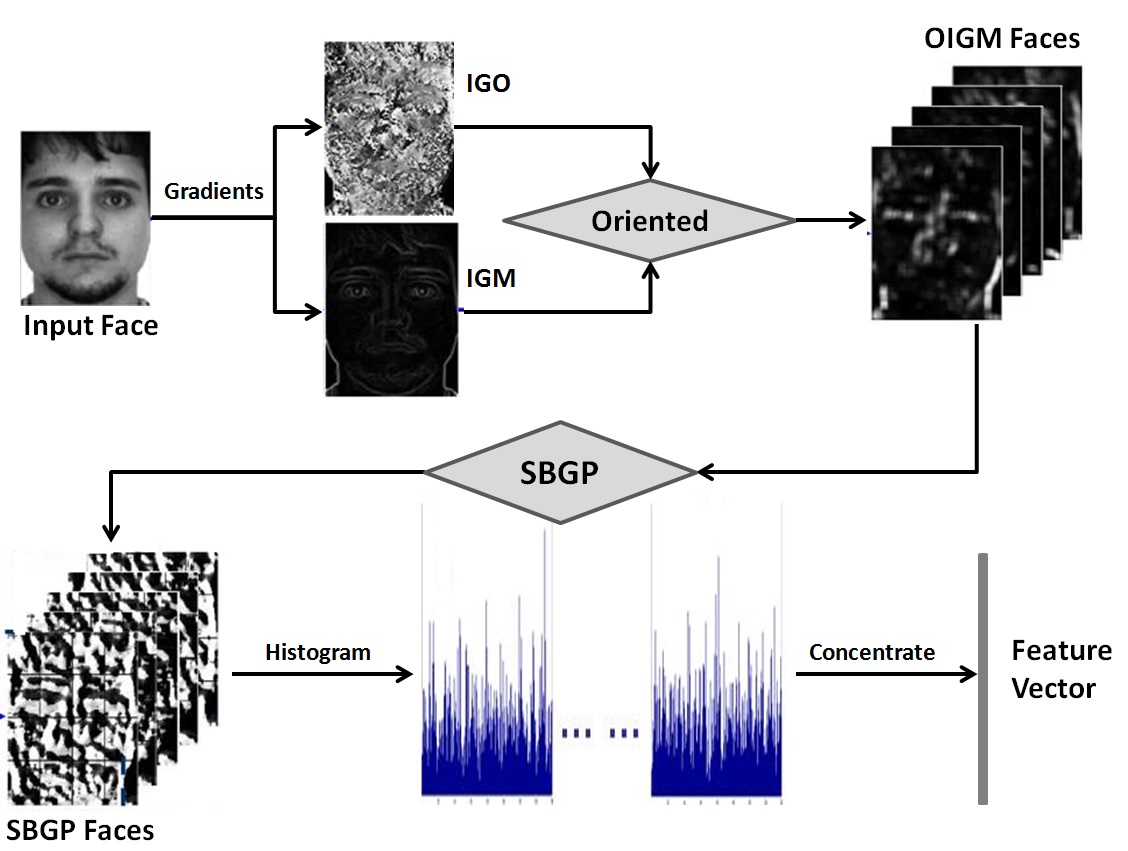}

    \end{center}
\caption{Framework of SBGPM descriptor.}\label{fig:SBGPM}
\end{figure}

Various extensions of Gabor and LBP representations have helped to yield some state-of-the-art local facial representations, for example by combining both properties and enforcing spatial locality, orientation and robustness, e.g. \cite{Zhang2005, Zhang2007, Lei2011, Tan2011, Chen2010, Vu2012}. Improving on these methods, we propose a framework by applying the SBGP descriptor on orientational IGM (OIGM), abbreviated as SBGPM, to enhance the discriminative power by further enforcing spatial locality and orientation. The framework of SBGPM is depicted in Fig.~\ref{fig:SBGPM} and its details given in the following steps:

1). Compute IGO and IGM from a given face image.

2). Generate a set of OIGM images from IGO and IGM. Similar to the orientations computed by WLD \cite{Chen2010}, IGO is first quantized into a number of dominant orientations. Then, an OIGM image, corresponding to a certain dominant orientation, is generated by computing the average values of IGMs of this dominant orientation in defined neighborhoods (e.g. with sizes of $7 \times 7$, referred as local resolution of OIGM),
\begin{eqnarray}
\overline{M}_{i,j}^t= \frac{1}{n}\sum_{k \in \Omega_{i,j}^t} M_k \quad t=1,2,\ldots,s
\end{eqnarray}
where $(i,j)$ is the location index of the given pixel, $M_k$ is the IGM value in location $k$, representing the 2D index such as $(i,j)$. $\Omega_{i,j}^t$ is a set of location indices corresponding to the $t$-th dominant orientation in the defined neighborhood of the given pixel. $n$ is the number of pixels in this neighborhood, e.g. $n=7 \times 7=49$.

3). Run SBGP on the OIGM images to yield a set of SBGPM images, on which local histogram is computed to generate the final feature vector.

In the SBGPM framework, the strength of edge information is enforced by using IGM image instead of the intensity image. It produces stronger orientational power by generating the OIGM images from different discrete dominant orientations, and further enforce spatial locality by computing the average IGM values in a certain local resolution. Effectively, SBGPM gains greater discriminant ability from these enhancements, while allowing an acceptable increase in complexity. Fortunately, SBGPM often achieves high performance in generally low complexity, different from the Gabor representation that would require a large number of Gabor faces (e.g. typically 40) and a large convolution neighborhood (e.g. $31 \times 31$). From our experiments, the typical number of OIGM and its local resolution are 3 and $7 \times 7$, respectively, leading to only 147 additional computational units (less than $0.4\%$ of Gabor faces) and 3 times of  dimensions (compared to 40 times of Gabor based fusion models).

\section{Robust Face Recognition}
%

We systematically evaluated the performance of SBGP based descriptors for facial representation and their robustness against multiple variations such as changes of lighting, expression, occlusion and age. Two groups of experiments were conducted. First, the performance of the SBGP was compared to the basic LBP and Gabor features, together with discussions on parameter selections. Second, the capability of the SBGP and SBGPM descriptors was further evaluated by comparing with recent methods on four publicly available databases: the AR \cite{Martinez1998, Martinez2001}, (extended) YaleB \cite{Georghiades2001, Lee2005}, FERET\cite{Phillips2000} and Labeled
Faces in the Wild (LFW) \cite{Huang2007} databases. For unbiased comparisons and reliable results, all implemented methods operated directly on the raw face images without any pre-processing, such as DoG filtering, Gamma correction and lighting equalization, some of which may prominently affect the experimental results.

\begin{figure}
\centering
\subfigure[]{     \includegraphics[height=1.25cm]{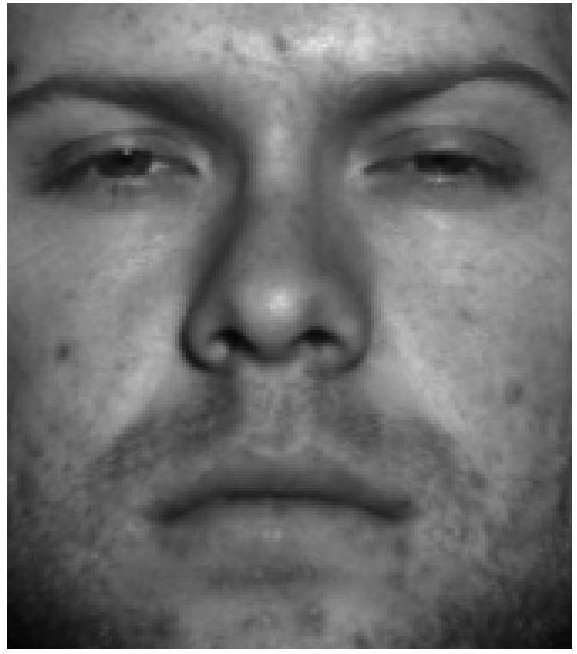}}
\subfigure[]{     \includegraphics[height=1.25cm]{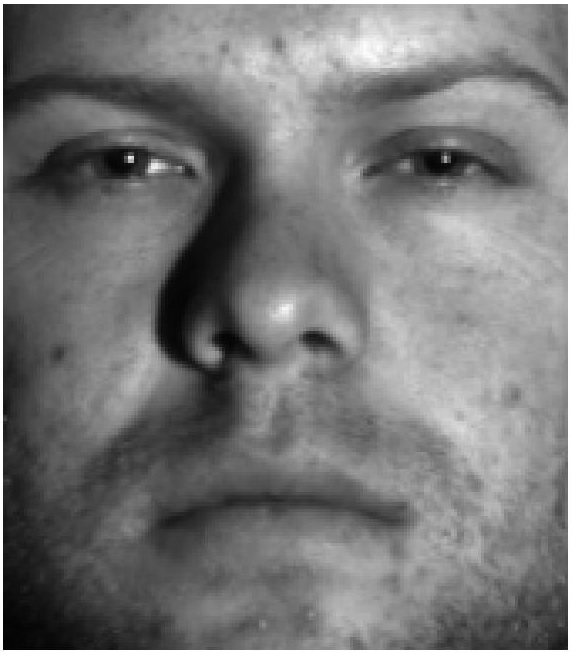}}
\subfigure[]{     \includegraphics[height=1.25cm]{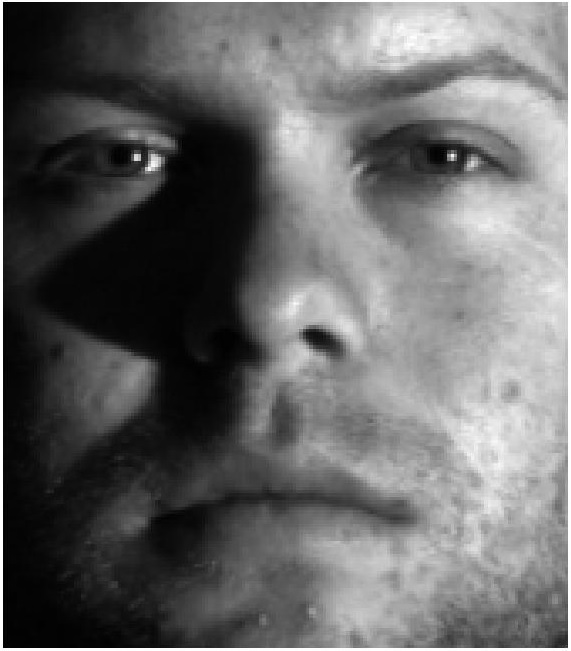}}
\subfigure[]{     \includegraphics[height=1.25cm]{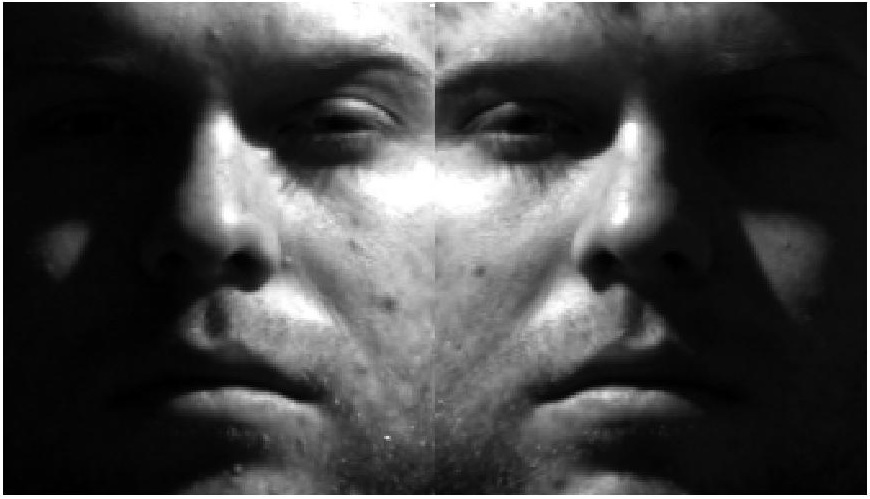}}
\subfigure[]{     \includegraphics[height=1.25cm]{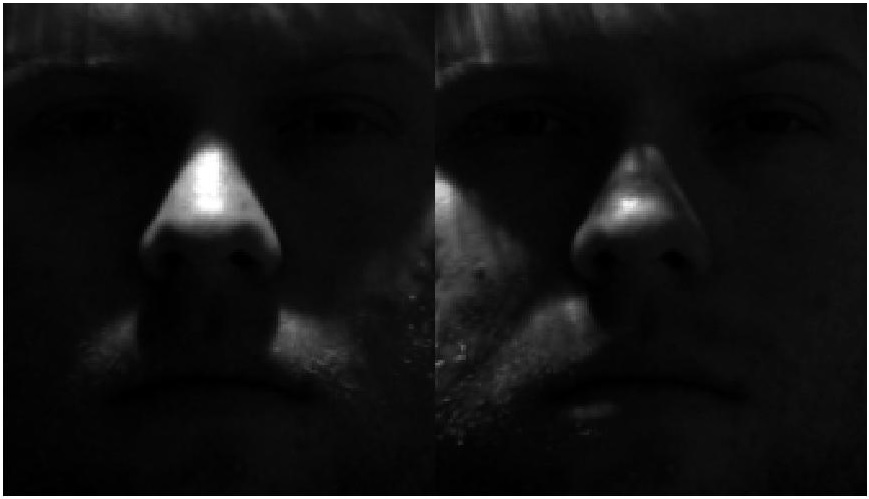}}\\

\subfigure[]               {     \includegraphics[height=1.5cm]{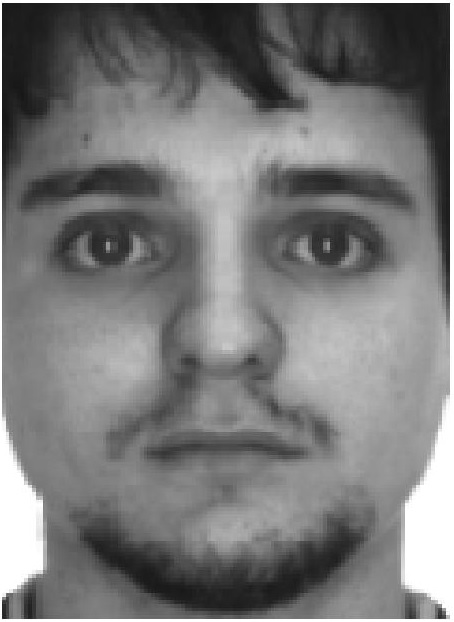}}
\subfigure[]           {     \includegraphics[height=1.5cm]{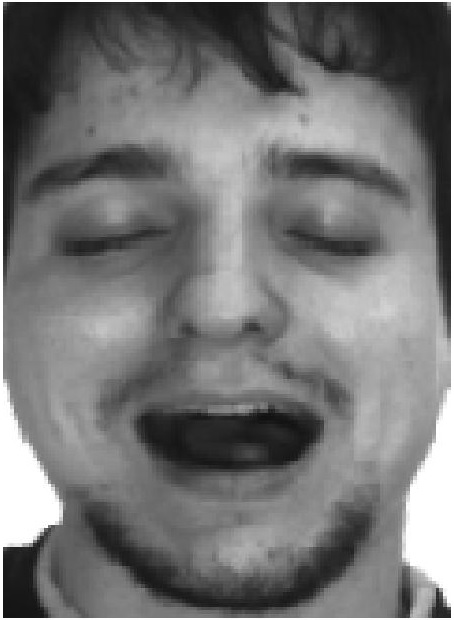}}
\subfigure[]             {     \includegraphics[height=1.5cm]{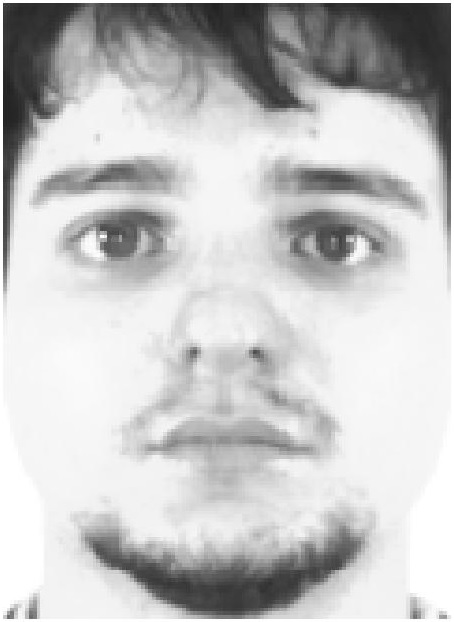}}
\subfigure[]{     \includegraphics[height=1.5cm]{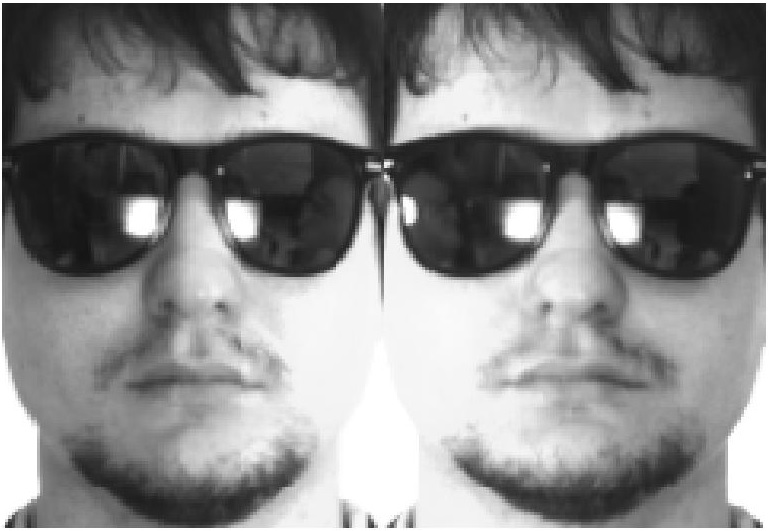}}
\subfigure[]    {     \includegraphics[height=1.5cm]{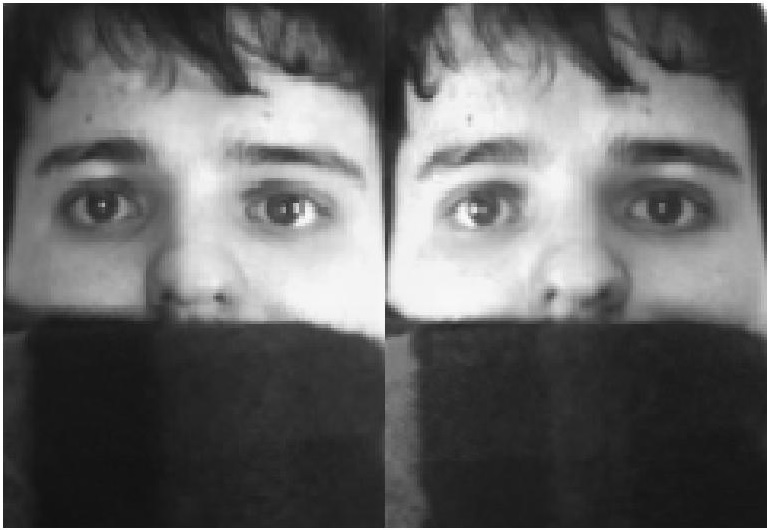}}
\caption{Face examples: (a)-(e) group 1-5 of the YaleB faces; (f)-(j) AR faces with Nature, Expression, Lighting Sunglass $\&$ Lighting and Scarf $\&$ Lighting.}\label{fig:face_examples}
\end{figure}


1). The YaleB database contains about 22000 face images of 38 subjects with 9 different poses and 64 illumination conditions for each subject. A widely used subset\cite{Georghiades2001, Lee2005}, which includes all faces from the frontal pose  ($64 \times 38 $ = 2432), was exploited in the experiments for testing the robustness to illumination variations. The dataset was divided into five different groups with increasing effect of illumination, according to \cite{Lee2005}. Exemplar faces are shown in Fig.~\ref{fig:face_examples}.


2). The AR database consists of over 4000 images of 126 subjects,
each having 26 facial images taken in two different sessions separated
by two weeks. Each session has 13 images with multiple variations in
expression, lighting and occlusion (sun glasses and/or scarf). A
subset of cropped faces (by its original authors\cite{Martinez2001}) of 50 male and 50 female subjects was used in the experiments. Examples on these variations are shown in Fig.~\ref{fig:face_examples}.

3). The FERET database\cite{Phillips2000} has five subsets, including a gallery set (Fa) and four probe sets (Fb, Fc, DupI and DupII). The gallery set contains 1196 frontal images of 1196 subjects. The DupI and Dup II sets, including 722 and 234 face images respectively, have been proven extremely challenging due to significant appearance variations caused by aging. Our experiments were conducted on both challenge sets. Following most existing methods, we cropped the original images into smaller sizes ($140 \times 120$) according to the available eye's coordinates, but without any further pre-processing.


4). The LWF dataset\cite{Huang2007} contains 13233 natural face images of 5749 people, collected in unconstrained environments from the web. They have large real-world variations in expression, lighting, pose, age, gender and even image scale and quality. The evaluation followed the standard\emph{ image-restricted test model}, verifying whether a pair of faces are from a same person. Our methods were evaluated on the widely used View 2 set, containing ten non-overlapping subsets, each having 600 pairs of images (300 matching and 300 non-matching pairs). Following the previous work\cite{Wolf2011,Vu2012}, we simply cropped out the main face area of size $150 \times 80$ from the images provided by Wolf \emph{et al}\cite{Wolf2009}.

\subsection{SBGP, LBP and Gabor Representation}

\begin{figure*}
\centering
\subfigure[Group three]{     \includegraphics[height=3cm,width=4cm]{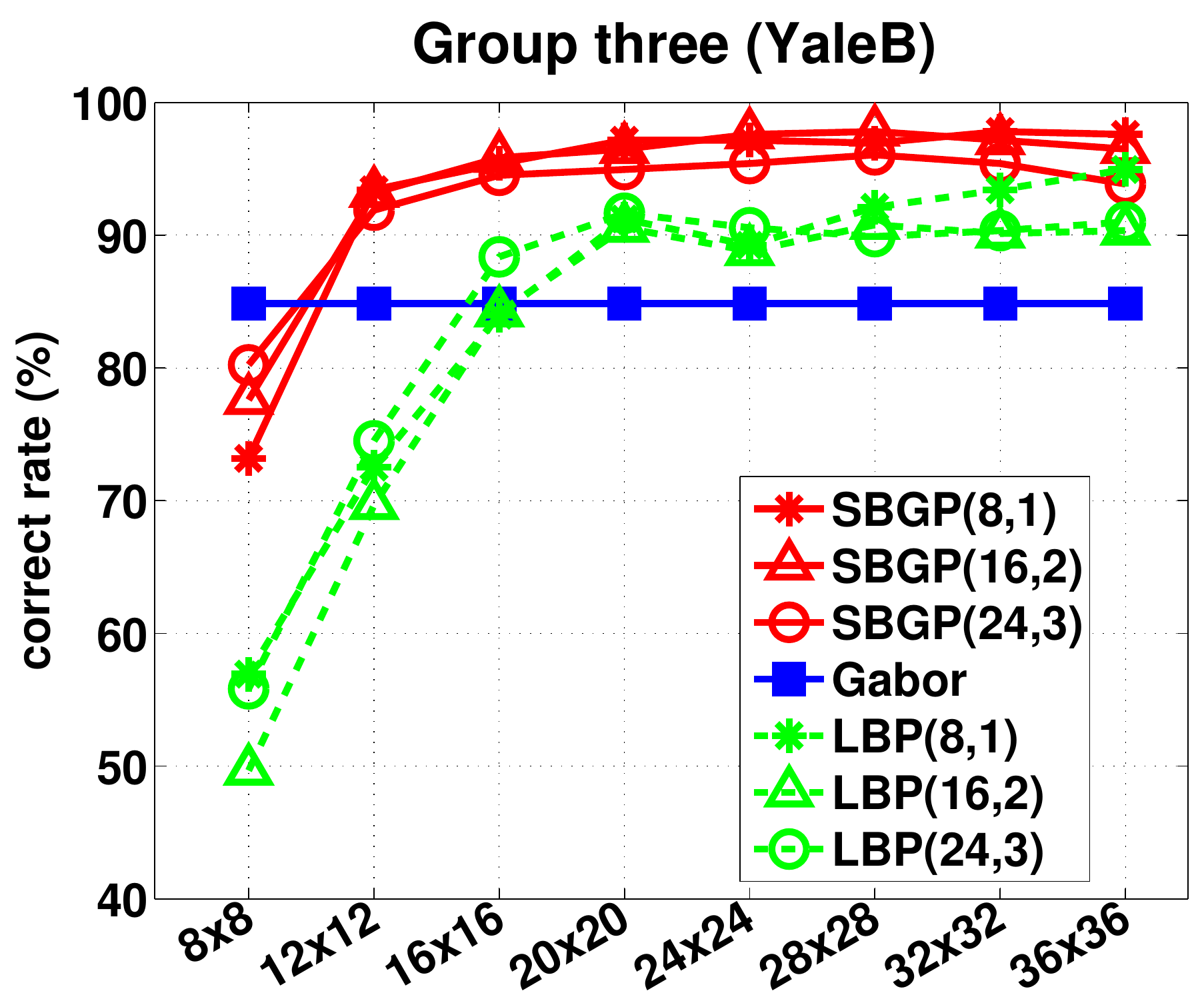}}
\subfigure[Group four]{     \includegraphics[height=3cm,width=4cm]{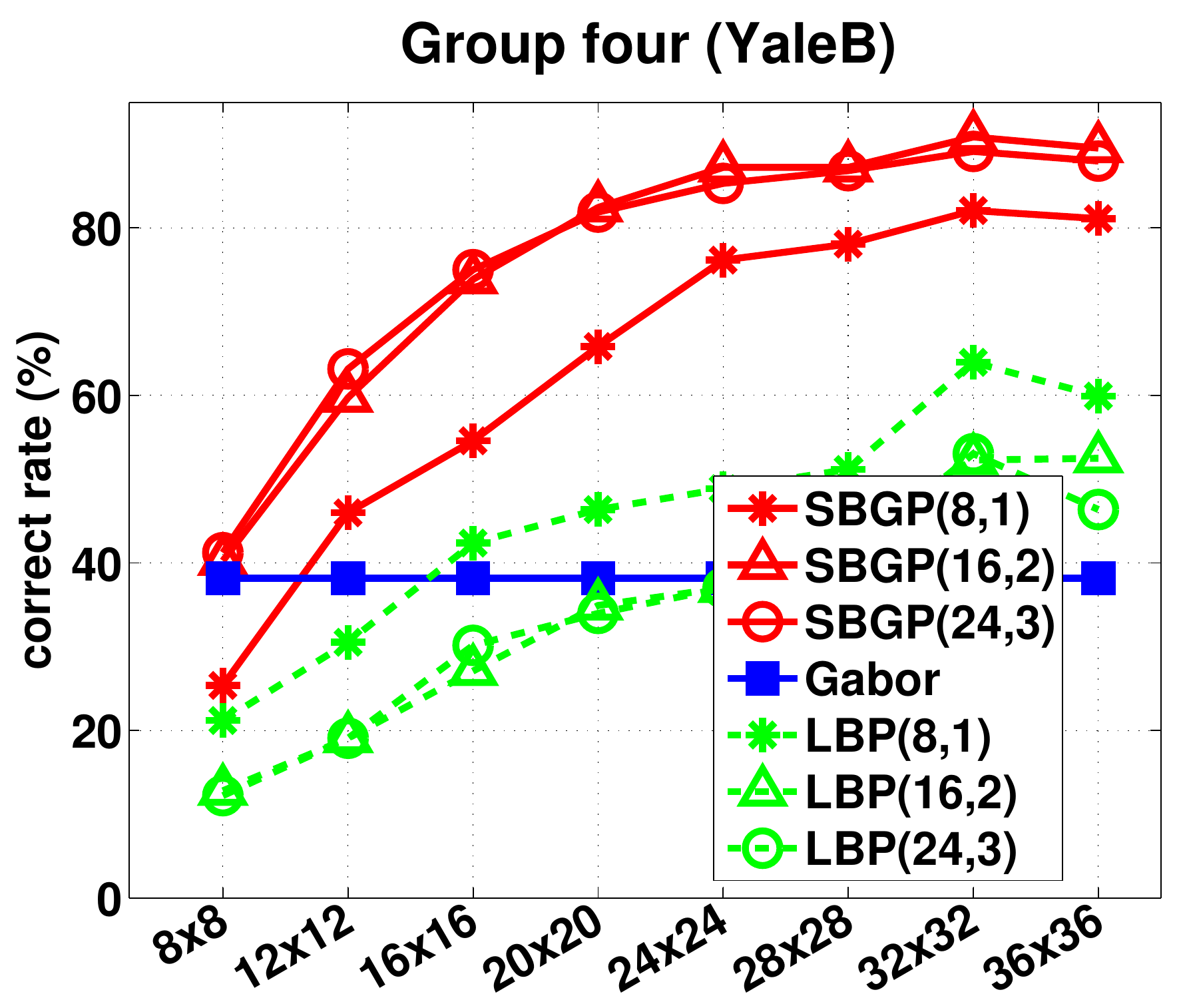}}
\subfigure[Group five]{     \includegraphics[height=3cm,width=4cm]{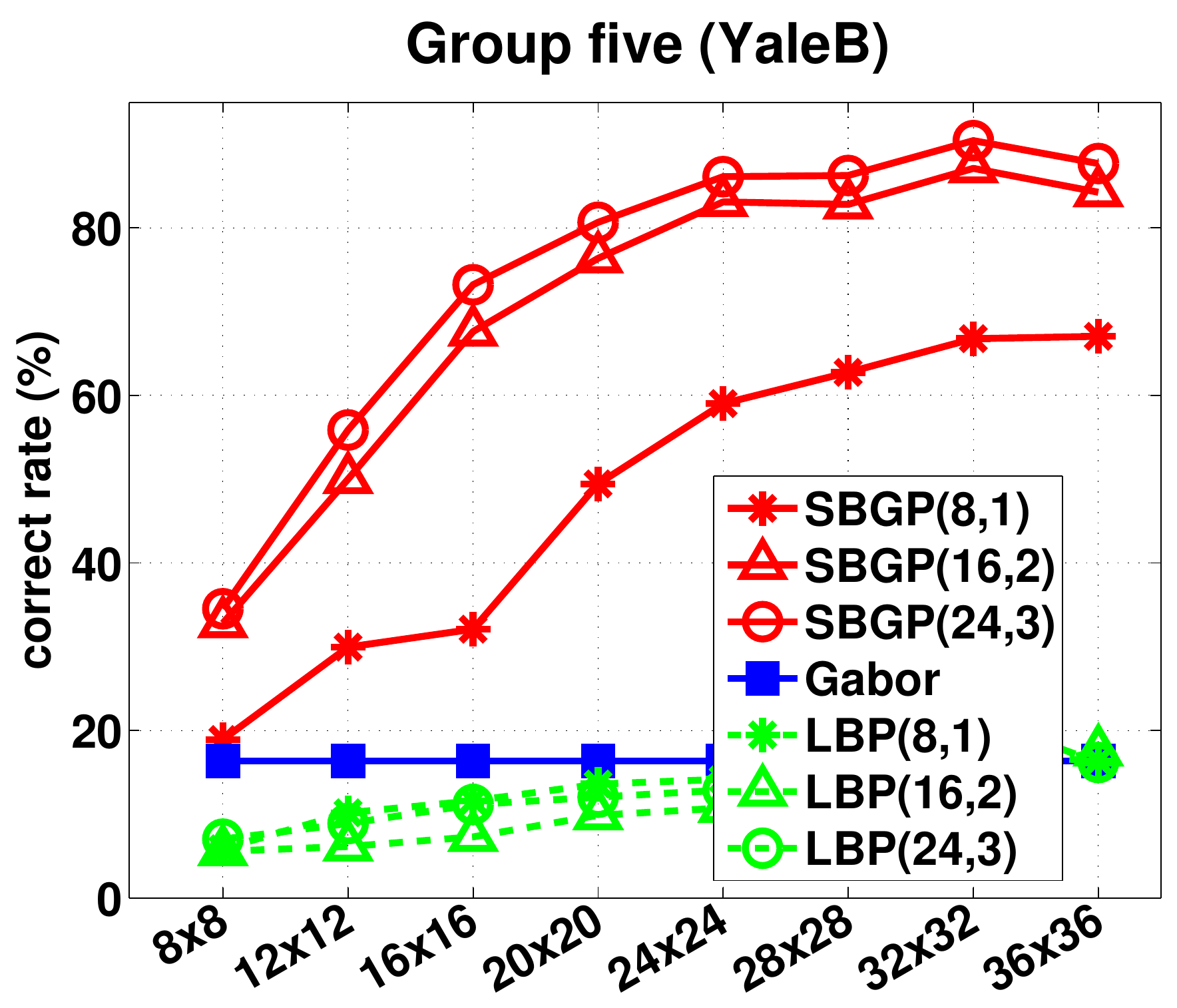}}\\

\subfigure[Expression]{     \includegraphics[height=3cm,width=4cm]{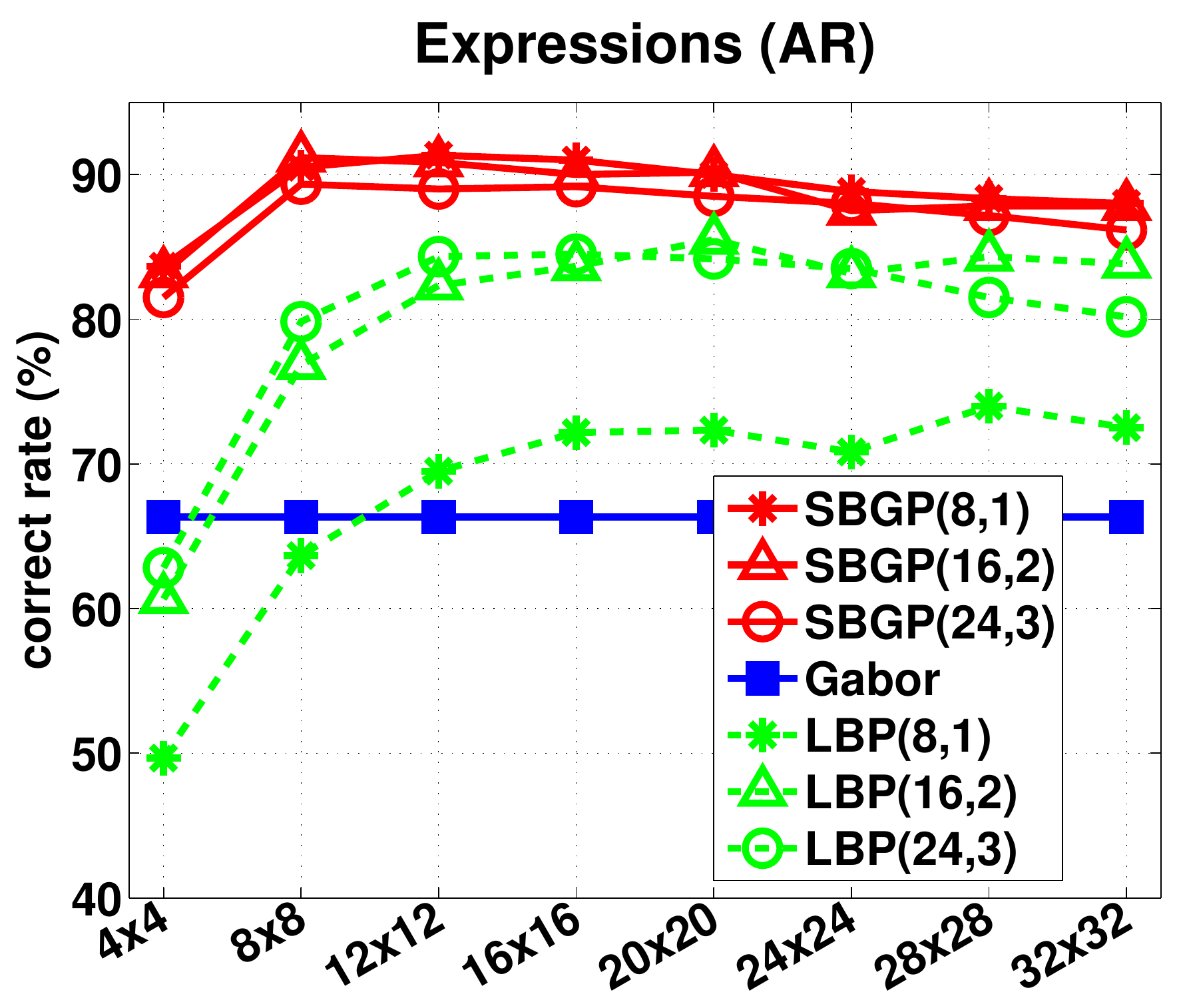}}
\subfigure[Lighting]{     \includegraphics[height=3cm,width=4cm]{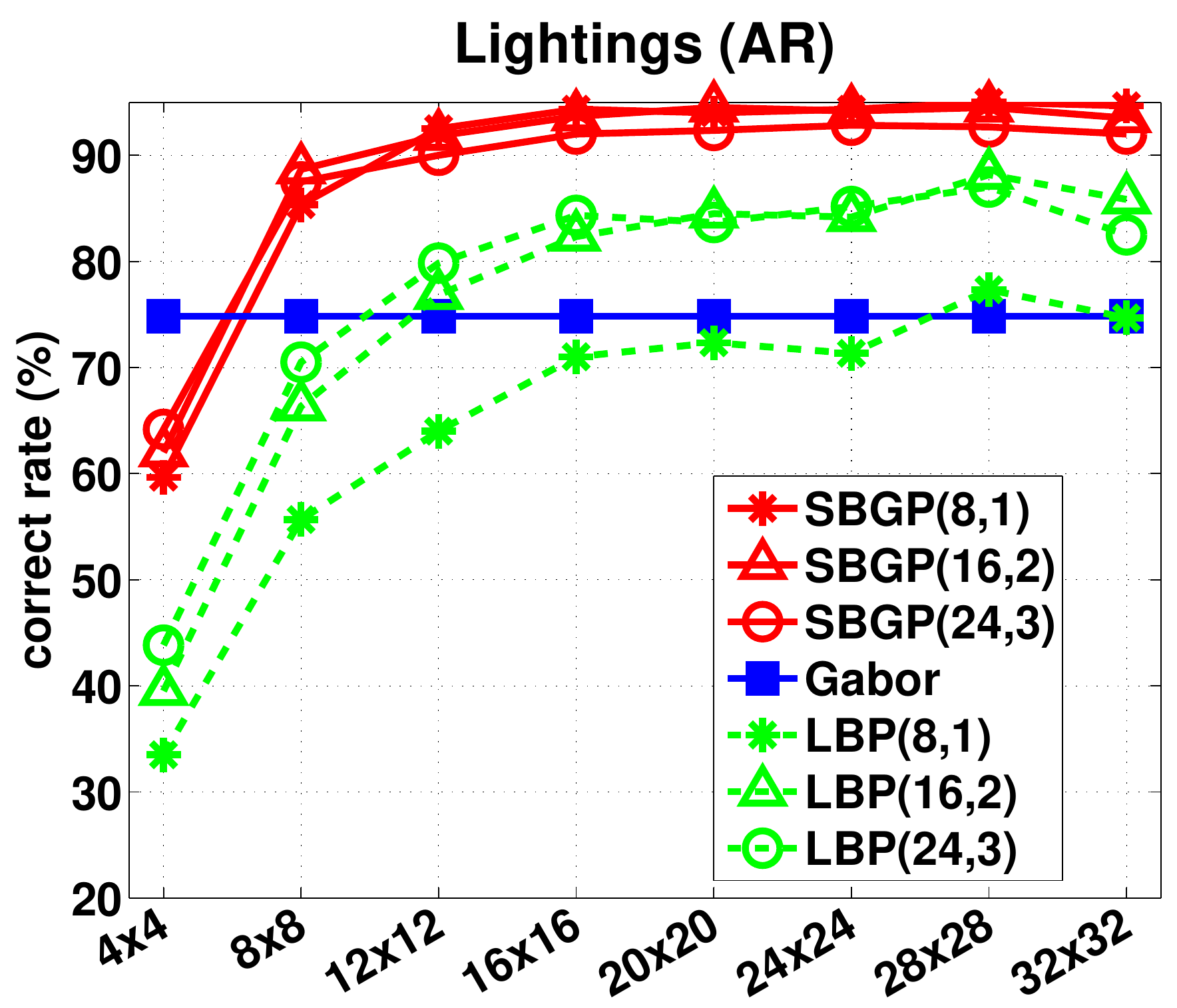}}
\subfigure[SG $\&$ Lig.]{     \includegraphics[height=3cm,width=4cm]{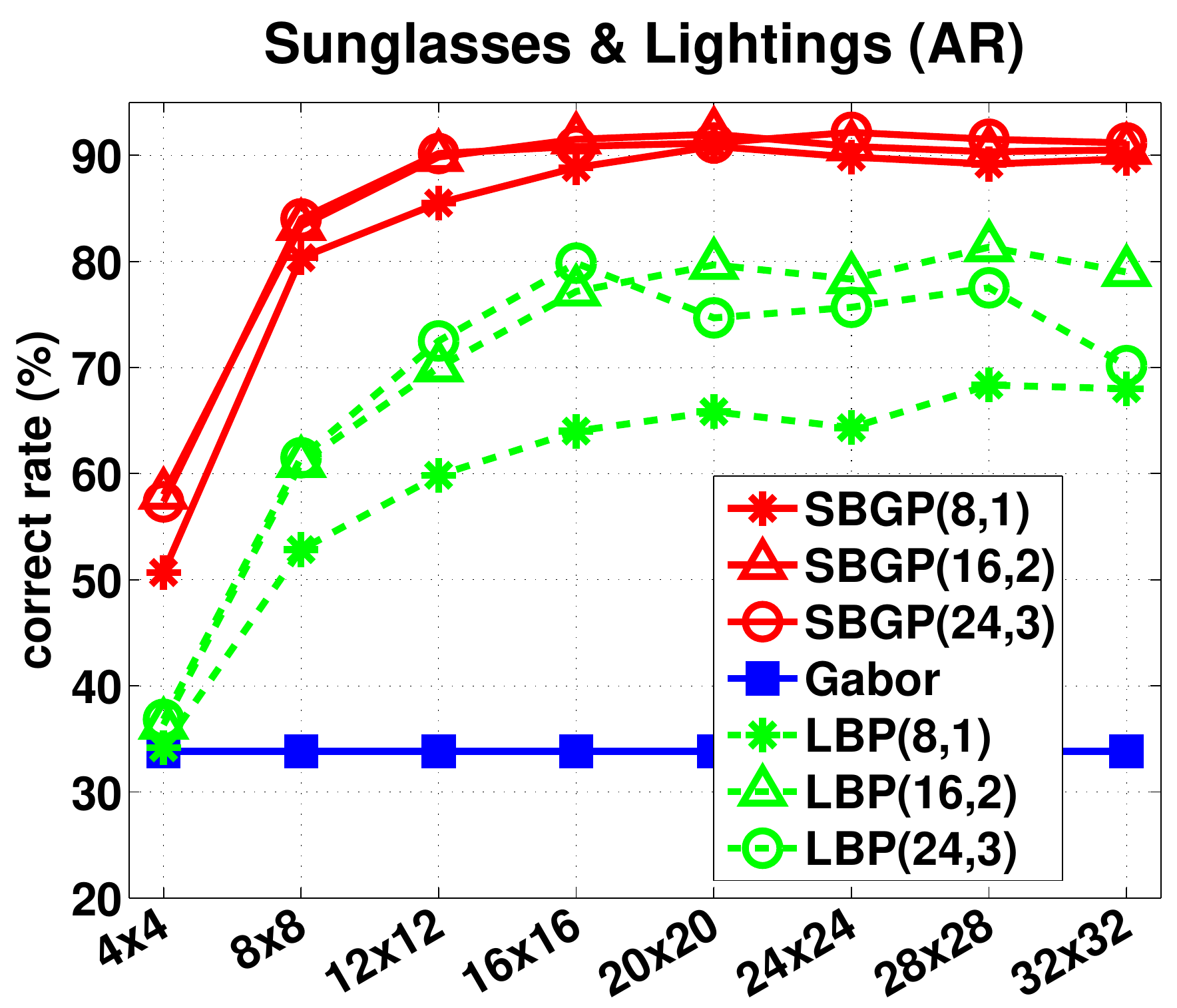}}
\subfigure[SC $\&$ Lig.]{    \includegraphics[height=3cm,width=4cm]{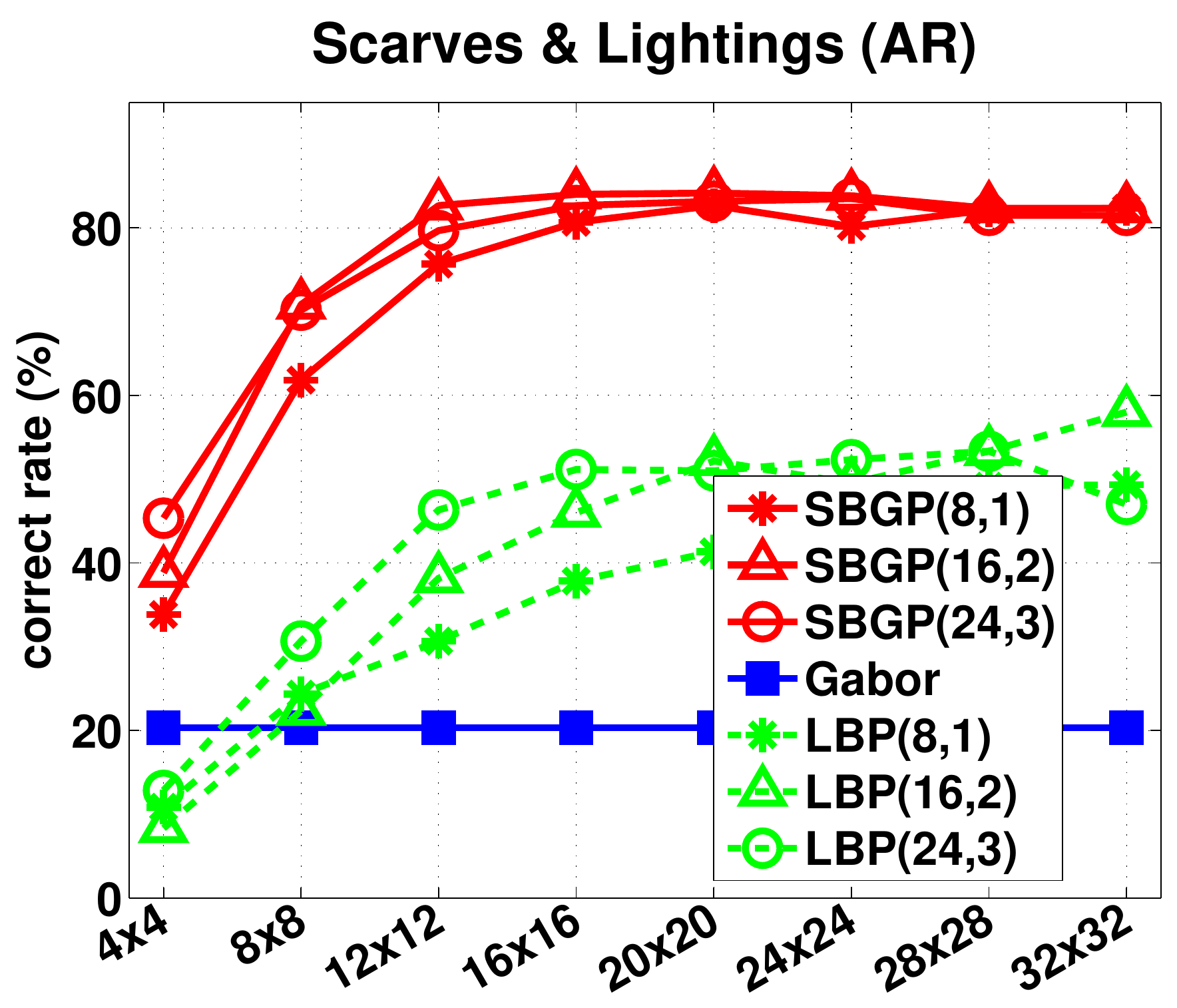}}
\caption{Performance of SBGP, LBP and Gabor on YaleB (top) and AR (bottom).}\label{fig:results_elementary_YaleBandAR}
\end{figure*}

We investigated the performance of three fundamental descriptors for face recognition. The proposed SBGP and the rotational invariant uniform LBP \cite{Ojala2002} were implemented in local histogram model as \cite{Ahonen2006}. There are only two parameters for both methods, spatial resolution, ($P,R$), and number of blocks, $N_{blk} \times N_{blk}$. The nearest neighbor (NN) classifier was used to assign the label of the most similar gallery image to the probe face. Similarities between feature vectors were computed by histogram intersection for LBP and SBGP, and by Euclidean distance for Gabor representation\cite{Liu2002}.

Their performances were evaluated on the YaleB and the AR databases. On the YaleB, a single face with natural illumination condition ("A+000E+00") per subject was used as gallery image. All five groups with different levels of illumination effects were tested. For the AR, the gallery images were the natural faces from the session one (also a single face per subject), and the probe images were grouped as expression, lighting, sunglass $\&$ lighting and scarf $\&$ lighting, each including 600 images. The results are presented in Fig.~\ref{fig:results_elementary_YaleBandAR}.

The recognition rate of all three methods reached 100$\%$ in groups one and two of the YaleB database. Fig.~\ref{fig:results_elementary_YaleBandAR} (top row) shows that, LBP and Gabor descriptors had similar performances, which were reasonable in group three with medium level of illumination effects but deteriorated drastically with increased illumination effects in groups four and five. In contrast, SBGP were consistently excellent (with recognition rate above $90\%$) even with the extreme lighting conditions (group five). The improvements of SBGP in groups four and five were highly significant, outperforming LBP or Gabor by more than $30\%$ and $70\%$, respectively. This demonstrates its enhanced robustness against illumination variation by learning local textural structures from the image gradient domain, which inherently contain gray-scale invariant features.

Similarly, the performance of SBGP was the best in all test groups on the AR database (shown in Fig.~\ref{fig:results_elementary_YaleBandAR} (bottom row)). It gave more than $90\%$ recognition rates in the groups of expression, lighting, and sunglass $\&$ lighting, and above $80\%$ for the groups seriously affected by both large-scale occlusions and illuminations. Again, the performances of LBP and Gabor were substantially affected by multiple variations in the last two groups. The excellent results of SBGP show that it is highly discriminative and robust to multiple facial variations.

It has also been found that the SBGP descriptor is fairly insensitive to the choice of its parameters. First, the overall performance of SBGP is stable in different spatial resolutions, especially for $(16,2)$ and $(24,3)$. By contrast, changes in spatial resolution cause large differences in recognition rate of the LBP. Second, the performance of both descriptors can be improved by increasing the number of blocks. As can be seen, in most test groups, the recognition rates of SBGP become stable when the number of blocks is equal or greater than $12 \times 12$, except for groups four and five of the YaleB, which require larger numbers of blocks to alleviate the effect of severe illumination conditions. Therefore, by trading off performance and computational complexity, the spatial resolution of the SBGP was set to $(16,2)$ in all our experiments, while the numbers of blocks were determined by the sizes of images.

\subsection{Lighting and Multiple Variations}

The efficiency of the proposed SBGP based descriptors was further evaluated by comparing with recent methods, including IGO based methods (Gradientfaces\cite{Zhang2009b} and IGOPCA\cite{Tzimiropoulos2012}), local feature methods (CS-LBP \cite{Heikkila2009}, WLD\cite{Chen2010,Hermosilla2012}, LTP\cite{Tan2011}, POEM\cite{Vu2012} and Volterrafaces\cite{Kumar2012}), and fusion of both (LGOBP \cite{Liao2009} and PHOG\cite{Bosch2007}), along with recent results directly quoted from related literature.

For SBGPM, the number of OIGM and local resolution were optimally set to 3 and $7 \times7$ in all experiments. For a fair and unbiased comparison, all implemented methods employed their optimal parameters and similarity measures suggested by the original authors. IGOPCA verified the number of reduced dimensions from 10 to its maximum number. The implementation of WLD was suggested by \cite{Hermosilla2012}, with the number of quantized orientations set to 8 and differential excitation value varied among $\{32, 48, 64\}$. The CS-LBP was implemented with 8 neighbors with radius of 2 and 0.01 as binary threshold, as suggested in \cite{Heikkila2009}. The number of bins for local IGO based methods (PHOG and LGOBP) was verified among $[4, 40]$. All local histogram methods (SBGP, SBGPM, CS-LBP, WLD, LTP, POEM, and LGOBP) were run by varying the numbers of blocks from $8 \times 8$ to $36 \times 36$. The pyramid level for the PHOG was optimized from 1 to 5. Finally, the best performance of each method, computed on three widely-used similarity measures: histogram intersection, $\chi^2$\cite{Tan2011} and Euclidean distance, was reported. The SBGP methods used histogram intersection.


\subsubsection{Illumination Variation}
This experiment evaluated the illumination invariant property of SBGP-based methods by exploiting the same gallery and probe images as in the previous experiment on the YaleB database. To provide more comprehensive results, the comparisons also included a group of methods based on the reflectance model specially developed to address the illumination effect. These methods include the logarithm total variation (LTV) model \cite{Chen2006}, logarithmic wavelet transform (LWT)\cite{Zhang2009a}, applying Multi-linear Principal Component Analysis on tensors-CT histograms (TCT-MPCA)\cite{Ruiz2009} and the reconstruction with normalized large- and small-scale feature images (RLS)\cite{Xie2011}. The results are presented in Table~\ref{tab:results_YaleB}.

\begin{table}[tb]
\footnotesize
\centering \caption{Performance of single training sample per person on YaleB database.}
\begin{tabular}{l|c|c|c|c}

\hline
\multirow{2}{*}{Method}  & \multicolumn {4}{c}{Error Rate ($\%$)}\\
\cline{2-5}
                          &Group 3     &Group 4    &Group 5  &Avg.      \\\hline \hline
LTV \cite{Chen2006}       &21.5        &24.2       &17.6     &20.7      \\\hline
LWT \cite{Zhang2009a}     &18.0        &18.0       &29.2     &22.7      \\\hline
TCT-MPCA\cite{Ruiz2009}   &5.3         &39.9       &- -      &23.9      \\\hline
RLS\cite{Xie2011}         &14.0        &14.7       &15.2     &14.7      \\\hline\hline
LGOBP                     &13.4        &48.3       &67.9     &47.3      \\\hline
WLD                       &\textbf{1.1}&15.3       &60.5     &30.6      \\\hline
LTP                       &2.4         &16.2       &39.5     &22.4      \\\hline
CS-LBP                    &4.4          &39.9       &85.7     &49.3      \\\hline
PHOG                      &6.4         &54.2       &77.9     &51.5     \\\hline
POEM                      &4.8         &11.3       &40.4     &21.9      \\\hline
Volterrafaces             &6.6         &32.3       &17.4     &19.3      \\\hline
Gradientfaces             &8.4         &12.6       &17.2     &13.4      \\\hline
IGOPCA                    &10.6        &12.0       &28.2     &18.5      \\\hline\hline

SBGP                      &2.8         &9.2        &12.9     &9.1       \\\hline
SBGPM                     &1.3 &\textbf{1.9}&\textbf{2.7} &\textbf{2.1} \\ \hline
\end{tabular}\label{tab:results_YaleB}
\end{table}

As one can see, IGO based methods yielded better overall performance than the intensity based methods (local features and reflectance models). As expected, SBGP based methods achieved the best performance in all groups. Even the basic SBGP outperformed all other methods, and the SBGPM had the lowest average error rate at only $2.1\%$, which is about one tenth of the errors of other methods. Large improvements were in the severe illumination conditions, groups four and five, in which only $1.9\%$ and $2.7\%$ errors occurred, respectively. The best performances of reflectance models and IGO methods were about $12\%$ in group four and around $15\%$ for group five. The improvements are statistically significant, showing the exceptional robustness of SBGPM against illumination.

Furthermore, WLD and LTP, extended from LBP, yielded low error rates in the low and medium levels of illumination change, groups three and four. But their errors increase drastically in the extreme illumination conditions of group five. Note that the CS-LBP is not robust to both medium and extreme illumination conditions. This may be dueo to partly by its noise patterns, and partly by threshing the intensity differences, which was originally developed to achieve the robustness on flat image regions \cite{Heikkila2009}. By contrast, the proposed SBGP methods consistently excelled in extreme illumination conditions. These results, along with the previous experiments, further verify that gray-scale invariance achieved in the gradient domain by discarding gradient contrast is stronger than that realized by discarding intensity contrast.



%

\subsubsection{Multiple Variations}

The robustness against multiple variations were analyzed on the AR database. The experiments were divided into two groups of different training schemes: \emph{a single training sample per person} and \emph{multiple training samples per person}.


\medskip

\emph{EXP I: A Single Training Sample Per Person}

\medskip
We used a neutral face per subject (N) in the first session as the gallery image and tested all other faces, 4 remaining groups in session one (E, L, GL and SL) and 5 groups in session two (N, E, L, GL and SL). Results are presented in Table.~\ref{tab:results_AR}. Recently published results achieved by the same experimental scheme, such as DMMA\cite{Lu2013} and ESRC-Gabor\cite{Deng2012}, are also included for comparison. These two methods build learning models on the local features.

\begin{table}[tb]
\footnotesize
\centering \caption{Performance of single training sample per person on AR database.}
\begin{threeparttable}
\begin{tabular}{l|c|c|c|c|c|c}

\hline
\multirow{2}{*}{Method} & \multicolumn {6}{c}{Error Rate ($\%$)}\\
\cline{2-7}

                       &$N$    &$E$   &$L$   &$GL$   &$SL$ &Ave.\\\hline \hline
&\multicolumn{6}{c}{session one}\\\hline
UP\cite{Deng2010}          &-      &18.0    &-     &-      &-    &- \\\hline

DMMA\cite{Lu2013}         &-      &13.0    &-     &-      &-    &- \\\hline

ESRC-G\cite{Deng2012}\tnote{a} &-      &5.8   &0.0     &\multicolumn {2}{c|}{7.1}  &5.0\\\hline

LGOBP                     &-      &12.3  &8.7  &10.7   &65.0 &24.2 \\\hline

WLD                       &-      &3.0   &6.0   &3.3    &7.3  &5.4\\\hline

LTP                       &-      &2.7   &1.3   &2.0      &9.7  &3.9\\\hline

CS-LBP                    &-      &4.0  &1.0   &1.7  &8.0    &3.7\\\hline

PHOG                      &-      &4.7   &1.0   &3.3      &6.7  &3.9     \\\hline

POEM                      &-      &5.0   &0.0   &3.7      &5.3  &3.5 \\\hline

Gradientfaces             &-      &16.3  &3.0     &8.0    &22.3 &12.4\\\hline
IGOPCA                    &-      &15.3  &3.0     &9.3    &15.7 &10.8\\\hline\hline

SBGP                      &--&2.0     &1.7   &2.7    &6.7  &3.0\\\hline
SBGPM                     &--&\textbf{0.7} &\textbf{0.0}   &\textbf{1.0}  &\textbf{2.3}  &\textbf{0.9}\\\hline\hline

& \multicolumn {6}{c}{session two}\\ \hline

UP\cite{Deng2010}         &23     &39.7  &-      &-      &-     &- \\\hline
DMMA\cite{Lu2013}         &12     &30.3  &-      &-      &-     &- \\\hline
ESRC-G\cite{Deng2012}\tnote{a} &-      &-     &-      &-      &-     &- \\\hline

LGOBP                      &9     &33.7  &37.7   &49.0   &85.3  &51.1 \\\hline

WLD                       &7      &20.7  &20     &24.3   &27.3  &21.9\\\hline

LTP                        &2      &18.7  &10.3   &16.0   &30.0  &17.5\\\hline

CS-LBP                    &3      &18.3  &11.3   &16.3  &25.7    &16.8\\\hline


PHOG                      &2      &22.3  &8.3    &17.0   &23.3  &16.5     \\\hline

POEM                      &2      &21.7  &8.7    &17.3   &24.3  &16.8\\\hline

Gradientfaces             &8      &36.7  &14.0   &28.3   &45.0  &29.2\\\hline
IGOPCA                    &3      &28.7  &10.3   &22.7   &34.7  &22.5\\\hline\hline

SBGP                      &3      &17.7  &9.3    &13.3   &25.0  &15.3\\\hline
SBGPM                     &\textbf{2}    &\textbf{14.7}  &\textbf{3.3}   &\textbf{13.0}    &\textbf{10.3}   &\textbf{9.7}\\ \hline

\end{tabular}
 \begin{tablenotes}
 \item [a] Only 80 subjects for test and the other 20 for training.
 \end{tablenotes}
\end{threeparttable}\label{tab:results_AR}
\end{table}

By contrast, the local feature methods outperformed IGO based holistic representations. Again, SBGP methods had the best overall performance in all implementations. SBGPM had the lowest error rates in all tests and the average error rates were less than $1\%$ and $10\%$ for sessions one and two, respectively, significantly surpassing the most closed performance at $3.5\%$ (by POEM) in session one and $16.5\%$ (by PHOG) in session two.


It can be seen from the table that the main gain of the local feature methods (e.g. POEM and LTP) over IGO methods lies in the tests of Expression(E), Scarf $\&$ Lighting (SL), both of which cause large-scale local distortions and can lead to  significant differences in performance of local feature and holistic methods. By integrating both approaches, the SBGPM has not only yielded excellent performances in single variations, but also achieved very low error rates even in the most severe cases affected by both large-scale scarf occlusion and lighting.

\medskip
\emph{EXP II: Multiple Training Samples Per Person}

\medskip
We further compared the performance of the SBGP based methods with recent local fusion models and SRC based methods by using their latest published results on the AR database, under the four different implementations used by the related publications. The implementations are described as follows and the results are shown in Table~\ref{tab:results_AR_impABCD}.

\textbf{Implementation A} followed the experimental setting of \cite{Lei2011} by using two neutral faces from both sessions as gallery images and testing all four variations including expression (E), lighting (L), sunglass $\&$ lighting (GL) and scarf $\&$ lighting (SL), each having six faces per subject. The SBGP methods were evaluated against two fusion models integrating LBP and Gabor representations. The SBGPM achieved perfect performance in the group of lighting, which were not reported in \cite{Lei2011} for the compared methods. The largest improvement lies in the group of sunglass $\&$ lighting, resulting in only $0.3\%$ error for SBGPM compared to $46.1\%$ for the best of the compared methods.

\textbf{Implementation B} evaluated the performance on variations of expression $\&$ lighting (E$\&$L) and occlusions. For E$\&$L, seven faces per subject were used for training, one neutral, three expressions and three lighting faces from session one, and tested the corresponding seven faces from session two. For occlusions, eight faces per subject including two neutral and six expression faces from both sessions, were used as gallery images, two faces of sunglasses or scarves in both sessions were tested.
The SRC-based approaches achieved low error rates for variations in expressions or lightings. However, their performances suffered seriously in large-scale occlusions such as scarves. A common remedy for mitigating this effect is to manually partition a face image into a number of regions, and discard the occluded parts. The GRRC with partitions improved the performance substantially with very low error rates of $2.7\%$, $0.0\%$ and $1.0\%$ \cite{Yang2012a}. The SBGPM consistently exceeded these, and obtained almost perfect performance with $0.3\%$, $0.0\%$ and $0.0\%$ error rates. Note that the performance of SRC approaches strongly depends on the manual partition scheme, while the SBGP methods are completely automatic.

\textbf{Implementation C} trained on seven non-occluded faces from session one (as in Implementation B) and tested on four sets of occluded faces. Each set contained three faces per subject, with sunglasses or scarves, including multiple effects by lighting, in session one or two. Four sets are indicated as GL[S1], GL[S2], SL[S1] and SL[S2] in the table.

\textbf{Implementation D} conducted on three separate experiments according to \cite{Chen2012}. The first one trained on seven non-occluded faces and one sunglass face (randomly selected from three in session one) and tested on seven non-occluded faces from session two and the remaining five sunglass faces in both sessions. The second experiment applied the similar training/test scheme for the faces with scarf occlusions. The last one evaluated both sunglass and scarf occlusions by using nine faces for training (seven non-occluded faces plus one (random) sunglass and one scarf faces from session one) and totally seventeen faces for testing, including seven non-occluded faces from session two and the remaining five sunglass and five scarf faces. The results of the SBGP methods were the average error rates of three, three and six cross-validated selections of training sets.

\begin{table}[tb]
\footnotesize
\centering \caption{Comparisons with recent local fusion models (Implementation A\cite{Lei2011}) and SRC based methods (Implementation B\cite{Yang2012a}, C\cite{Yang2012a} and D\cite{Chen2012}) from published results.}
\begin{threeparttable}
\begin{tabular}{l|c c c c}

\hline

\multirow{2}{*}{Method} & \multicolumn {4}{c}{Error Rate }\\
& \multicolumn {4}{c}{($\%$)}\\\hline\hline


\textbf{Implement. A}    &\textbf{E}  &\textbf{L}  &\textbf{GL} &\textbf{SL} \\\hline
LGBP-M \cite{Zhang2005}  &13.9    &- -     &62.4    &17.4  \\
LGBP-P\cite{Ruiz2009}    &14.1    &- -     &63.0    &16.5  \\
GVLBP-M\cite{Lei2011}    &9.4     &- -     &46.1    &12.6\\
GVLBP-P\cite{Lei2011}    &8.9     &- -     &53.9    &9.6\\\hline

SBGP                     &2.5     &0.5     &1.0     &4.7 \\
SBGPM                    &\textbf{2.2} &\textbf{0.0}&\textbf{0.3} &\textbf{1.2}  \\ \hline\hline

\textbf{Implement. B}\tnote{a}   &\textbf{E$\&$L}\tnote{b}  &\textbf{Sunglass}  &\textbf{Scarf} & \\\hline

SRC \cite{Wright2009}        &5.3     &13.0(2.5)     &40.5(6.5)  & \\
LRC \cite{Naseem2010}        &23.3    &4.0 (- -)     &74.0(4.5)  & \\
CESR\cite{He2011}            &- -     &30.0(- -)     &- - (1.7)  & \\
CRC$_{RLS}$\cite{Zhang2011}  &6.3     &31.5(8.5)     &9.5 (5.0)  & \\
RSC$_{GS.}$\cite{Theodorakopoulos2011} &10.0       &- -         &- - & \\
RCR \cite{Yang2012b}         &4.1      &- - (1.5)   &- - (3.5)   & \\
GRRC\cite{Yang2012a}         &2.7      &7.0 (0.0)   &21.0(1.0)   & \\\hline

SBGP                         &2.0      &0.5         &1.0         & \\
SBGPM                        &\textbf{0.3} &\textbf{0.0}   &\textbf{0.0} & \\ \hline \hline

\textbf{Implement. C}    &\textbf{GL[S1]}  &\textbf{GL[S2]}  &\textbf{SL[S1]} &\textbf{SL[S2]} \\\hline
SRC \cite{Wright2009}       &16.7    &51.3    &51.0    &71.0  \\
CRC$_{RLS}$\cite{Zhang2011} &22.0    &47.7    &55.3    &70.7  \\
GRRC\cite{Yang2012a}        &7.7     &48.3    &5.0     &15.7\\\hline

SBGP                        &\textbf{0.0}  &6.0             &2.3          &11.0 \\
SBGPM                       &0.3           &\textbf{2.3}    &\textbf{0.3} &\textbf{4.3}  \\ \hline\hline

\textbf{Implement. D}    &\textbf{EGL}  &\textbf{ESL}  &\textbf{EGSL} & \\\hline

SRC \cite{Wright2009}     &15.8    &23.7    &22.0   &   \\
LLC \cite{Wang2010}       &15.5    &23.4    &21.0   &   \\
LR \cite{Chen2012}        &14.6    &15.6    &18.4   &   \\\hline

SBGP                      &2.6     &3.7     &3.6    &   \\
SBGPM                     &\textbf{1.6} &\textbf{1.2}    &\textbf{1.9}&\\\hline

\end{tabular}

 \begin{tablenotes}
 \item [a] The error rates presented in parentheses were achieved by using manually partition scheme.
 \item [b] SIFT\cite{Lowe2004}, and its extension, Partial-Descriptor-SIFT, were test on this group with error rate of 6.1\% and 4.5\% in \cite{Geng2009}.
 \end{tablenotes}
\end{threeparttable}\label{tab:results_AR_impABCD}

\end{table}

Implementations C and D tested the SBGP methods on more complex variations such as sunglass $\&$ lighting, or scarf $\&$ lighting, which had been rarely evaluated by the SRC-based methods. Clearly, complex variations do not seem to hinder the extraordinary capabilities of the SBGP and SBGPM in these implementations, while the performance of SRC-based approaches suffered poorly. SBGPM again achieved extremely low average error rates, $1.8\%$  and $1.6\%$ for implementations C and D, a fraction of the average error rates of the best of the compared methods ($19.2\%$ by GRRC and $16.2\%$ by LR).

\subsection{Aging and Unconstrained Variations}
\begin{table}[tb]
\footnotesize
\centering \caption{Performance on  LWF for ageing and unconstrained variations (CR-Correct Rate).}
\begin{threeparttable}
\begin{tabular}{l|c|c||l|c}

\hline
\multicolumn {3}{c||}{FERET}&\multicolumn {2}{c}{LFW}
 \\\hline
\multirow{2}{*}{Method}     &\multicolumn {2}{c||}{CR ($\%$)}&\multirow{2}{*}{Method}&\multirow{2}{*}{CR ($\%$)}\\
\cline{2-3}
                           &DupI   &DupII    &                          &     \\\hline \hline
PS-SIFT\cite{Luo2007}       &61.0   & 53.0    & --&-- \\\hline
Gabor-WPCA\cite{Deng2005}   & 78.8  & 77.8    &V1-like\cite{Pinto2008}   &64.2 \\\hline

LBP-WPCA\cite{Vu2012}       & 79.4  & 70.0    &V1-like+\cite{Pinto2008}  &68.1 \\\hline

WLGBP\cite{Zhang2005}       & 74.0  & 71.0    & Gabor (C1)\cite{Wolf2011} &68.4 \\\hline

WHGPP\cite{Zhang2007}       & 79.5  & 77.8    &LBP\cite{Wolf2011}     &67.9 \\\hline

G-LDP\cite{Zhang2010}\tnote{b}& 78.8 & 77.8& FPLBP\cite{Wolf2011}      &67.5 \\\hline
LGBP-WPCA\cite{Nguyen2009}  & 83.8 & 81.6     &TPLBP\cite{Wolf2011}       & 69.0 \\\hline

Zou's Result\cite{Zou2007}  &85.0  & 79.5     &Comb.\cite{Wolf2011}\tnote{a}  & 74.5 \\\hline

Tan's Result\cite{Tan2007}\tnote{c}& 90.0 & 85.0 &SIFT\cite{Wolf2011}  & 69.9 \\\hline

POEM-WPCA\cite{Vu2012}\tnote{c} & 88.8 & 85.0 &POEM\cite{Vu2012} & 73.7 \\\hline
IGOPCA\cite{Tzimiropoulos2012}  & 88.9 & 85.4 &--  & --\\\hline
\hline
\textbf{SBGPM}         & \textbf{94.3} &\textbf{89.7} &\textbf{SBGPM}& \textbf{78.7} \\\hline
\end{tabular}

 \begin{tablenotes}
 \item [a] Fusion of multiple features: Gabor, LBP, FPLBP and TPLBP.
 \item [b] The highest approximated results reported  by curve in \cite{Zhang2010}, using Gabor pre-processing.
 \item [c] A pre-precessing step was applied for getting higher performance, i.e. Gamma correction and DoG filter were used in \cite{Tan2007}, and Retina filtering was processed before POEM \cite{Vu2012}.
 \end{tablenotes}
 \end{threeparttable}\label{tab:results_FERETandLFW}
\end{table}

The performance of the SBGPM descriptor was further investigated on age changes (DupI and DupII of the FERET database) and unrestricted real-world images (the LFW database). On these databases, our experiments followed most of previous work by using square root of the features for representation and cosine distance for similarity measure \cite{Wolf2011, Vu2012}. Note that good performance on the two challenging databases heavily relies on sophisticated machine learning models for learning high-level features and advanced classifiers. Development of advanced machine learning methods is beyond the scope of this work.  For an unbiased evaluation, our descriptors were fairly compared to a set of manually-designed features. In our implementation, we applied the Fisher's Linear Discriminant Analysis (FDA) \cite{Belhumeur1997} for classification. Because the FDA cannot be applied directly for face verification on the LFW, the performance on this dataset was evaluated without any learning processing. The correct rates of the SBGPM on two databases are compared to the recent published results in Table~\ref{tab:results_FERETandLFW}.

  The results show that the SBGPM achieved competitive performance to recent descriptors with correct rates reaching $94.3\%$ and $89.7\%$ on DupI and DupII, respectively. The margins between the SBGPM and the closest methods on the list are about $4\%$ on both subsets, which are significant for this challenging dataset. Similarly, the proposed descriptor obtained $78.7\%$ correct rate for face verification on the LFW database, further improving over the closest single descriptor (POEM) by $5\%$ and the best multiple fusion descriptors by about $4\%$ in correct rate. The favorable performance of the SBGPM on these challenging variations further illustrates its highly discriminative power and strong robustness for facial representation.


\section{Conclusion}
This paper has introduced a novel framework for robust facial representation. The proposed structural binary gradient pattern (SBGP) effectively enforces spatial locality in the gradient domain to enhance robustness against both illuminations and local distortions, yet still being compact and computationally efficient by encoding local structures to a set of binary patterns. Theoretical analysis shows that the defined \emph{structural} patterns of the SBGP work extraordinarily as orientational micro edge detectors and thus gain strong spatial locality and orientation properties, leading to effective discrimination. Furthermore, the SBGP is generic and suitable for building fusion models. As an example, the enhanced SBGPM has also been presented as the resulting of combining SBGP and orientational image gradients. Extensive justifications and experimental verifications demonstrate the efficiency of SBGP and SBGPM, and their markedly improved recognition performances over the existing methods on a variety of robustness tests against lighting, expression, occlusion and aging.

%
%
%


\bibliographystyle{natbib/plainnat.bst}

{\footnotesize
\bibliography{mybibfile_PR}}

\begin{thebibliography}{10}
\expandafter\ifx\csname url\endcsname\relax
  \def\url#1{\texttt{#1}}\fi
\expandafter\ifx\csname urlprefix\endcsname\relax\def\urlprefix{URL }\fi
\expandafter\ifx\csname href\endcsname\relax
  \def\href#1#2{#2} \def\path#1{#1}\fi

\bibitem{Huang2012}
W.~Huang, H.~Yin, On nonlinear dimensionality reduction for face recognition,
  Image and Vision Computing 30 (2012) 355--366.

\bibitem{Turk1991}
M.~Turk, A.~Pentland, Eigenfaces for recognition, Journal of Cognitive
  Neuroscience 3 (1991) 71--86.

\bibitem{Belhumeur1997}
P.~N. Belhumeur, J.~P. Hespanha, D.~J. Kriegman, Eigenfaces vs. fisherfaces:
  Recognition using class specific linear projection, IEEE Trans. Pattern Anal.
  Mach. Intell. 19 (1997) 711--720.

\bibitem{Scholkopf1998}
B.~Sch{\"o}lkopf, A.~Smola, K.-R. M{\"u}ller, Nonlinear component analysis as a
  kernel eigenvalue problem, Neural Computation 10 (1998) 1299--1319.

\bibitem{Roweis2000}
S.~T. Roweis, K.~S. Laurence, Nonlinear dimensionality reduction by locally
  linear embedding, Science 290 (2000) 2323--2326.

\bibitem{Yin2010}
H.~Yin, W.~Huang, Adaptive nonlinear manifolds and their applications to
  pattern recognition, Information Sciences 180 (2010) 2649--2662.

\bibitem{Huang2009}
W.~Huang, H.~Yin, Linear and nonlinear dimensionality reduction for face
  recognition, in: IEEE Int’l Conf. on Image Processing (ICIP), 2009, pp.
  3337--3340.

\bibitem{Zhang2009b}
T.~Zhang, Y.~Y. Tang, B.~Fang, Z.~Shang, X.~Liu, Face recognition under varying
  illumination using gradientfaces, IEEE Trans. on Image Processing 18 (2009)
  2599--2606.

\bibitem{Tzimiropoulos2012}
G.~Tzimiropoulos, S.~Zafeiriou, M.~Pantic, Subspace learning from image
  gradient orientations, IEEE Trans. Pattern Anal. Mach. Intell. 34.

\bibitem{Heisele2003}
B.~Heisele, P.~Ho, J.~Wu, T.~Poggiob, Face recognition: component-based versus
  global approaches, Computer Vision and Image Understanding 91 (2003) 6--21.

\bibitem{Vu2012}
N.-S. Vu, A.~Caplier, Enhanced patterns of oriented edge magnitudes for face
  recognition and image matching, IEEE Trans. on Image Processing 21 (2012)
  1352--1365.

\bibitem{Huang2010}
W.~Huang, H.~Yin, A dissimilarity kernel with local features for robust facial
  recognition, in: IEEE Int’l Conf. on Image Processing (ICIP), 2010, pp.
  3785--3788.

\bibitem{Manjunath1996}
B.~Manjunath, W.~Ma, Texture features for browsing and retrieval of image data,
  IEEE Trans. Pattern Anal. Mach. Intell. 18 (1996) 837--842.

\bibitem{Ojala1996}
T.~Ojala, M.~Pietik{\"a}inen, D.~Harwood, A comparative study of texture
  measures with classification based on featured distributions, Pattern
  Recognition 29 (1996) 51--59.

\bibitem{Ojala2002}
T.~Ojala, M.~Pietik{\"a}inen, T.~Maenpaa, Multiresolution gray-scale and
  rotation invariant texture classification with local binary patterns, IEEE
  Trans. Pattern Anal. Mach. Intell. 24 (2002) 971--987.

\bibitem{Liu2002}
C.~Liu, H.~Wechsler, Gabor feature based classification using the enhanced
  fisher linear discriminant model for face recognition, IEEE Trans. on Image
  Processing 11 (2002) 467--476.

\bibitem{Zhang2005}
W.~Zhang, S.~Shan, W.~Gao, X.~Chen, H.~Zhang, Local gabor binary pattern
  histogram sequence (lgbphs): A novel non-statistical model for face
  representation and recognition, in: IEEE Int’l Conf. Computer Vision
  (ICCV), 2005, pp. 786--791.

\bibitem{Zou2007}
J.~Zou, Q.~Ji, G.~Nagy, A comparative study of local matching approach for face
  recognition, IEEE Trans. on Image Processing 16 (2007) 2617--2628.

\bibitem{Chen2010}
J.~Chen, S.~Shan, C.~He, G.~Zhao, M.~Pietik{\"a}ine, X.~Chen, W.~Gao, Wld: A
  robust local image descriptor, IEEE Trans. Pattern Anal. Mach. Intell. 32
  (2010) 1705--1720.

\bibitem{Tan2011}
X.~Tan, B.~Bill~Triggs, Enhanced local texture feature sets for face
  recognition under difficult lighting conditions, IEEE Trans. on Image
  Processing 20 (2011) 1635--1650.

\bibitem{Zhang2007}
B.~Zhang, S.~Shan, X.~Chen, W.~Gao, Histogram of gabor phase patterns (hgpp): A
  novel object representation approach for face recognition, IEEE Trans. on
  Image Processing 16 (2007) 57--67.

\bibitem{Xie2011}
X.~Xie, W.~S. Zheng, J.~Lai, P.~C. Yuen, C.~Y. Suen, Normalization of face
  illumination based on large-and small-scale features, IEEE Trans. Pattern
  Anal. Mach. Intell. 20 (2011) 1807--1821.

\bibitem{Ruiz2009}
J.~A. Ruiz-Hernandez, J.~L. Crowley, A.~Meler, A.~Lux, Face recognition using
  tensors of census transform histograms from gaussian features maps, in:
  British Machine Vision Conference (BMVC), 2009, pp. 1--8.

\bibitem{Chen2006}
T.~Chen, W.~Yin, X.~S. Zhou, D.~Comaniciu, T.~S. Huang, Total variation models
  for variable lighting face recognition, IEEE Trans. Pattern Anal. Mach.
  Intell. 28 (2006) 1519--1524.

\bibitem{Lei2011}
Z.~Lei, S.~Liao, M.~Pietik$\ddot{a}$inen, , S.~Z. Li, Face recognition by
  exploring information jointly in space, scale and orientation, IEEE Trans. on
  Image Processing 20 (2011) 247--256.

\bibitem{Heikkila2009}
M.~Heikkil$\ddot{a}$, M.~Pietik$\ddot{a}$inen, C.~Schmid, Description of
  interest regions with local binary patterns, Pattern Recognition 42 (2009)
  425--436.

\bibitem{Lowe2004}
D.~Lowe, Distinctive image features from scale-invariant keypoints,
  International Journal of Computer Vision 60 (2004) 91--110.

\bibitem{Georghiades2001}
A.~Georghiades, P.~Belhumeur, D.~Kriegman, From few to many: Illumination cone
  models for face recognition under variable lighting and pose, IEEE Trans.
  Pattern Anal. Mach. Intell. 23 (2001) 643--660.

\bibitem{Ahonen2006}
A.~Ahonen, A.~Hadid, M.~Pietik{\"a}inen, Face description with local binary
  patterns: Application to face recognition, IEEE Trans. Pattern Anal. Mach.
  Intell. 28 (2006) 2037--2041.

\bibitem{Kumar2012}
R.~Kumar, A.~Banerjee, B.~C. Vemuri, P.~Hanspeter, Trainable convolution
  filters and their application to face recognition, IEEE Trans. Pattern Anal.
  Mach. Intell. 34 (2012) 1423--1436.

\bibitem{Olshausen1996}
B.~A. Olshausen, D.~J. Field, Emergence of simple-cell receptive field
  properties by learning a sparse code for natural images, Nature 381 (1996)
  607--609.

\bibitem{Marr1980}
D.~Marr, E.~Hildreth, Theory of edge detection, The Royal Society of London.
  Series B, Biological Sciences (1934-1990) 207 (1980) 187--217.

\bibitem{Yang2010}
M.~Yang, L.~Zhang, Local gabor binary pattern histogram sequence (lgbphs): A
  novel non-statistical model for face representation and recognition, in:
  Euro. Conf. Computer Vision (ECCV), 2010, pp. 448--461.

\bibitem{Martinez1998}
A.~M. Martinez, R.~Benavente, The AR Face Database, CVC Technical Report No.24,
  1998.

\bibitem{Martinez2001}
A.~M. Martinez, A.~C. Kak, Pca versus lda, IEEE Trans. Pattern Anal. Mach.
  Intell. 23 (2001) 228--233.

\bibitem{Lee2005}
K.~C. Lee, J.~Ho, D.~Kriegman, Acquiring linear subspaces for face recognition
  under variable lighting, IEEE Trans. Pattern Anal. Mach. Intell. 27 (2005)
  684--698.

\bibitem{Phillips2000}
P.~J. Phillips, H.~Moon, P.~J. Rauss, S.~Rizvi, The feret evaluation
  methodology for face recognition algorithms, IEEE Trans. Pattern Anal. Mach.
  Intell. 22 (2000) 1090--1104.

\bibitem{Huang2007}
G.~Huang, M.~Ramesh, T.~Berg, E.~Learned-Miller, Labeled Faces in the Wild: A
  Database for Studying Face Recognition in Unconstrained Environments,
  Technical report, 07-49, University of Massachusetts, Amherst, 2007.

\bibitem{Wolf2011}
L.~Wolf, T.~Hassner, Y.~Taigman, Effective unconstrained face recognition by
  combining multiple descriptors and learned background statistics, IEEE Trans.
  Pattern Anal. Mach. Intell. 33 (2011) 1978--1990.

\bibitem{Wolf2009}
L.~Wolf, T.~Hassner, Y.~Taigman, The one-shot similarity kernel, in: IEEE
  Int’l Conf. Computer Vision (ICCV), 2009.

\bibitem{Hermosilla2012}
G.~Hermosilla, J.~Ruiz-del Solar, R.~Verschae, M.~Correa, A comparative study
  of thermal face recognition methods in unconstrained environments, Pattern
  Recognition 45 (2012) 2445--2459.

\bibitem{Liao2009}
S.~Liao, A.~Chung, Face recognition with salient local gradient orientation
  binary patterns, in: IEEE Int’l Conf. on Image Processing (ICIP), 2009, pp.
  3317--3320.

\bibitem{Bosch2007}
A.~Bosch, A.~Zisserman, X.~Munoz, Representing shape with a spatial pyramid
  kernel, in: ACM International Conference on Image and Video Retrievel, 2007.

\bibitem{Zhang2009a}
T.~Zhang, B.~Fang, Y.~Yuan, Y.~Y. Tang, Z.~Shang, D.~Li, F.~Lang, Multiscale
  facial structure representation for face recognition under varying
  illumination, Pattern Recognition 42 (2009) 251--258.

\bibitem{Lu2013}
J.~Lu, Y.-P. Tan, G.~Wang, Discriminative multi-manifold analysis for face
  recognition from a single training sample per person, IEEE Trans. Pattern
  Anal. Mach. Intell. 35 (2013) 39--51.

\bibitem{Deng2012}
W.~Deng, J.~Hu, J.~Guo, Extended src: Undersampled face recognition via
  intra-class variant dictionary, IEEE Trans. Pattern Anal. Mach. Intell. 34
  (2012) 1864--1870.

\bibitem{Deng2010}
W.~Deng, J.~Hua, J.~Guo, W.~Cai, D.~Feng, Robust, accurate and efficient face
  recognition froma single training image: A uniform pursuit approach, Pattern
  Recognition 43 (2010) 1748--1762.

\bibitem{Yang2012a}
M.~Yang, L.~Zhang, S.~Shiu, D.~Zhang, Gabor feature based robust representation
  and classification for face recognition with gabor occlusion dictionary,
  Pattern Recognition 45 (2012) 2445--2459.

\bibitem{Chen2012}
C.-F. Chen, W.~C.-P., Y.-C. Wang, Low-rank matrix recovery with structural
  incoherence for robust face recognition, in: IEEE Int’l Conf. Computer
  Vision and Pattern Recognition (CVPR), 2012.

\bibitem{Wright2009}
J.~Wright, A.~Y. Yang, A.~Ganesh, S.~S. Sastry, Y.~Ma, Robust face recognition
  via sparse representation, IEEE Trans. Pattern Anal. Mach. Intell. 31 (2009)
  210--227.

\bibitem{Naseem2010}
I.~Naseem, R.~Togneri, M.~Bennamoun, Linear regression for face recognition,
  IEEE Trans. Pattern Anal. Mach. Intell. 32 (2010) 2106--2112.

\bibitem{He2011}
R.~He, W.-S. Zheng, B.-G. Hu, Maximum correntropy criterion for robust face
  recognition, IEEE Trans. Pattern Anal. Mach. Intell. 33 (2011) 1561--1576.

\bibitem{Zhang2011}
L.~Zhang, M.~Yang, X.~Feng, Sparse representation or collaborative
  representation: Which helps face recognition?, in: IEEE Int’l Conf.
  Computer Vision (ICCV), 2011.

\bibitem{Theodorakopoulos2011}
I.~Theodorakopoulos, I.~Rigas, G.~Economou, S.~Fotopoulos, Face recognition via
  local sparse coding, in: IEEE Int’l Conf. Computer Vision (ICCV), 2011.

\bibitem{Yang2012b}
M.~Yang, L.~Zhang, D.~Zhang, S.~Wang, Relaxed collaborative representation for
  pattern classification, in: IEEE Int’l Conf. Computer Vision and Pattern
  Recognition (CVPR), 2012.

\bibitem{Wang2010}
J.~Wang, J.~Yang, K.~Yu, F.~Lv, T.~Huang, Y.~Gong, Locality-constrained linear
  coding for image classification, in: IEEE Int’l Conf. Computer Vision and
  Pattern Recognition (CVPR), 2010, pp. 3360--3367.

\bibitem{Geng2009}
C.~Geng, X.~Jiang, Face recognition using sift features, in: IEEE Int’l Conf.
  on Image Processing (ICIP), 2009, pp. 3313--3316.

\bibitem{Luo2007}
J.~Luo, Y.~Ma, E.~Takikawa, S.~Lao, M.~Kawade, L.~L.-B., Person-specific sift
  features for face recognition, in: IEEE Int’l Conf. on Acoustics, Speech
  and Signal Processing (ICASSP), 2007, pp. 593--596.

\bibitem{Deng2005}
W.~Deng, J.~Hu, J.~Guo, Gabor-eigen-whiten-cosine: A robust scheme for face
  recognition, in: Analysis and Modelling of Faces and Gestures (AMFG), 2005,
  pp. 336--349.

\bibitem{Pinto2008}
N.~Pinto, J.~DiCarlo, D.~Cox, Establishing good benchmarks and baselines for
  face recognition, in: Faces in Real- Life Images Workshop in European
  Conference on Computer Vision, 2008.

\bibitem{Zhang2010}
B.~Zhang, Y.~Gao, S.~Zhao, J.~Liu, Local derivative pattern versus local binary
  pattern: Face recognition with high-order local pattern descriptor, IEEE
  Trans. on Image Processing 19 (2010) 533--544.

\bibitem{Nguyen2009}
H.~V. Nguyen, L.~Bai, L.~Shen, Local gabor binary pattern whitened pca: A novel
  approach for face recognition from single image per person, in: Advances in
  Biometrics, 2009, pp. 269--278.

\bibitem{Tan2007}
X.~Tan, B.~Triggs, Fusing gabor and lbp feature sets for kernel-based face
  recognition, in: Analysis and Modelling of Faces and Gestures (AMFG), 2007,
  pp. 235--249.

\end{thebibliography}

\end{document}